\documentclass{article}
\usepackage{graphicx} % Required for inserting images
\usepackage[margin=0.7in]{geometry}
\usepackage{amsthm,amsmath,amsfonts,amssymb,dsfont,placeins}
\usepackage{hyperref,natbib,babel,algpseudocode,algorithm,bm}

\newcommand{\mat}[1]{\mathbf{#1}}

\newtheorem{theorem}{Theorem}[section]
\newtheorem{lemma}[theorem]{Lemma}
\newtheorem{proposition}[theorem]{Proposition}
\newtheorem{corollary}[theorem]{Corollary}
\newtheorem{definition}[theorem]{Definition}
\newtheorem{claim}[theorem]{Claim}
\newtheorem{remark}[theorem]{Remark}

\newtheorem*{theorem*}{Theorem}
\newtheorem*{lemma*}{Lemma}
\newtheorem*{proposition*}{Proposition}
\newtheorem*{corollary*}{Corollary}
\newtheorem*{definition*}{Definition}
\newtheorem*{claim*}{Claim}
\newtheorem*{remark*}{Remark}

\title{Towards minimax optimal algorithms for Active Simple Hypothesis Testing.}
\author{Sushant Vijayan,\\
Tata Institute of Fundamental Research,\\
email: sushant.vijayanq@tifr.res.in.
}
\date{}
\begin{document}
\maketitle
\begin{abstract}
We study the Active Simple Hypothesis Testing (ASHT) problem, a simpler variant of the Fixed Budget Best Arm Identification problem. In this work, we provide novel game theoretic formulation of the upper bounds of the ASHT problem. This formulation allows us to leverage tools of differential games and Partial Differential Equations (PDEs) to propose an approximately optimal algorithm that is computationally tractable compared to prior work. However, the optimal algorithm still suffers from a curse of dimensionality and instead we use a novel link to Blackwell Approachability to propose an algorithm that is far more efficient computationally. We show that this new algorithm, although not proven to be optimal, is always better than static algorithms in all instances of ASHT and is numerically observed to attain the optimal exponent in various instances. 
\end{abstract}

\section{Introduction.}\label{intro}
We study the problem of Active Simple Hypothesis Testing (ASHT) where an agent is faced with the problem of choosing between $m$ different simple hypotheses after observing $T$ samples. At the end of $T$ samples, it has to output one of the $m$ hypothesis. The distinguishing difference from the usual hypothesis testing scenario is the ability to choose one of $K$ actions and observe the corresponding sample for that action. This ability to control the samples in this way makes the problem more interesting and difficult compared to the usual hypothesis testing with no control over the sample generation. The performance of the agent is measured in terms of the error probability its decision incurs. The above theoretical problem is a model for many practical scenarios- A cosmetic drug trial often involve a testing period where the outcome of interest is to identify the best product after the trial period, choosing a channel from a set of channels before commencing communications, placement of sensors in certain set of positions so as to minimize signal error. Any situation which require a period of testing before committing to a final decision with only certain fixed budget of samples (that is an inability to request additional samples) can be modeled effectively using ASHT and its more general version -  Fixed Budget Best Arm Identification (FB-BAI).\\

We intend to study the ASHT problem in the large deviation setting with the quantity of interest being the minimax error exponent over the hypotheses, that is, the worst case error exponent over the hypotheses. The relationship between FB-BAI and ASHT is a straightforward one - ASHT is a special variant of the FB-BAI \footnote{In the literature  of FB-BAI, we have $m=K$, but those results easily extend to $m \neq K$ case.}, wherein we consider only simple singleton hypotheses. The FB-BAI allows for more general composite (even continuum) hypotheses. The elements of these hypotheses are assumed to be multi-armed bandits or memoryless channels whose samples can be controlled through choosing actions or inputs. We choose the minimax error exponent as the performance measure, since that would imply at least that minimax exponential rate of error probability decay on all hypotheses. This is a good measure to choose when one is interested in guaranteeing a good performance across all instances. \\

We study the ASHT problem since often the more general FB-BAI problem is quite difficult to understand. Further, ASHT problem itself is not well understood in terms of the optimal exponents and the algorithmic behavior attaining those optimal exponents. Further, ASHT is also a useful model for practical scenarios like channel identification. We hope the insights gained in the study of ASHT problem will help in  understanding the FB-BAI problem as well.\\

\textbf{Literature Review:} The basic problem of hypothesis testing with active sample control and a sample budget has been studied in many different research communities. In particular the problem has evoked considerable interest within the Simulation, Operations Research, Multi-Armed  Bandit and Information Theory communities. The problem has been given different names in these communities. We will try to present the development of this research work in the various communities. We do not attempt to be exhaustive and only give a selection of the various works and results on the problem from the various communities.\\
In the operations research community the problem has been studied under the name 'Ranking \& Selection' (RS) problem. The RS literature often assumes the underlying distributions are gaussians with unknown means and variances. The problem was often cast as a Dynamic Programming problem and studied under both frequentist and bayesian frameworks. Based on a notion of optimal static allocation for a given instance a number of algorithms have been proposed. The Optimal Computing Budget Allocation (OCBA) was studied in \cite{chen1996lower} and \cite{chen2000simulation}. \cite{glynn2004large} showed that this procedure is optimal among all static strategies in a large deviation sense by . However, these procedures presume the knowledge of the parameters of the unknown underlying hypotheses and in general cannot be achieved. 
\cite{he2007opportunity} and \cite{gao2017new} proposed a variant of OCBA  with the error probability replaced with the simple regret as the objective. The latter motivated their procedures using large deviation theory. Static allocation based strategies like OCBA develop strategies based on estimates of quantities like means and variances. A different approach taken by using a Bayesian formalism was to approximate the corresponding DP solution with finite step look ahead policies based on various heuristics. \cite{chick2001new}, \cite{chick2010sequential} develop Expected Value of Information (EVI) approach which can be viewed as an approximated dynamic programming solution. An alternative one step look ahead policy was the Knowledge Gradient (KG) approach developed in \cite{frazier2008knowledge}. As far as we are aware, these policies do not come with any guarantee of exact or approximate optimality. More details and references can be obtained from the surveys of \cite{kim2006selecting} and \cite{hong2021review}.\\

In information theory community the problem has been studied under the names of Active Hypothesis testing (AHT), Controlled Sensing (CS) and Channel Discrimination/Identification (CI). In the fixed budget setting \cite{berlekamp1964block} studied the problem of block channel coding with fixed messages. This can be considered as a special case of the CS problem where the agent has access to the hypothesis. A multi-channel identification was studied in \cite{mitran2005error} with the constraint of finite number of past observations available to the agent. These results were subsumed by the result on the fixed budget setting of \cite{nitinawarat2013controlled}. We will take a more detailed look of \cite{nitinawarat2013controlled} results in section \ref{prev_relevant_work}. In 
\cite{naghshvar2013sequentiality} they try to study and approximate the underlying DP solutions and manage to obtain upper and lower bounds for the bayesian error probability. \cite{hayashi2009discrimination} studied the binary CD problem and managed to characterize the optimal minimax exponent in terms of the Chernoff Distances between arms/actions. We will examine the results of \cite{hayashi2009discrimination} in detail in section \ref{prev_relevant_work}. In \cite{kartik2021fixed} they study the problem of minimizing the error probability of wrongly declaring hypothesis $i$ when the true hypothesis $j \neq i$. In addition the agent may also choose to declare the result as inconclusive and has to ensure the probability of correctly identifying the hypothesis is moderately large. They characterize the optimal exponent and propose algorithms that achieve the optimal exponent.\\

The FB-BAI was introduced in the bandit literature by the papers \cite{audibert2010best} \cite{bubeck2011pure}. They introduced the Successive Rejection (SR) algorithm and proved its order-wise optimality of the error-probability upto $\log(K)$ factors in the exponent. \cite{gabillon2012best} proposed a single meta algorithm UGapE for both the sequential version - Fixed confidence Best Arm Identification (FC-BAI) and FB-BAI and gave sample complexity bounds for the former and error probability bounds for the latter. \cite{karnin2013almost} proposed Sequential Halving (SH) algorithm and improved the $O(K^2)$ factor multiplying the error exponent in SR to $O(\log(K))$. \cite{carpentier2016tight} improved the error probability lower bounds by $O(\log(K))$ factor in the error exponent and showed the order-wise optimality of SR and SH algorithms \footnote{It is to be noted that the gap between order-wise optimal in exponent and actually attaining the optimal minimax exponent can be significant.}. In \cite{garivier2016optimal} an instance specific sample complexity lower bound was derived and an optimal algorithm asymptotically matching this lower bound for every bandit instance was proposed. In the same work, a similar data processing based instance specific upper bound for the error exponent was derived for FB-BAI and the question of obtaining an asymptotically instance wise optimal algorithm was posed. This remained the outstanding question in FB-BAI for a few years (for example see \cite{pmlr-v178-open-problem-qin22a}). It was recently shown by \cite{degenne2023existence} for the important case of two armed bernoulli bandits that no such algorithm could exist. A stronger result showing that no algorithm could dominate even the simple static uniform strategy in the error exponent was shown by \cite{wang2023universally} in the same specific case of two armed bernoulli bandits. It is to be remarked that a Hoeffding style result (Theorem 2)  of \cite{hayashi2009discrimination} can be used to prove similar negative results for FB-BAI problems from other distributions. In  \cite{wang2023best} they prove a general large deviation instance specific lower bound for the error exponent and utilize it to characterize the error exponent of SR. In an important work \cite{komiyama2022minimax} studied the problem of achieving the minimax exponent in FB-BAI and showed through data processing inequality a series of upper bounds for the minimax error exponent and gave a theoretically optimal algorithm for achieving it. Our work relies on and extends on their work in ASHT setting. We will present a detailed review of the results in \cite{komiyama2022minimax} in section \ref{prev_relevant_work}.\\
In general, it has been difficult to characterize the optimal minimax exponent and provide optimal algorithms achieving them in FB-BAI (see \cite{pmlr-v178-open-problem-qin22a} and the conclusion section of \cite{komiyama2022minimax}). Even the simpler variant ASHT has considered difficult to resolve for even $m=3$ (see section 6 of \cite{nitinawarat2013controlled}).\\

\textbf{Contribution of the current work:} 
\begin{enumerate}
    \item We recast the upper bounds of minimax error exponent of \cite{komiyama2022minimax} into a game theoretic framework. Using this we can characterize the optimal exponent as the value of a zero sum differential game.
    \item The game theoretic interpretation allows us to use PDE methods to establish alternate upper bounds, effective approximation schemes for computing the optimal exponent when $m$ is small and propose an approximately optimal algorithm which is computationally tractable for small $m$. This unlike the algorithm proposed by \cite{komiyama2022minimax} which assumes access to a computationally intractable oracle and hence is not implementable even for small $m$.
    \item The PDE based algorithms that we propose suffer from curse of dimensionality in $m$. Instead we propose an alternative algorithm based on Blackwell Approachability. We show theoretically that this algorithm is uniformly better than any static strategy for all instances of ASHT. Further, we observe numerically that this algorithm can achieve the optimal exponent in a number of problem instances and is not as computationally expensive as PDE methods or that of \cite{komiyama2022minimax}.
\end{enumerate}
In terms of technical novelty our reinterpretation of the upper bounds of \cite{komiyama2022minimax} facilitates introduction of methods of differential games and PDEs to study this problem. In terms of algorithmic novelty the introduction of Blackwell approachability to study ASHT seems new in the literature. The analysis of approachability also makes us ideas from convex and non-smooth analysis.\\

\textbf{Organization of the paper:} We introduce the problem formally in section \ref{prelim}. We review in detail the most relevant work in \ref{prev_relevant_work}. We present our results of recasting the upper bounds of \cite{komiyama2022minimax} into game-theoretic framework in section \ref{info_th_lb}. In section \ref{sec:numerical_pde_scheme} we present a simple explicit upwind scheme to compute the optimal minimax exponent and disprove a conjecture of \cite{komiyama2022minimax}. Further, in the same section we propose and prove an approximately optimal algorithm. In section \ref{sec:approachability}, we propose a meta strategy that encompasses a wide variety of algorithms and consider the relation of Blackwell approachabilty in studying the minimax exponent problem.Section \ref{sec:construction b_set} shows the construction of a suitable $B$-set, a crucial ingredient that allows us to prove the better than static performance of the approachability algorithm. Section \ref{Summary_of_work} provides a summary of this work and Section \ref{future_work} concludes with some new directions and problems to facilitate further research into the problem. The proofs of all the results are presented in the Appendix.

\section{Preliminaries.}\label{prelim}
We will denote a K-armed bandit \emph{instance} as $\nu$ and the the arm distribution of each arm as $\nu_a$. The mean of the arms are given by $\mu_a$. $D(p||q)$ is the KL divergence between two distributions $p,q$. $\xi_a$ is the  set of all the bandits with the best arm as $a$. Then, $\xi=\cup^{K}_{a=1}\xi_a$ is the collection of all bandits under consideration and will be called the \emph{bandit class}.  We assume there is a unique best arm value for each instance $\nu$. Hence, we can define the \emph{best arm map} as a surjective function $i^{*}: \xi \rightarrow [K]$, such that $i^{*}(\nu)=a$ iff $\nu \in \xi_a$.

\subsection{Assumptions.}
\begin{enumerate}
    \item All arm distributions are assumed to be have full support over a common finite set.
    \item Each $\xi_a$ is a singleton \footnote{In many places particularly in Sections \ref{prev_relevant_work} and \ref{info_th_lb} we will sometimes relax this assumption to finiteness or even allow $\xi_a$ to be a continuum.}.
\end{enumerate}
This simpler variant of the fixed budget problem under assumption 1 is called as the \textbf{Active Simple Hypothesis Testing (ASHT)}. As each hypothesis $\xi_a$ is a singleton we call this a simple hypothesis testing problem. The active part refers to the ability of the agent to choose/control the samples it gets.
\subsection{Problem Formulation.}
  We have, as in a typical fixed budget problem, that there is a sample horizon $T$ and that any algorithm $\mathbb{A}_T$ has $T$ sampling rules denoted by $a_t$, $\forall t \in [T],$ which are adapted to the sample filtration and a decision rule $\hat{i}_T \in [K]$ adapted to the final filtration at $T$. 
Define the \emph{error probability} of an algorithm $\mathbb{A}_T$ on the instance $\nu$ as:
\begin{equation}\label{err_prob_eq}
P_e(\nu,\mathbb{A}_T)= P_{\nu,\mathbb{A}_T}(i^{*}(\nu) \neq \hat{i}_T),
\end{equation}
where the randomness is from the stochasticity of the samples generated by $\nu$ and any internal randomness of the algorithm.
\begin{remark}
We have chosen to call algorithm as one associated with a specific horizon $T$. We will call a \emph{sequence of algorithms}, as $\mathbb{A}=(\mathbb{A}_T)_{T \in \mathbb{N}}$. We denote the set of all possible sequence of algorithms $\mathbb{A}$ as $\mathcal{A}$.
\end{remark}
\begin{definition}
The \textbf{error exponent} of a sequence of algorithms $\mathbb{A}$ on an instance $\nu$ is defined as:
\begin{equation}\label{err_exp_eq}
e(\nu,\mathbb{A}) = \liminf_{T \to \infty}-\frac{\log(P_e(\nu,\mathbb{A}_T))}{T}
\end{equation}
and the \textbf{minimal} exponent over bandit class $\xi$ as 
\begin{equation}\label{min_exp_eq}
e_m(\xi,\mathbb{A}) = \underset{\nu \in \xi}{\inf}e(\nu,\mathbb{A}).
\end{equation}
\end{definition}

  Henceforth, we  use the word "algorithm" interchangeably for algorithmic sequence $\mathbb{A}$ and horizon dependent algorithm $\mathbb{A}_T$ and where necessary, we will distinguish the two notions using their respective symbols.

\begin{definition}[Minimax optimality]
An algorithm $\mathbb{A}^{*}$ is said to be \textbf{minimax optimal} if:
\begin{equation*}
e_m(\xi,\mathbb{A}^{*})=\underset{\mathbb{A} \in \mathcal{A}}{\max}\hspace{0.2cm}e_{m}(\xi,\mathbb{A}).
\end{equation*}
\end{definition}
  The problem this work attempts to solve is:
\\
\\
\emph{\large Can we come up with a computationally tractable algorithm $A$ which is minimax optimal for any given generic bandit class $\xi$?}

\section{Previous relevant work.}\label{prev_relevant_work}
There are three main works that have considered the problem and are relevant to this work.
\\

  \underline{\textbf{\cite{hayashi2009discrimination}}:} This work considers the simplest possible setting where the hypotheses are simple and binary. Let us assume the the two possible bandit hypothesis are : $\nu, \nu^{'}$. Under certain regularity assumptions on the underlying distribution densities it shows the following result:
\begin{theorem}[Theorem 2 \cite{hayashi2009discrimination}]\label{hoeffding_result}
Let us define the following quantity between two distributions with a common support
\begin{equation}
 \phi(s,p,q) = \log \left( \sum_{l \in \text{supp(p)}} p^{s}(l)q^{1-s}(l) \right).
\end{equation}
Suppose the following condition holds:\\
(a) The second derivative of $\phi$ wrt $s$ is bounded for all arms and both pairs of hypothesis, that is,
\begin{equation*}
\underset{a \in [K]}{\max}\underset{s \in [0,1]}{\sup}\frac{d^{2} \phi(s,\nu_a,\nu^{'}_a)}{d s^{2}} < \infty.
\end{equation*}
Then, if an algorithm $\mathbb{A}$ achieves an error exponent of atleast $r$ on the instance $\nu$, that is $e(\nu,\mathbb{A}) \geq r$, then we necessarily have:
\begin{equation}
e(\nu^{'},\mathbb{A}) \leq \underset{a \in [K]} {\max} \hspace{0.2cm} \underset{Q:\forall a, D(Q_a||\nu_a) \leq r}{\min} \hspace{0.2cm} D(Q_a || \nu^{'}_a).    
\end{equation}
\end{theorem}
  Clearly if the role of $\nu,\nu^{'}$ interchanges a similar result will hold. The result tells us that if an algorithm $\mathbb{A}$ does well in an instance then its performance in an alternate instance is limited. Note that in this case $\xi=\{\nu,\nu^{'}\}$. Hence, from the above theorem one can conclude that:
\begin{equation}
\underset{\mathbb{A} \in \mathcal{A}} {\max} \hspace{0.2cm} e_{m}(\xi,\mathbb{A}) \leq \underset{r \in [0,1]}{\max} \min \left \{r,  \underset{a \in [K]} {\max} \hspace{0.2cm} \underset{Q:\forall a, D(Q_a||\nu_a) \leq r}{\min} \hspace{0.2cm} D(Q_a || \nu^{'}_a) \right \}.
\end{equation}
On the RHS, in the minimum expression the first quantity is strictly increasing in $r$ and the other quantity \[\underset{a \in [K]} {\max} \hspace{0.2cm} \underset{Q:\forall a, D(Q_a||\nu_a) \leq r}{\min} \hspace{0.2cm} D(Q_a || \nu^{'}_a) \]is continuous and decreasing in $r$. Hence, the optima on the RHS is achieved exactly at $r^{*}$ where
\begin{equation*}
r^{*}=\underset{a \in [K]} {\max} \hspace{0.2cm} \underset{Q:\forall a, D(Q_a||\nu_a) \leq r^{*}}{\min} \hspace{0.2cm} D(Q_a || \nu^{'}_a)
\end{equation*}
Now, one definition of Chernoff information between two distributions $\nu_a, \nu^{'}_a$ is the unique fixed point of the following optimization function:
\begin{equation*}
f(r)=\underset{D(Q_a|| \nu_a) \leq r}{\min}D(Q_a|| \nu^{'}_a)
\end{equation*}
and let us denote it by $C(\nu_a,\nu^{'}_a)$ (that is $r=f(r)=C(\nu_a,\nu^{'}_a)$). Note that Chernoff information is symmetric in its arguments. Then, we can conclude that:
\begin{equation*}
r^{*}=\underset{a \in [K]} {\max} \hspace{0.2cm} C(\nu_a,\nu^{'}_a).
\end{equation*}
Now this lower bound could be extended to more general $\xi$ in the following fashion:
\begin{equation}\label{hayashi_lb}
\underset{\mathbb{A} \in \mathcal{A}} {\max} \hspace{0.2cm} e_{m}(\xi,\mathbb{A}) \leq  \underset{a \neq a^{'}}{\min}\underset{\underset{\nu^{'} \in \xi_{a^{'}}}{\nu\in \xi_a}}{\min} \underset{b \in [K]} {\max} \hspace{0.2cm} C(\nu_b,\nu^{'}_b) \triangleq R^{ub}. 
\end{equation}
\\

  \underline{\textbf{\cite{nitinawarat2013controlled}}:} Building off \cite{hayashi2009discrimination}, this work also showed the lower bound \eqref{hayashi_lb} for the case when the $\xi_a$ are singletons (that is simple hypothesis) but for arbitrary arms not just simple binary hypotheses  like \cite{hayashi2009discrimination}. Of course, \eqref{hayashi_lb} is valid for the case when $\xi_a$ are composite hypotheses. An interesting result obtained by \cite{nitinawarat2013controlled} is that they showed in the case of simple hypotheses the best open loop or static strategies can achieve is the following:
\begin{theorem}[Theorem 1 \cite{nitinawarat2013controlled}]\label{best_static_result}
Assume that the the sampling rules are given by pre-defined strategy that does not adapt to the incoming samples. Let us denote the class of such static algorithms as $\mathcal{A}_{static}$. Then if $\xi_a$ are simple hypotheses (that is $\xi_a$ are singleton sets) we have:
\begin{equation}\label{best_static_exp}
\underset{\mathbb{A} \in \mathcal{A}_{static}} {\max} \hspace{0.2cm} e_{m}(\xi,\mathbb{A}) = \underset{w \in \Delta_K}{\max} \underset{\underset{\nu^{'} \in \xi_{a^{'}}}{\nu \in \xi_a}}{\min}\underset{s \in [0,1]}{\max} -\sum_{b \in [K]}w_b \log \left( \sum_{x \in \mathcal{X}} (\nu^{'}_b(x))^{s} \nu^{1-s}_b(x)\right) \triangleq R_{static},
\end{equation}
where $\Delta_K$ denotes the probability simplex over the arms $[K]$. 
\end{theorem}
\begin{remark}
Of course, one can use their methods to extend the result to composite hypotheses as well, just as was done for \hyperref[hoeffding_result]{Theorem 2.1}.Further, it is straightforward to show the following consistency result between \hyperref[best_static_result]{Theorem 2.2} and \eqref{hayashi_lb}:
\begin{equation}\label{consistency_1}
\underset{w \in \Delta_K}{\max} \underset{\underset{\nu^{'} \in \xi_{a^{'}}}{\nu \in \xi_a}}{\min}\underset{s \in [0,1]}{\max} -\sum_{b \in [K]}w_b \log \left( \sum_{x \in \mathcal{X}} (\nu^{'}_b(x))^{s} \nu^{1-s}_b(x)\right) \leq \underset{a \neq a^{'}}{\min}\underset{\underset{\nu^{'} \in \xi_{a^{'}}}{\nu\in \xi_a}}{\min} \underset{b \in [K]} {\max} \hspace{0.2cm} C(\nu_b,\nu^{'}_b),
\end{equation}
where we utilize the alternate description of Chernoff information as 
\begin{equation}\label{alt_ci_desc}
C(\nu_b,\nu^{'}_b)= \underset{s \in [0,1]}{\max} -\log \left( \sum_{x \in \mathcal{X}} (\nu^{'}_b(x))^{s} \nu^{1-s}_b(x)\right).
\end{equation}
\end{remark}
  In the simple binary hypothesis case \eqref{hayashi_lb} reduces to :
\begin{equation*}
\underset{\mathbb{A} \in \mathcal{A}} {\max} \hspace{0.2cm} e_{m}(\xi,\mathbb{A}) \leq \underset{b \in [K]} {\max} \hspace{0.2cm} C(\nu_b,\nu^{'}_b). 
\end{equation*}
Hence, a way to achieve the minimax exponent in this special simple binary hypothesis testing case is to simply play the arm that discriminates the most between the two instances in terms of the Chernoff Information. This is a static strategy! One might be tempted to think that adaptivity might not play role based only on this special case. This is refuted by an example given in Section 5 of \cite{nitinawarat2013controlled}. They show that even in the case of simple hypotheses but with $K=3$ an adaptive algorithm outperforms the optimal static benchmark \eqref{best_static_exp}.
\\

  \underline{\textbf{\cite{komiyama2022minimax}}:} This work gave stronger upper bounds on $\underset{\mathbb{A} \in \mathcal{A}} {\max}\hspace{0.2cm} e_{m}(\xi,\mathbb{A})$. Let us define certain preliminaries to succinctly state their result. For any $B \in \mathbb{N}$, let us denote the empirical bandit distributions observed between $(lT/B+1)^{th}$ sample to $(l+1)T/B^{th}$ sample as $Q_{l}$ (here $l \in \mathbb{N}$). Note that $Q_{l}=(Q_{l,1},Q_{l,2},\dots,Q_{l,K})$ is a K-tuple of empirical distributions $Q_{l,a}$ associated with each arm $a$. Basically, $Q_j$ is the empirical bandit distributions we create from the $l^{th}$ batch of $T/B$ samples. 
\begin{remark}
This construction maybe carried out for arbitrarily large $B$ as we are in the asymptotic large deviation setting where we assume the a-priori limit $T \to \infty$. Practically, we are talking of finite but large $T$ such that $B=o(T)$.
\end{remark}
  Let $Q^{l}=(Q_1,Q_2,\dots,Q_l)$. Using this notation to define
\begin{equation*}
r^{B}(Q^{B})=(r_1(Q^{1}),r_2(Q^{2}),\dots, r_B(Q^{B})).
\end{equation*}
Here $r_l(Q^l)$ is assumed to be an element from $\Delta_K$ that depends on  $Q^{l}$. The interpretation is that $r_l(Q^{l})$ determines the allocation of samples to each arm during the sampling period between $(l-1)T/B+1$ th sample to $lT/B$ sample. Note that crucially $r_l$ is \emph{not adapted} to the usual causal filtration of samples, instead it is allowed to peak ahead by a batch and then decide the allocation (thus breaking causality and is not adapted to the filtration). $r^{B}$ is then just the collection of these causality breaking sampling allocation rules.
\\

  Finally, we also define an \emph{causality adapted} decision rule $J$ that only depends on $Q^{B}$, that is $J(Q^{B}) \in [K]$.
\begin{theorem}[Theorem 2 \cite{komiyama2022minimax}]\label{batched_B_lb}
We state that for any $B \in \mathbb{N}$, we have that:
\begin{equation}\label{batched_B_lb_eq}
\underset{\mathbb{A} \in \mathcal{A}} {\max}\hspace{0.2cm} e_{m}(\xi,\mathbb{A}) \leq \underset{r^{B}(.),J(.)}{\sup} \hspace{0.2cm} \underset{Q^{B}}{\inf} \underset{\underset{J(Q^{B}) \neq i^{*}(\nu)}{\nu \in \xi}}{\inf} \frac{1}{B} \sum^{B}_{l=1}\sum_{a \in [K]}r_{l}(Q^{l})(a)D(Q_{l,a}||\nu_a) \triangleq R^{go}_{B},
\end{equation}
where the outermost supremum is taken over all possible causality breaking allocation rules $r(.)$ and causality adapted decision rule $J(.)$, where they are treated as functions taking the argument $Q^{B}$.
\end{theorem}
  The work then shows that $R^{go}_{B} \leq R^{go}_1$ and that the limit $B \to \infty$ exists,that is:
\begin{equation}\label{lim_defn_r_go_infty}
\underset{B \to \infty}{\lim} R^{go}_B=R^{go}_{\infty}.
\end{equation}
Hence we have the following bound:
\begin{equation}
\underset{\mathbb{A} \in \mathcal{A}} {\max}\hspace{0.2cm} e_{m}(\xi,\mathbb{A}) \leq R^{go}_{\infty}.
\end{equation}
Further, they showed that assuming access to the optimal oracle $r^{*}(.),J^{*}(.)$ in the defining optimization of $R^{go}_{B}$ they could provide an algorithm (Delayed Optimal Tracking (DOT)) that matches the exponent $R^{go}_{\infty}$. This presumed access to these oracles, however, is impractical as they cannot be feasibly implemented (\emph{curse of dimensionality} kicks in even for more moderate sized $B$). Instead, they run experiments where these oracles are replaced with suitably trained neural networks \footnote{They call this algorithm Tracking by Neural Network (TNN).} and report good empirical performance.
\section{Recasting Information theoretic bounds.}\label{info_th_lb}
\subsection{Consistency Results}
We first show an equivalent representation of $R^{go}_B$:
\begin{proposition}\label{eq_rep_R_go_B}
\begin{equation}\label{eqv_rep_R_go_B_eq}
R^{go}_B= \underbrace{\underset{Q_1}{\inf} \hspace{0.1cm}\underset{w_1}{\sup}\hspace{0.1cm}\underset{Q_2}{\inf}\hspace{0.1cm}\underset{w_2}{\sup}\ldots\underset{Q_B}{\inf}\hspace{0.1cm}\underset{w_B}{\sup}}_{B \text{-times}} \underset{j}{\max}\hspace{0.1cm}\underset{j \neq i}{\min} \underset{\nu \in \xi_i}{\inf} \left( \frac{1}{B} \sum^{B}_{l=1}\sum_{a \in [K]}w_{l}(Q^{l})(a)D(Q_{l,a}||\nu_a) \right)
\end{equation}
where each $w_i$ is supremized over the probability simplex $\Delta_K$.
\end{proposition}
  The proof is given in Appendix \ref{proof_prop_1}. This representation shows that this information theoretic lower bound (for error probability) is the value of an adversarial repeated game played between nature and the agent. This has the Stackelberg structure where nature plays $Q_1$, the agent responds to it with $w_1$ and then nature plays $Q_2$ and so on. Here, nature is the leader and the agent is the follower. This is in reverse of the actual structure of play in the practical scenario, wherein the agent plays first and then nature responds. Nature tries to minimize the quantity :
  \begin{equation}\label{terminal_cost}
  \underset{j}{\max}\hspace{0.1cm}\underset{j \neq i}{\min} \underset{\nu \in \xi_i}{\inf} \left( \frac{1}{B} \sum^{B}_{l=1}\sum_{a \in [K]}w_{l}(Q^{l})(a)D(Q_{l,a}||\nu_a) \right)
  \end{equation}
  while the agent tries to maximize the same quantity. For any bandit $\nu \in \xi_i$ the term:
  \begin{equation*}
 \frac{1}{B} \sum^{B}_{l=1}\sum_{a \in [K]}w_{l}(Q^{l})(a)D(Q_{l,a}||\nu_a)    
  \end{equation*}
  refers to its negative log-likelihood of the samples. Let us assume for now that the decision rule is fixed to be the maximum likelihood rule, that is, the bandit with the highest likelihood over the samples is the output. Now the quantity \eqref{terminal_cost} is the negative likelihood of the bandit with the smallest negative likelihood of being wrong under the maximum likelihood rule. Interpreted differently, for a given sample path, the second most likely bandit is the one with the highest likelihood of being in the error set under the maximum likelihood decision rule. Nature tries to maximize this likelihood \footnote{Here we are talking about the likelihood while in eq. \eqref{terminal_cost} we were discussing the negative log-likelihood. } while the agent tries to minimize it.
  \begin{remark}
   The maximum likelihood rule can be heuristically argued as the optimal rule for our problem as follows: For an ASHT problem, the optimal bayes exponent under a fixed prior is the same as the optimal minimax exponent. This is because the effect of the prior washes away in the large deviation setting of $T \to \infty$. But for the Bayesian problem we know the optimal decision rule is always the Maximum-a-Posteriori (MAP) rule. The MAP rule is equivalent to the maximum likelihood rule for a uniform prior over the bandits.      
  \end{remark}
  Next, we have the following consistency result:
\begin{proposition}\label{consistency_result}
\begin{equation}\label{consist_eq}
R_{static}\leq R^{go}_{\infty} \leq R^{go}_{B}\leq R^{go}_1 \leq R^{ub}    
\end{equation} 
\end{proposition}
  The proof is presented in Appendix \ref{proof_prop_2}.
  \begin{remark}
We note that all the results did not need the assumption of finite support or $\xi_a$ being singleton sets. These are general results. The next subsection does make use of the assumption that $\xi_a$ are singleton sets (although this could further be generalized to $\xi_a$ being nonempty finite sets).
  \end{remark}
\subsection{Zero sum differential game value interpretation of $R^{go}_\infty$.}\label{subsec:differential_game_interp}
From here on, we will introduce the symbol $\nu^{i}$ to refer to the unique hypothesis in $\xi_i$. \emph{We will assume that $i \in [m]$, For the best arm identification setting where the number of hypothesis is equal to the number of arms, we have $m=K$. However, we will choose to introduce $m$, since our results extend in straightforward manner to the more general setting where the agent has $K$ actions and is testing against $m$ active hypotheses, with $m \neq K$}.
\\

The goal of this subsection is to interpret $R^{go}_\infty$ as the value of a zero sum differential game. We refer the reader for more precise details and preliminaries from the monograph \cite{souganidis1999two}. 
Consider the following zero sum differential game, where two players are the agent and nature. The first player or \emph{leader} is nature who uses the previous sample history (also the knowledge of agent strategy) to choose the empirical bandit distribution $Q_t$ in the infinitesmial interval $(t,t+dt]$. The agent is the second player or \emph{follower} who chooses an allocation function $w_t$ based on sample history and the choice of the leader $Q_t$ in the same infinitesmial interval. It is to be noted that the leader is aware that the follower can base their strategy on their choice $Q_t$.
\begin{remark}
The above game is in a sense reverse of the actual gameplay that occurs where the role of leader is taken by the agent and that of the follower by nature. We work with the current setup to relate the differential game to the lower bound $R^{go}_{\infty}.$
\end{remark}
The dynamical equation of the differential game and initial state vector are given by:
\begin{equation}\label{dyn_eq}
\begin{aligned}
\dot x_{t,i} &= \sum_{a \in [K]}w_t(a)D(Q_{t,a}||\nu^{i}_a), \\
x_{0,i} &=0,\hspace{0.4cm} \forall i \in [m],
\end{aligned}
\end{equation}
where each $i$ refers to one of the bandit instances in $\xi$. We have assumed here that $\xi_a$ consists of singleton sets (it need not be singleton but just finite). Let us assume that $|\xi|=m$ \footnote{it is true that $m=K$ in the case $\xi_a$ are all singletons but let us keep using $m$ to denote the number of bandit instances under consideration}. Hence $x \in \mathbb{R}^{m}$. We will assume the game is played over the finite unit interval $[0,1]$. Both players aim to control a final terminal cost $g(x_1)$ at $t=1$. The running cost is presumed to be zero. The leader tries to minimize it while the follower tries to maximize it:
\begin{equation}\label{term_cost}
g(x_1)=\underset{j}{\max}\hspace{0.1cm}\underset{j \neq i}{\min} \hspace{0.1cm} x_{1,i}.
\end{equation}
This terminal cost can also be re-written as the second order statistic of $x$, that is $g(x)=x_{(2)}$ and is again interpreted as the negative log-likelihood of bandit making an error under the maximum likelihood decision rule (recall the discussion after eq. \eqref{terminal_cost}).\\

In \cite{souganidis1999two} they assume certain bounded Lipschitzness on the dynamical system and terminal cost. We will now verify those assumptions so that we can utilize the theory presented in \cite{souganidis1999two}. As $w \in \Delta_K$ and by assumption $Q_{t,a}$ belong to compact convex sets and $\nu_a^{i}$ have full supports \footnote{thus making $Q \to D(Q||\nu^{i}_a)$ continuous.} we conclude that:
\begin{equation*}
\sum_{a \in [K]}w_t(a)D(Q_{t,a}||\nu^{i}_a) \leq C_1, 
\end{equation*}
for some appropriate $C_1$. Further, the dynamics directly depends only on the controls $w_t$ and $Q_t$ in \eqref{dyn_eq} and not on $x_t,t$. Thus, the dynamics is Lipschitz in $x,t$ trivially. We conclude that assumption (1.2) in \cite{souganidis1999two} is satisfied for our game. As minimum of finite  number of lipschitz functions is lipschitz we have that $\underset{j \neq i}{\min} \hspace{0.1cm} x_{i}$ is Lipschitz. Similarly, as maximum of finite  number of lipschitz functions is lipschitz we have that $\underset{j}{\max}\hspace{0.1cm}\underset{j \neq i}{\min} \hspace{0.1cm} x_{i}$ is also lipschitz and hence $g(x)$ is lipschitz. As the initial vector $x_0=0$ and the dynamics are bounded we can conclude that $x_t$ always lies within a bounded set. This naturally implies that $g(x_t)$ is bounded as well. Thus, assumption (1.4) of \cite{souganidis1999two} is satisfied as well. The fact that the running cost is zero immediately implies assumption (1.5) of \cite{souganidis1999two}. Now we have by Theorem 7 in \cite{komiyama2022minimax} that:
\begin{equation*}
R^{go}_{\infty}=\underset{B \to \infty}{\lim} R^{go}_B.
\end{equation*}
But under assumptions (1.2),(1.4) and (1.5) of \cite{souganidis1999two} we can apply Theorem 4.4 in \cite{souganidis1999two} and the observation that the RHS in the above equation before the limit $B \to \infty$ is discrete approximation upper value to the differential game we have that:
\begin{theorem}\label{hji_pde_R_go_infty}
Consider the following Hamilton-Jacobi-Isaac (HJI) equation associated with the differential game and its unique viscosity solution $V(x,t)$ :
\begin{equation}\label{HJI_pde}
\begin{aligned}
& \partial_tV+ \underset{Q}{\min}  \underset{w \in \Delta_K}{\max}  \bigg \{ \sum^{m}_{i=1}\sum_{a \in [K]}w_t(a)D(Q_{t,a}||\nu^{i}_a)\partial_{x_i}V \bigg  \} =0\\
& V(x,1)=g(x), \quad  \forall x \in {R^{m}},
\end{aligned}
\end{equation}
which represents the upper value of the differential game, then 
\begin{equation}
R^{go}_{\infty}= V(0,0).
\end{equation}
\end{theorem}
Thus we have given an alternative PDE representation of $R^{go}_\infty$ as terminal upper value function of the differential game. 
\begin{remark}
The assumption that $g(x)$ needs to be bounded can be relaxed when showing that the value function of the differential game satisfies the HJI equation (See Chapter 4 in \cite{krasovskij1988game} or Chapter 9 in \cite{elliott1972existence}). Alternatively, we could have clipped the function $g$ to be between $0$ and $max_{i,w,Q}\sum_a w_a D(Q_a||v^{i}_a)$ and used the $g_{clipped}$ as the terminal value function and applied the results of \cite{souganidis1999two} to derive a HJI equation. In the next subsection we give an equivalent and well known dynamical programming representation of the value function in terms of 'non-anticipating' strategies.
\end{remark}
The Hamiltonian associated with the PDE is 
\begin{equation*}
H^{+}(p)=\underset{Q}{\min}  \underset{w \in \Delta_K}{\max}  \bigg \{\sum^{m}_{i=1}\sum_{a \in [K]}w(a)D(Q_{a}||\nu^{i}_a)p_i \bigg  \}. 
\end{equation*}
The lower hamiltonian $H^{-}(p)$ is:
\begin{equation*}
H^{-}(p)=\underset{w \in \Delta_K}{\max}  \underset{Q}{\min} \bigg \{ \sum^{m}_{i=1}\sum_{a \in [K]}w(a)D(Q_{a}||\nu^{i}_a)p_i \bigg  \}.    
\end{equation*}
Then we have the following result:
\begin{lemma}\label{Hamiltonian_formula_lemma}
The upper and lower Hamiltonians are equal, that is, they satisfy the Isaacs condition. They are expressed by the following formula :
\begin{equation*}\label{Hamiltonian_formula}
H^{+}(p)=H^{-}(p)=
\begin{cases}
    \max_a -(\sum_i p_i) \log \left( \sum_{x \in \mathcal{X}}\prod^{m}_{i=1}(\nu^{i}_a(x))^{\frac{p_i}{(\sum_i p_i)}} \right), & \text{ if } \sum_i p_i>0\\
    \max_a \min_{x \in \mathcal{X}}-\sum_i p_i \log(\nu^{i}_a(x)), & \text { if } \sum_i p_i \leq 0.
\end{cases}
\end{equation*}
Further, under our assumptions $\nu^{i}_a(x) \geq \epsilon$ $\forall x \in \mathcal{X}$, $a \in [K]$ and $i \in [m]$ for some $\epsilon>0$,  we then have that $H^{+}(p)=H^{-}(p)=H(p)$ is Lipschitz continuous with a Lipschitz constant $ \sqrt{m} \log \left( \frac{1}{\epsilon}\right).$ The Hamiltonian is also non-decreasing in each co-ordinate $p_i$.
\end{lemma}
The proof is in Appendix \ref{Hamiltonian_formula_lemma_proof}. Due to Isaac's condition holding, the lower value of the game (where the agent plays first and nature follows) is equal to the upper value (where the order of play ins reversed). This explains why $R^{go}_{\infty}$ is achievable by an algorithm (see also the proof of Theorem 5 in \cite{komiyama2022minimax} for computationally infeasible algorithm achieving $R^{go}_{\infty}$ and hence a direct proof of the next corollary.). We record the result as corollary:
\begin{corollary}\label{achievability_R_go_inty}
The exponent $R^{go}_\infty$ is achievable by some algorithm. 
\end{corollary}
As the upper and lower hamiltonians are equal we will denote them as just $H(p)$ without any further distinction from hereon. 
Next, as a consistency check, we utilize the PDE formulation to show how we can get back the results of \cite{hayashi2009discrimination} and \cite{nitinawarat2013controlled}.  Let us consider the time reversed PDE system of Theorem \ref{hji_pde_R_go_infty}, gotten by using the transform $\tilde{t}=1-t$:
\begin{equation}\label{time_revr_pde}
\begin{aligned}
& \partial_{\tilde{t}}\tilde{V}+ \underset{Q}{\max}  \underset{w \in \Delta_K}{\min}  \bigg \{ \sum^{m}_{i=1}\sum_{a \in [K]}-w_t(a)D(Q_{t,a}||\nu^{i}_a)\partial_{x_i} \tilde{V} \bigg  \} =0\\
& \tilde{V}(x,0)=g(x), \quad  \forall x \in {R^{m}}.
\end{aligned}
\end{equation}
Note that the new hamiltonian for eq \eqref{time_revr_pde} is :
\begin{equation*}
\tilde{H}(p)=\underset{Q}{\max}  \underset{w \in \Delta_K}{\min}  \bigg \{ \sum^{m}_{i=1}\sum_{a \in [K]}-w(a)D(Q_{a}||\nu^{i}_a)p_i \bigg  \}=-H(p).
\end{equation*}
and that:
\begin{equation*}
\tilde{V}(0,1)=V(0,0)=R^{go}_{\infty}.
\end{equation*}
We note that when the dimension $m=2$, the function $g(x)$ simplifies to  be $g(x_1,x_2)=\max \{ x_1,x_2 \}$ which is convex and lipschitz (with a lipschitz constant 1).
Now this means we can apply the Hopf formula (Theorem 3.1 in \cite{bardi1984hopf}) to the time reversed PDE system to obtain:
\begin{equation*}
\tilde{V}(x,\tilde{t})=\underset{p}{\sup} \{ \langle p,x \rangle+\tilde{t}H(p)-g^{*}(p) \}
\end{equation*}
where $g^{*}$ is the fenchel conjugate of $g$. The conjugate $g^{*}$ (for $m=2$) is calculated as:
\begin{equation*}
 g^{*}(p)=\begin{cases}
     0 & \text{ if } p \in \Delta_2\\
     +\infty & \text{ otherwise.}
 \end{cases}
\end{equation*}
As we are interested in $\tilde{V}(0,1)$ we have from Lemma \ref{Hamiltonian_formula_lemma}:
\begin{equation*}
R^{go}_{\infty}=\underset{p \in \Delta_2}{\sup} \max_a -\log \left(\sum_{x \in \mathcal{X}} (\nu^{1}_a(x))^{p_1}(\nu^{2}_a(x))^{p_2} \right)
\end{equation*}
which is equivalent to \cite{hayashi2009discrimination}'s characterization of the minimax exponent in the binary case $(m=2)$.
\begin{remark}
The above example also shows the apparent difficulty between $m=2$ and $m>2$ cases: the terminal condition $g(x)$ is not convex for $m>2$ while it is for $m=2$. The Hopf formula is only valid for convex terminal conditions.
\end{remark}
Next, we combine Hopf formula with a convex upper bound for $g$, by using the comparison principle for viscosity solutions (Proposition 1.1 in \cite{crandal1984two}) to obtain upper bounds on $R^{go}_\infty$, the minimax exponent. We note that $g(x) \leq  g_{i,j}(x)=\max\{ x_i,x_j \}$ for all $i \neq j$. Hence from the comparison principle and Hopf formula we have that:
\begin{equation*}
\begin{aligned}
\tilde{V}(0,1) & \leq \underset{p}{\sup} \{H(p)-g_{i,j}^{*}(p)\}\\
&=\underset{p \in \Delta_{\{i,j\}}}{\sup} \{H(p)\}\\
&=\underset{a}{\max} \hspace{0.2 cm} C(\nu^{i}_a,\nu^{j}_a)
\end{aligned}
\end{equation*}
As this is true for any $(i,j), i \neq j$ we have that:
\begin{equation*}
R^{go}_{\infty}\leq \underset{i \neq j}{\min} \hspace{0.2cm}\underset{a}{\max} \hspace{0.2 cm} C(\nu^{i}_a,\nu^{j}_a),
\end{equation*}
which is the upper bound result from \cite{nitinawarat2013controlled}.

Using similar ideas one can derive the following upper bound for $R^{go}_{\infty}$:
\begin{proposition}\label{alt_R_go_infty_ub}
Denote by the set of all pairs $(i,j)$, $i<j$ by $S$. For each pair $(i,j)$ let $\beta_{i,j} \in \Delta_{m}$ be any probability distribution with support in the two point index set $\{i,j\}$.  Now for any $\lambda \in \Delta_S$ and choice of $\beta=(\beta_{i,j})_{(i,j) \in S}$ we define:
\begin{equation*}
P(\lambda,\beta)= \sum_{(i,j) \in S} \lambda_{i,j}\beta_{i,j}
\end{equation*}
which is the induced distribution $P(.) \in \Delta_m$. Now we have:
\begin{equation*}
R^{go}_{\infty} \leq \underset{\lambda \in \Delta_S}{\inf}\hspace{0.2cm}\underset{\beta}{\sup} \hspace{0.2cm} \underset{a}{\max}-\log \left(\sum_{x \in \mathcal{X}}\prod^{m}_{k=1}(\nu^{k}_a(x))^{P_k(\lambda,\beta)} \right):=R_{Hopf}
\end{equation*}
\end{proposition}
The proof is in appendix \ref{alt_R_go_infty_ub_proof}.
\subsection{A Dynamic Programming Representation of the Value Function $V(x,t)$.}
In this section we will give a Dynamic Programming (DP) representation of $V(x,t)$ in terms of non-anticipating strategies (defined below) first defined in \cite{elliott1972existence}. First let us define set of measurable controls:
\begin{definition}
A measurable control for the agent from time $t$ is a measurable mapping :
\begin{equation*}
w_{[t,1]}: [t,1] \longrightarrow \Delta_K
\end{equation*}
for each $t \in [0,1)$. Similarly a  measurable control for nature from time $t$ is defined as the measurable mapping:
\begin{equation*}
 Q_{[t,1]} : [t,1] \longrightarrow (\Delta_{\mathcal{X}})^{K}. 
\end{equation*}
\end{definition}
Denote the set of such measurable controls as $\mathcal{W}(t)$ and $\mathcal{Q}(t)$ respectively for agent and nature respectively. Next, we define non-anticipative strategies for both nature and agent:
\begin{definition}[Non-anticipative strategy for agent.]\label{def: non_anticipative_agent}
A non-anticipative strategy for the agent beginning from time $t$ is given by the mapping:
\begin{equation}
\alpha : \mathcal{Q}(t) \longrightarrow \mathcal{W}(t)
\end{equation}
such that for each $s \in [t,1]$ and $Q^{0}$,$Q^{1} \in \mathcal{Q}(t)$, if $Q^{0}=Q^{1}$ almost everywhere in $[t,s]$ then we have that $\alpha(w^{0})=\alpha(w^{1})$ almost everywhere in $[t,s]$.
\end{definition}
Denote the set of all non-anticipative strategies for the agent beginning from $t$ as $\mathcal{A}(t)$. Similarly, we can defined non-anticipative strategies for nature:
\begin{definition}[Non-anticipative strategy for nature.]\label{def: non_anticipative_nature}
A non-anticipative strategy for the agent beginning from time $t$ is given by the mapping:
\begin{equation}
\gamma : \mathcal{W}(t) \longrightarrow \mathcal{Q}(t)
\end{equation}
such that for each $s \in [t,1]$ and $w^{0}$,$w^{1} \in \mathcal{W}(t)$, if $w^{0}=w^{1}$ almost everywhere in $[t,s]$ then we have that $\gamma(w^{0})=\gamma(w^{1})$ almost everywhere in $[t,s]$.
\end{definition}
Denote the set of such non-anticipative strategies of nature starting from $t$ as $\mathcal{G}(t)$. We then have that the value function can be formalized as follows:
\begin{lemma}\label{DP_value_formulation}
We have under the Isaacs condition that:
\begin{equation}\label{terminal_value_representation}
\begin{aligned}
V(x,t) & =\underset{w \in \mathcal{A}(t)} {\sup} \underset{Q_{[t,1]} \in \mathcal{Q}(t)}{\inf} g(x(1))\\
&= \underset{Q \in \mathcal{G}(t)}{\inf} \underset{w_{[t,1]} \in \mathcal{W}(t)}{\sup} g(x(1)).
\end{aligned}
\end{equation}
Further, we have the following DP recursion satisfied for any $s >t$ :
\begin{equation}\label{DP_recursion}
\begin{aligned}
V(x,t)&=\underset{w \in \mathcal{A}(t)} {\sup} \underset{Q_{[t,1]} \in \mathcal{Q}(t)}{\inf} V(x_{s},s)\\
&=\underset{Q \in \mathcal{G}(t)}{\inf} \underset{w_{[t,1]} \in \mathcal{W}(t)}{\sup} V(x_{s},s),
\end{aligned}
\end{equation}
where $x_s$ is the value of the solution at time $s$ to eq. \eqref{dyn_eq} in response to measurable control $Q_{[t,1]}$ and non-anticipative strategy $w$ with the initial condition $x_t=x$ in the first equality. A similar interpretation, with roles appropriately reversed, holds in the second equality above.
\end{lemma}
A proof of this standard fact from differential game theory can be found in Theorem 2.2 \cite{souganidis1999two} \footnote{More details about the above construction is available in section 2 of\cite{souganidis1999two} and in greater detail in \cite{elliott1972existence}.} The above lemma interprets the value function $V(x,t)$ as a sort of equilibrium value of the terminal function $g$ resulting from the play of both agent and nature.
\section{Numerical schemes based on PDE formulation \eqref{HJI_pde}.}\label{sec:numerical_pde_scheme}
In this section we propose numerical schemes to compute $R^{g}_{\infty}$ approximately (subsection \ref{FD_scheme}) and an approximately optimal algorithm ( \ref{matching_opt_algo})based on such approximations of the value function. We, utilize heavily previous results from the PDE literature presented in \citet{crandal1984two,souganidis1985approximation,taras1994approximation,tarasyev1999control}.
\subsection{Finite Difference Explicit Upwind Scheme to approximate $R^{go}_{\infty}$.}\label{FD_scheme}
From the previous section, we know that for the PDE eq. \eqref{HJI_pde} the Hamiltonian $H(.)$ is independent of $t,x$, lipschitz and one-homogeneous in $p$. The terminal condition $g(x)$ is Lipschitz and one-homogeneous as well.
\begin{lemma}\label{viscosity_properties}
The viscosity solution to PDE eq. \eqref{HJI_pde} has the following properties:
\begin{enumerate}
    \item The solution $V(x,t)$ is lipschitz jointly in $t$ and $x$ over compact subsets of $\mathbb{R}^{m} \times [0,1]$.
    \item The solution $V(x,t)$ is unique.
    \item The solution satisfies the scaling condition, $\forall t \in [0,1)$: 
    \begin{equation*}
        V(x,t)=(1-t) V\left(0,\frac{x}{1-t} \right).
    \end{equation*}
\end{enumerate}
\end{lemma}
The proof of the lemma is given in Appendix \ref{proof_viscosity_properties}. 
Next, we utilize the fact that the speed of propagation in the dynamics is finite.
\begin{lemma}\label{domain_of_dependence}
From our assumptions we have that :
\begin{equation*}
\underset{w \in \Delta_K}{\max} \underset{\max}{Q}\sum_a w_a D(Q_a)||\nu^{a}_i) \leq \log \left(\frac{1}{\epsilon} \right),
\end{equation*}
for each $i \in [m]$ \footnote{see lemma \ref{Hamiltonian_formula_lemma} for the definition of $\epsilon$.}. That is the velocity field of the state vector $x$ is bounded or finite. In the PDE eq. \eqref{HJI_pde} suppose we replace the terminal condition $g$ with a new function $g^{'}$ such that $g(x)=g^{'}(x)$, for each $x \in [0,\log \left(\frac{1}{\epsilon} \right)]^{m}$. Further assume that $g^{'}$ is bounded, has compact support and is lipschitz and denote the unique viscosity solution to eq. \eqref{HJI_pde} with the terminal condition replaced by $g^{'}$ as $V^{'}$. Then, we have that:
\begin{equation*}
V(x,t)=V^{'}(x,t),
\end{equation*}
whenever $0 \leq x_i \leq \log \left(\frac{1}{\epsilon} \right) t$, for each $i \in [m]$ and $t \in [0,1]$.
\end{lemma}
The proof is given in appendix \ref{proof_domain_of_dependence}. We will compute $R^{go}_{\infty}$ by using a suitable $g^{'}$ instead of $g$. We do this instead of using $g$ directly as the terminal function, because a lot of the standard convergence results for finite difference schemes like \cite{crandal1984two} and \cite{souganidis1985approximation} assume the terminal function is bounded and further it helps keeping the domain of discretization bounded if $g^{'}$ has a compact support.
\begin{lemma}\label{alt_g_properties}
Let $a=\log \left( \frac{1}{\epsilon} \right)$ and $C=[0,a]^{m}$. Let the $l _{\infty}$ expansion of the hypercube $C$ by a radius $a/2$ be denoted as $C_{a/2}$, that is:
\begin{equation*}
C_{a/2}=[-a/2,3a/2]^{m}.    
\end{equation*}
Define the following function:
\begin{equation}\label{modified_terminal_func}
g^{'}(x)=\begin{cases}
    g \left(\prod^{(2)}_C(x) \right) \left(1-\frac{2d_{\infty}(x,C)}{a}\right), & \hspace{0.2cm} x \in C_{a/2}\\
    0 & \hspace{0.2cm} x \notin C_{a/2},
\end{cases}
\end{equation}
where $\prod^{(2)}_C(x)$ is the $l_2$ projection of the point $x$ onto the set $C$ and $d_{\infty}(x,C)$ denotes $l_{\infty}$ distance of the point $x$ from the set $C$. Then $g^{'}(x)$ is a bounded, lipschitz function with a compact support such that $g(x)=g^{'}(x)$ whenever $x \in C$. Further, we have that $V^{'}(x,t)=0$\footnote{recall the definition of $V^{'}$ from lemma \ref{domain_of_dependence}.} whenever $x \notin [-3a/2,3a/2]^{m}$ for all $t \in [0,1]$.
\end{lemma}
The proof of the lemma is given in appendix \ref{proof_alt_g_properties}. The above lemma gives us a $g^{'}$ that has the necessary boundedness and compact support characteristics we sought for. The boundedness allows us to leverage the most straightforward upwind schemes proposed in \cite{crandal1984two}. The function $g^{'}$ has $C_{a/2}$ as the compact support. This implies that the discretization is just a centered hypercube $[-3a/2,3a/2]^{m}$. At the boundary of this hypercube we can simply set the boundary condition to be zero because of the last statement of the lemma.
\begin{remark}
The projection in $g^{'}$ is taken wrt to $l_2$ norm so that there exists a unique projection point. The distance was taken as $d_{\infty}$ because the sets $C,C_{a/2}$ were axis aligned hyper-rectangles and the $d_{\infty}$ distance is easy to compute. For the same reason, it is also true that $\prod^{(2)}_C(x) \subseteq \prod^{(\infty)}_C(x)$ for every $x$, that is the $l_2$ projection point is also a projection point under the $l_{\infty}$ norm.
\end{remark}
We propose the following simple explicit upwind scheme:
\begin{equation}\label{explicit_upwind}
V^{n+1}_{i}=V^{n}_i-\Delta t \tilde{H} \left( \left(\frac{\Delta_j V^{n}_{i}}{\Delta h} \right)_{j \in [m]} \right),
\end{equation}
where $V^{n}_{i}$ refers to the approximate value function at time $n \Delta t$ at grid point $i$. $\Delta_j V^{n}_{i}=V^{n}_{(i_1,i_2, \dots,i_{j+1}, \dots,i_m)}-V^{n}_i$, that is difference of approximated value functions at the forward point $(i_1,\dots, i_{j+1}, \dots, i_{m})$ from $i=(i_1,\dots,i_{j}, \dots,i_m)$. $\Delta t$ and $\Delta h$ are time and space discretizations respectively. Note, that we use the same spatial discretization in all directions for simplicity. $\tilde{H}$ is the hamiltonian of time reversed PDE eq. \eqref{time_revr_pde}. Each spatial grid point $i$ corresponds to spatial point $(i_1\Delta h, i_2 \Delta h, \dots, i_m \Delta h)$. The indices are in the range $\lceil \frac{-3a}{2 \Delta h} \rceil \leq i_j \leq \lfloor \frac{3a}{2\Delta h} \rfloor$. For any evaluation of the approximate value function at a spatial point outside this range we set the evaluation to be zero (this is justified from the results of lemma \ref{alt_g_properties}). The eq. \eqref{explicit_upwind} gives us a way to compute the approximate value function in the next time step from the evaluations of the current time step. We set the initial condition as follows:
\begin{equation}\label{init_cond_upwind}
V^{0}_i=g^{'}((i_1\Delta h, i_2 \Delta h, \dots, i_m \Delta h)).
\end{equation}
\begin{remark}
We have setup the finite difference scheme eq \eqref{explicit_upwind} as stepping forward in time. The original differential game had a terminal value function and would intuitively suggest to step backwards in time. The equivalence between these two setups is trivial to verify. We use the forward in time setup to be consistent with the setup used in \cite{crandal1984two}.
\end{remark}
In \cite{crandal1984two} they introduce the notion of a \emph{consistent} and \emph{monotone} finite difference scheme \footnote{See section 1 and 2 in \cite{crandal1984two} for more details.}. In case of the upwind scheme proposed in eq. \eqref{explicit_upwind} consistency is trivially true since we only use one forward difference $\frac{\Delta_j V^{n}_{i}}{\Delta h}$ per direction. The notion of monotone scheme means that the RHS of eq. \eqref{explicit_upwind} is non-decreasing in $V^{n}_i$ for each $i$ appearing on the RHS. We then have:
\begin{lemma}\label{explicit_scheme_mono}
The explicit upwind scheme proposed in eq. \eqref{explicit_upwind} is a monotone scheme, provided $\Delta h$ and $\Delta t$ are chosen to satisfy the following stability criteria:
\begin{equation}\label{stability_criteria}
1-\frac{m \Delta t}{\Delta h} \log \left( \frac{1}{\epsilon}\right) \geq 0.
\end{equation}
\end{lemma}
The proof is given in appendix \ref{proof_explicit_scheme_mono}. The stability criteria eq. \eqref{stability_criteria} is a Courant-Friedrichs-Lewy (CFL) type condition, which guarantees the stability of upwind schemes like eq. \eqref{explicit_upwind}. This is a well known condition in numerical solution of such hyperbolic PDEs \cite{courant1967partial}.\\

Given that the explicit upwind scheme eq. \eqref{explicit_upwind} has been shown to be consistent and monotone under the CFL stability condition, we can utilize Theorem 1 of  \cite{crandal1984two} to provide the following estimate:
\begin{theorem}[Theorem 1 \cite{crandal1984two}]\label{convergence_thm_fd_scheme}
 The explicit upwind scheme eq. \eqref{explicit_upwind}, assuming the step sizes $\Delta t$ and $\Delta h$ satisfy the CFL stability criteria eq. \eqref{stability_criteria}, has the following error estimate:
 \begin{equation}\label{upwind_err_estimate}
 |V^{n}_i- V^{'}((i_1\Delta h, i_2 \Delta h, \dots, i_m \Delta h), n \Delta t)| \leq c \sqrt{\Delta t},  
 \end{equation}
 where $c$ is a constant dependent only on $\sup_{x} |g^{'}(x)|$ and the lipschitz constants of $g^{'}$ and $\tilde{H}$. Here $V^{'}(,,.)$ is the evaluation of the actual viscosity solution while $V^{n}_i$ is the approximated value function evaluated at the corresponding grid point.
\end{theorem}
This theorem states that if we hold the ratio $\kappa$ of time step by the spatial step constant, and send the time step $\Delta t \to 0$, we get finer and finer approximation of the value function. One can thus use these finite difference schemes to numerically compute the optimal exponent $R^{go}_{\infty}$. A pseudocode for implementing the procedure described above is given below:
\begin{algorithm}
\caption{Explicit Upwind Finite Difference Scheme.}
\label{alg:upwind_nd}
\begin{algorithmic}[1] % Line numbering enabled

\State \textbf{Input:} % Inputs
\State $m$, $a$, $T$,$N_x$,$N_t$.
\State Initial  function \(g^{'}(\mathbf{x})\).

\State \textbf{Output:} % Output
\State m-dimensional solution array \(V\) approximating \(V^{'}\) at final time \(T\).

\vspace{0.5em}
\State \textbf{Initialization:}
\State \( \Delta h = 3a / N_x \), \( \Delta t = T / N_t \).
\State  \textbf{if} \(1 < \frac{m \Delta t}{\Delta h} \log \left( \frac{1}{\epsilon}\right) \) \textbf{then}
\State \quad Warn of violation of CFL condition.
\State  \textbf{end if}

\State Define grid points: \(\mathbf{x}_{\bm{i}} = (-3a/2 + i_1 \Delta x, \dots, -3a/2 + i_m \Delta x)\) for multi-index \(\bm{i} = (i_1, \dots, i_m)\) where \( 0 \le i_k \le N_x \) for all \( k=1,\dots,m \).
\State Initialize m-dimensional solution arrays \(V_{\text{current}}\) and \(V_{\text{next}}\) of size \((N_x+1) \times \dots \times (N_x+1)\).
\State \textbf{for} each \(\bm{i}\)  \textbf{do}
    \State \quad \textbf{if} \(\bm{i}\) is on the boundary (i.e., any \(i_k=0\) or \(i_k=N_x\)) \textbf{then}
        \State \quad \quad \( V_{\text{current}}[\bm{i}] \gets 0 \) 
        \State \quad \quad \( V_{\text{next}}[\bm{i}] \gets 0 \)    \Comment{Boundary values set to zero}
    \State \quad \textbf{else} 
        \State \quad \quad \( V_{\text{current}}[\bm{i}] \gets g^{'}(\mathbf{x}_{\bm{i}}) \) \Comment{Apply initial condition}
    \State \quad \textbf{end if}
\State \textbf{end for}

\vspace{0.5em}
\State \textbf{Time Stepping Loop:}
\State \textbf{for} \( n = 0 \) \textbf{to} \( N_t - 1 \) \textbf{do} 
    \State \quad \textbf{for} each \(\bm{i}\) \textbf{do}
        \State \quad \quad Initialize 1-D array $Grad$ of size $m$.
        \State \quad \quad \textbf{for} \( k = 1 \) \textbf{to} \( m \) \textbf{do} \Comment{Compute FD approx in each direction.}
            \State \quad \quad \quad \(Grad[k] = \frac{V_{current}[i_k]-V_current[i]}{\Delta h}\).
        \State \quad \quad \textbf{end for} 
        \State \quad \quad \( V_{\text{next}}[\bm{i}] \gets V_{\text{current}}[\bm{i}] - \Delta t H(Grad) \) \Comment{Upwind Scheme.}
    \State \quad \textbf{end for} 

    \vspace{0.5em}
    \State \quad \( V_{\text{current}} \gets V_{\text{next}} \) 
\State \textbf{end for}

\vspace{0.5em}

\State \Return \( V_{\text{current}} \) 

\end{algorithmic}
\end{algorithm}
\subsubsection{Computational effort in calculating $R^{go}_{\infty}$.}
In \cite{komiyama2022minimax} had given the representation that $R^{go}_{\infty}=\lim_{B \to \infty} R^{go}_{B}$. Calculating $R^{go}_{\infty}$ numerically from this representation is quite difficult. One has to solve a repeated stackelberg game (see eq. \eqref{eq_rep_R_go_B}) for a large enough $B$. This entails solving $B$- times nested minimax problems, which themselves can be quite challenging \footnote{This had led \cite{komiyama2022minimax} to term their algorithm Delayed Optimal Tracking (DOT) as just of theoretical interest and computationally intractable.}. In general we will require $O(me^{B K })$. See more details in appendix \ref{compute_effort_R_go_infty}. Instead the PDE approach is suitable for this whenever $m$ is small, typically $m \leq 4$. The Explicit Upwind Finite Difference eq. \eqref{explicit_upwind} is computationally tractable for small $m$, but suffers from the curse of dimensionality for moderate to large $m$. It typically require computational effort of $O(K B^{m+1})$ (with $B=\frac{1}{\Delta t}$). See again, appendix \ref{compute_effort_R_go_infty} for the details. In contrast however, using eq. \eqref{eq_rep_R_go_B} is intractable even for such small $m$. Thus, the PDE formulation eq. \eqref{HJI_pde} can be useful computationally when $m$ is small.

\subsubsection{Numerical calculations of $R^{go}_{\infty}$ and a conjecture of \cite{komiyama2022minimax}.}\label{subsec: num_compute_conj_disprov}
Using the Explicit Upwind Scheme \eqref{explicit_upwind} one can compute the $R^{go}_{\infty}$ numerically. We contrast this with the situation in \cite{komiyama2022minimax}. There the formulation as a limit of $R^{go}_{B}$ \eqref{lim_defn_r_go_infty} is computationally intractable. Further, \cite{komiyama2022minimax} had conjectured if $R^{go}_{1}$ was achievable (see footnote 10 in page 4 of \cite{komiyama2022minimax}.). We show in the following example that $R^{go}_{\infty}< R^{go}_1$ and hence not achievable.

\begin{table}[h]
\centering
\begin{tabular}{|c|c|}
\hline
Quantity &Value \\
\hline
K & 3 \\
\hline
m & 3 \\
\hline
$\nu^{(1)}$ & $ (0.6, 0.3, 0.23)$\\
$\nu^{(2)}$ & $ (0.35, 0.8, 0.35)$\\
$\nu^{(3)}$ & $ (0.2, 0.3, 0.75)$\\
\hline
$R^{go}_{\infty}$ & $0.1234 \pm 2 \times 10^{-4}$  \\
\hline
$R^{go}_{1}$ & $0.13205 \pm 1 \times 10^{-5}$\\
\hline
$R_{static}$ & $0.07814 \pm 3 \times 10^{-5}$\\
\hline
\end{tabular}
\caption{A simple bandit class $\xi=  \{ \nu^{(1)}, \nu^{(2)}, \nu^{(3)} \}$, with three $(m=3)$ three-armed bernoulli bandits ($K=3$).}
\label{tab:num_example_1}
\end{table}
The simple bandit class $\xi$ shown in Table \ref{tab:num_example_1} contains three bernoulli bandits $\nu^{(1)}, \nu^{(2)}, \nu^{(3)}$. Each of these bernoulli bandit has three arms whose parameters are as shown in Table \ref{tab:num_example_1}. This example shows that their conjecture is false in the simple setting we consider \footnote{They conjectured this in the more general bandit setting where $\xi_a$ can be a continuum.}. Given our repeated Stackelberg game formulation of $R^{go}_{B}$ (\eqref{eqv_rep_R_go_B_eq}) and our interpretation of $R^{go}_{\infty}$ as the value of a differential game \eqref{HJI_pde} this seems natural. Indeed, most simple examples have the property $R^{go}_{\infty} < R^{go}_{1}$. 
\subsection{An optimal algorithm based on the PDE formulation \eqref{HJI_pde}.}\label{matching_opt_algo}
In this section we will describe an optimal algorithm. There are many works, for example  \citet{crandal1984two,souganidis1985approximation,barles1991convergence}, which show convergence of various finite difference schemes to value function/viscosity solution of PDEs like eq. \eqref{HJI_pde}, but question of deriving approximate optimal feedback controls in differential games is still very much an open question (see \cite{falcone2006numerical,falcone2013semi}). In our specific problem, we construct and prove a $\delta$-optimality of an algorithm based on an alternative approximation scheme (different from explicit scheme \eqref{explicit_upwind}) for the value function. This approach is based on the works \cite{taras1994approximation,tarasyev1999control}.\\

In the problems where the value function $V$ is known to be continuously differentiable everywhere it can be shown that the optimal strategy for the agent is to play :
\begin{equation}\label{typical_optimal_strat}
 a_t \in \underset{a \in [K]}{argmax} \hspace{0.2cm} H_a \left( \nabla_x V(t,x_t) \right),
\end{equation}
that is play the action maximizing the Hamiltonian where the momentum parameter $p$ is set to the gradient $\nabla_x V(t,x_t)$. See Theorem 4.1.1 in \cite{krasovskij1988game} for a rigorous proof of a similar general  assertion. The issue is that (this is well-known in the PDE literature, see Chapter 3 \cite{evans2022partial}.) that very often Hamilton Jacobi equations have no classical solution where the gradient $\nabla_x V(t,x)$ is continuous. In fact, regardless of how smooth the Hamiltonian and the terminal conditions are, a discontinuity or "shock" can develop in the gradient.\\

In this subsection, we will first develop an alternative scheme that converges. This alternative scheme is based on replacing the approximate value function with its local convex hull. If this replacement is carefully chosen, one can use the sub-differential of local convex hull in eq.\eqref{typical_optimal_strat} instead of the actual gradient $\nabla_x V(t,x)$. Then, our strategy will be to choose an appropriate sub-differential and play the maximization action in \eqref{typical_optimal_strat}, wherein we replace the gradient with the chosen sub-differential. In this way we preserve the intuitive idea of maximizing the Hamiltonian, but ameliorate aforementioned gradient discontinuity issue. 

\subsubsection{Spatio-temporal grid construction.}
Let us first discretize the time interval $[0,1)$ into $B$ disjoint sub-intervals, each with length $\Delta t$. Here, $1=B \Delta t$. Denote $t_l=l \Delta t$. The temporal grid is denoted then as:
\begin{equation*}
GR_{\Delta t}= \{ t_l \hspace{0.2cm} | \hspace{0.2cm} 0 \leq t_l=l \Delta t \leq 1 \text{ and } 0 \leq l \leq B\}.
\end{equation*}
Let us now define the symmetric hypercube:
\begin{equation*}
H_{\sqrt{m}a+\kappa}=\{ x \hspace{0.2cm} | \hspace{0.2cm} \forall i \in [m], -(\sqrt{m}a+\kappa) \leq x_i \leq (\sqrt{m}a+\kappa) \},
\end{equation*}
where as in susbsection \ref{FD_scheme}, Lemma \ref{alt_g_properties}, $a=\log \left( \frac{1}{\epsilon} \right)$. This is the background domain we will work with. Let us discretize $H_{\sqrt{m}a+\kappa}$ into a grid with spatial grid parameter $\Delta h$ (just as in sub-section \ref{FD_scheme}) which is symmetric about origin. Let us denote the set of spatial grid points as:
\begin{equation*}
GR_{\Delta h}=\{ x \hspace{0.2cm} | \hspace{0.2cm}  ||x ||_{\infty} \leq \sqrt{m}a+\kappa \text{ and } \forall i \in [m], x_i=n_i \Delta h , \text{ with } n_i \in \mathbb{Z}\}.
\end{equation*}
We will denote an element $x^{(n)}$ according to the multi-index $n=(n_1,n_2,\dots,n_m)$ with the condition:
\begin{equation*}
x^{(n)}_i=n_i \Delta h
\end{equation*}
holding for each $i \in [m]$. We will restrict our computation of the approximate value function during each $t_l$ to a time-varying subset of the $GR_{\Delta h}$. The spatio-temporal cone is:
\begin{equation*}
G_{\sqrt{m}a,\kappa}= \{ (t,x) \hspace{0.2cm} | \hspace{0.2cm} ||x|| \leq \sqrt{m} a t+\kappa t^2, \hspace{0.2cm} \forall t \in [0,1] \}.
\end{equation*}
The cone $G_{\sqrt{m}a,\kappa}$ is all the space-time points $(t,x)$, that can be affected by the dynamics starting from the origin. The spatial section of $G_{\sqrt{m}a,\kappa}$ at a given time $t$ is defined as:
\begin{equation*}
G^{t}_{\sqrt{m}a,\kappa}= \{ x \hspace{0.2cm} | \hspace{0.2cm} (t,x) \in G_{\sqrt{m}a, \kappa} \}.
\end{equation*}
The aforementioned subset is basically the intersection of $GR_{\Delta h}$ and a spatial section at time $t_l \in GR_{\Delta t}$ of the spatio-temporal cone:
\begin{equation*}
D_l= GR_{\Delta h} \cap G^{t_l}_{\sqrt{m}a, \kappa}, \text{ with } t_l \in GR_{\Delta t}.
\end{equation*}
\begin{remark}
We have chosen to use a spherically symmetric $G_{\sqrt{m} a, \kappa}$. It would have been more efficient and faithful to the dynamics \eqref{dyn_eq} to have used the set $\{ (t,x) \hspace{0.2cm} | \hspace{0.2cm} 0 \leq x_i \leq a t, \forall t \in [0,1], \forall i \in [m] \}$ instead. We have chosen $G_{\sqrt{m} a,\kappa}$ because this follows the setup in \cite{taras1994approximation,tarasyev1999control} and allows us to leverage their techniques directly. The gain in efficiency by using a more faithful to dynamics set, would only be in constant factors. The parameter $\kappa$ is chosen to enable us to derive certain guarantees that the scheme is consistent (see Lemma \ref{lemma_grid_scheme_consistency} below). It can be viewed as the positive ``acceleration" added to the maximum speed of dynamics \eqref{dyn_eq}. 
\end{remark}
We will also need to take simplicial decomposition of $D^{conv}_l$. This decomposition will be used to linearly interpolate the approximate value function to points within $G^{t}_{\sqrt{m} a}$ at any time $t \in [0,1]$. Well known triangulation or simplicial decomposition procedures like Freudenthal triangulation (Section 8.5 \cite{kochenderfer2022algorithms}) can be utilized for this.
\subsubsection{Backward induction and the time stepping operator $F$.}\label{subsec: backward_time_stp_op_alg}
Given the modified terminal function $g^{'}$ at $t=1$, we need to define an operator that backward inducts the approximate value function to $t=0$ recursively along $GR_{\Delta t}$. 
\begin{definition}\label{defn:time_step_operator}
A time stepping operator is a map, formally defined as:
\begin{equation}\label{generic_time_step_operator}
F(l,u_{l+1},GR_{\Delta h}): D_l \longrightarrow \mathbb{R}.
\end{equation}
Here $u_{l+1}$ is a lipschitz function on $D^{conv}_{l+1}$, the closed convex hull of $D_l$. 
\end{definition}
The operator takes a lipschitz function $u_{l+1}$ and modifies it into a function over the pertinent subset $D_l$ in the previous time step $t_l$. The output will be considered as the approximate value function at time $t_l$. We will use the following specific operator:
\begin{equation}\label{lce_time_stp_oper}
F_{lce}(l,u_{l+1},GR_{\Delta h})(x)=\underset{y \in B_x(\sqrt{m}a\Delta t)}{\sup} \hspace{0.2cm} \underset{s \in \partial g_{l+1}(y)}{\max} \{ \Delta t H(s)+ g_{l+1}(y)+\langle s , x -y \rangle \}
\end{equation}
where $g_{l+1}(.)$ is the lower convex envelope (l.c.e) of $u_{l+1}$ over $B_x(\sqrt{m}a\Delta t)$, the ball centered at $x$ and of radius $\sqrt{m} a \Delta t$. $\partial g_{l+1}(y)$ is the sub-differential set of $g_{l+1}$ at $y$ (see the formal definition below). The subscript $lce$ in $F_{lce}(l,u_{l+1},GR_{\Delta h})$ reminds us  that this operator is based on lower convex envelope of $u_{l+1}$. The l.c.e $g_{l+1}$ is obtained by the following variational representation:
\begin{equation}\label{var_rep_lce}
g_{l+1}(y)=\inf \bigg \{ \sum^{p}_{j=1} \alpha_j u_{l+1}(y_j) \hspace{0.2cm}| \hspace{0.2cm} \sum^{p}_{j=1} \alpha_j=1, \text{ and } y=\sum^{p}_j \alpha_j y_j, \text{ with } y_j \in B_x(\sqrt{m}a\Delta t), \alpha_j \geq 0, \forall j \in [p] \text { and } p \in \mathbb{N}. \bigg \}
\end{equation}
for any $y \in B_x(\sqrt{m}a\Delta t)$. The notion of local sub-differential set is defined as :
\begin{equation*}
\partial g_{l+1}(y)= \{ s \hspace{0.2cm}| \hspace{0.2cm} g_{l+1}(z) \geq g_{l+1}(y)+\langle s, z-y \rangle, \forall z \in B_x(\sqrt{m}a\Delta t) \}.
\end{equation*}
\begin{remark}
As mentioned previously, the motivation for eq. \eqref{lce_time_stp_oper} is to allow us to approximate the gradient $\nabla_x V(t,x)$ by the sub-differential of convex envelope of the approximate value function. There are additional motivations arising from using Dini derivatives to define minimiax solutions to eq. \eqref{HJI_pde}. See the introduction in \cite{taras1994approximation}.
\end{remark}
The values obtained by using the operator $F_{lce}(l,u_{l+1},GR_{\Delta h})$ is only on the finite grid $D_l$.  To continue the backward induction over $l$, we need to linearly interpolate to get the values over $D^{conv}_l$. Now for any point $y \in D^{conv}_l$, we have that it must be in one of the hypercubes $H^{x}_{\Delta h}=[x_1,x_1+\Delta h] \times \dots \times [x_m,x_m+\Delta h]$, with $x \in GR_{\Delta h}$. We take the unique triangle in the simplicial decomposition $\Omega$ of this hypercube, which contains $y$. We define the value of $y$ as:
\begin{equation*}
F^{int}_{lce}(l, u_{l+1}, GR_{\Delta h})(y)= \sum^{m}_{j=0}\rho_j(\Omega)F_{lce}(l, u_{l+1}, GR_{\Delta h})(y_j)
\end{equation*}
where $y_j$ are vertices of the aforementioned unique triangle and $\rho_j$ are the barycentric co-ordinates of $y$ in this triangle. We note that this linear interpolant satisfies the consistency condition : $F^{int}_{lce}(l, u_{l+1}, GR_{\Delta h})(y)=F_{lce}(l,u_{l+1},GR_{\Delta h})(y)$, whenever $y \in D_l$. Finally, we define the backward induction process:
\begin{equation}\label{backward_ind_lce_timestp}
\begin{aligned}
& \hat{V}_N(x)=g^{'}(x), \hspace{0.2cm} \forall x \in D^{conv}_B\\
& \hat{V}_l(x)=F^{int}_{lce}(l, \hat{V}_{l+1}, GR_{\Delta h})(x),\ \hspace{0.2cm} \forall x \in D^{conv}_{l},  \forall l=0,1,\dots,(B-1).
\end{aligned}
\end{equation}
We note that $\hat{V}_{l+1}$ is defined only within the region $D^{conv}_{l+1}$, while the computation of $F$ (eq. \eqref{lce_time_stp_oper}) requires defining $\hat{V}_{l+1}$ over the set $\cup_{x \in D_l}B_x(\sqrt{m}a \Delta t)$. The following lemma assures us that this inconsistency will not occur provided $\Delta h$ is chosen appropriately:
\begin{lemma}\label{lemma_grid_scheme_consistency}
If $\Delta h < \frac{\kappa (\Delta t)^2}{\sqrt{m}}$, then we have that :
\begin{equation*}
\underset{x \in D_l}{\cup} B_x(\sqrt{m}a \Delta t) \subseteq D^{conv}_{l+1}, \hspace{0.2cm} \forall l \in  \{0,1,\dots,B-1\}.
\end{equation*}
\end{lemma}
The proof is given in Appendix \ref{proof_lemma_grid_scheme_consistency}.
The function $\hat{V}_l$ (for all $l=0,1,\dots,B$) is a lipschitz function since it is piecewise linear function. Further, its lipschitz constant is bounded by $c(m) L_{g^{'}}$, where $c(m)$ is just a dimension dependent constant since the simplicial decomposition of $D^{conv}_l$ can be made using a simplicial decomposition of the hypercubes intersecting $D^{conv}_l$. These simplicial decompositions (Freudenthal decompoisition) will maintain a mesh quality of the simplices that is dependent only on the dimension $m$. To get the exact dependence $c^{'}(m)$ one can combine Theorem 3 in \cite{taras1994approximation} with property $(F7)$ of Theorem 1 in \cite{taras1994approximation}. We now adapt Theorem 4 in \cite{taras1994approximation} to our setting:
\begin{theorem}[Theorem 4 in \cite{taras1994approximation}]\label{convg_thm_backwrd_stp_op}
The backward induction process eq. \eqref{backward_ind_lce_timestp} is convergent when $\Delta h =O(\Delta t)$ and satisfies an error estimate of 
\begin{equation}\label{backwd_stp_est}
\underset{(t,x): t=t_l \in GR_{\Delta t}, x \in D_l}{\sup}|\hat{V}_l(t,x)-V(t,x)| \leq C \sqrt{\Delta t}.
\end{equation}
Here, $C$ is constant that depends on the Lipschitz constants of $H, g^{'}$ and the constant $a$. 
\end{theorem}
Thus the backward induction scheme eq \eqref{backward_ind_lce_timestp} is convergent just like the simple explicit upwind scheme eq. \eqref{explicit_upwind} in subsection \ref{FD_scheme}. The pseudocode for Backward time-stepping algorithm \eqref{lce_time_stp_oper}, \eqref{backward_ind_lce_timestp} is given as Algorithm \ref{alg: backward_timestepping}.
\begin{algorithm}
\caption{Backward Time-stepping.}
\label{alg: backward_timestepping}
\begin{algorithmic}[1]
\Statex \textbf{Inputs:}
\Statex $\Delta t, \Delta h, a,H(.)$. 
\Statex Grids -$D_B,\dots, D_1,D_0$, Terminal function- $g^{'}(x)$

\Statex \textbf{Output:}
\Statex Multidimensional array $\mat{V}$ storing approximate solution values. $\mat{V}$ has dimensions $(B+1) \times |D_N|$.
\Statex Multidimensional array $\mat{A}$ storing optimal arm pulls. $\mat{A}$ has dimensions $N \times |D_B|$.
\Statex
\Statex \textbf{Initialization:}
\State Initialize the array $\mat{V}$ of size $(B+1) \times |D_B|$  with a special 'invalid' value.

\State Initialize the array $\mat{A}$ of size $B \times |D_B|$  with a special 'invalid' value.
\For{$y_j \in D_B$} 
    \State $\mat{V}[B, j] \gets g^{'}(y_j)$ \Comment{Assign terminal cost}
\EndFor
\For{$l$ from $B-1$ down to $0$} \Comment{Backward time stepping.}
    \State $S_{l+1} \gets$ Simplicial decomposition of $D^{conv}_{l+1}$. \Comment{Triangulation of domain.}
    \State Get $\hat{V}_{l+1}(\cdot) $ by linearly interpolating on values $\mat{V}[l+1, \cdot]$ based on $S_{l+1}$. 
    \For{$y_j \in D_l$} 
        \State $g_{l+1}(.) \gets$ Local convex envelope around $y_j$ of $\hat{V}_{l+1}$.
        \State $\mat{V}[l, j]  \gets \max_a \sup_{\mat{y} \in O(\mat{y}_j, \sqrt{m} a \Delta t)} \max_{\mat{s} \in Df(\mat{y})} \left\{ \Delta t H_a(\mat{s}) - g^{*}_{l+1}(\mat{y}) + \langle \mat{s}, \mat{y}_j \rangle \right\}$.
        \State $\mat{A}[l, j]  \gets \underset{a \in [K]}{argmax} \sup_{\mat{y} \in O(\mat{y}_j, \sqrt{m} a \Delta t)} \max_{\mat{s} \in Df(\mat{y})} \left\{ \Delta t H_a(\mat{s}) - g^{*}_{l+1}(\mat{y}) + \langle \mat{s}, \mat{y}_j \rangle \right\}$. 
        \State \Comment{Based on the reformulation presented in eq. \eqref{alt_rep_tim_stp_op}.}
    \EndFor
\EndFor
\Statex
\State \Return $\mat{V}, \mat{A}$
\end{algorithmic}
\end{algorithm}
The computational complexity of calculating the time stepping operator used in Algorithm \ref{alg: backward_timestepping} is  discussed in \ref{compute_time_time_stpp_op}. In particular we utilize the representation eq. \eqref{alt_rep_tim_stp_op} of the time stepping operator developed there in Algorithm \ref{alg: backward_timestepping} (see lines 11-13 in Algorithm \ref{alg: backward_timestepping}).
\subsubsection{Approximately Optimal Algorithm.}
We now utilize the backward timestepping process (eq.\eqref{backward_ind_lce_timestp} and Algorithm \ref{alg: backward_timestepping}) defined in the previous subsection \ref{subsec: backward_time_stp_op_alg} in building an approximately optimal algorithm for ASHT problem. We first run the backward time stepping algorithm \ref{alg: backward_timestepping} and obtain the optimal arm pull array $\mat{A}$. Next, we break the samples into uniform batches, with the number of batches $B$ being determined by $B=\frac{1}{\Delta t}$. The agent decides their action for the next batch of $T/B$ samples at time intervals $t_l \in GR_{\Delta t}$. As the speed of the dynamics \eqref{dyn_eq} is upper bounded by $\sqrt{m}a$, we note that at any time $t \in [0,1]$, for any trajectory starting at the origin, that  is $x_0=0$, we have:
\begin{equation*}
x_t \in G^{t}_{\sqrt{m}a,\kappa}.
\end{equation*}
For any $t_l \in GR_{\Delta t}$, we note that $D_l$ is the spatial discretization of $G^{t_l}_{\sqrt{m}a,\kappa}$. The algorithm will first project the state vector $x_{t_l}$ onto the grid $D_l$, that is find a $y_{j_l} \in D_l$ such that:
\begin{equation}\label{proj_x_l_grid}
y_{j_l} \in \underset{y \in D_l}{argmin}||x_{t_l}-y||.
\end{equation}
The optimal action for $y_l$ was recorded in the optimal arm pull array $A$. The algorithm will play the optimal action $A[l,j_l]$ for the next $T/B$ samples, that is:
\begin{equation}\label{optimal_action_policy}
 \mat{A}[l, j_l]  \gets \underset{a \in [K]}{argmax} \sup_{\mat{y} \in O(\mat{y}_{j_l}, \sqrt{m} a \Delta t)} \max_{\mat{s} \in Df(\mat{y})} \left\{ \Delta t H_a(\mat{s}) - g^{*}_{l+1}(\mat{y}) + \langle \mat{s}, \mat{y}_{j_l} \rangle \right\},   
\end{equation}
where $j_l$ is grid index of the point $y_{j_l}$. The above Algorithm \ref{alg: GOAP} is called Grid-based Optimal Action Policy (GOAP).
\begin{algorithm}
\caption{Grid-based Optimal Action Policy (GOAP).}
\label{alg: GOAP}
\begin{algorithmic}[1]
\Statex \textbf{Inputs:}
\Statex Unknown bandit $\nu \in \xi$, No. of Samples $T$, Bandit class $\xi$.
\Statex $\Delta t, \Delta h, a,H(.)$. 
\Statex Grids -$D_B,\dots, D_1,D_0$, Terminal function- $g^{'}(x)$

\Statex \textbf{Initialization:}
\State $B \gets \lfloor \frac{1}{\Delta t} \rfloor$.
\State Run  Backward Timestepping (Algorithm \ref{alg: backward_timestepping}) with the given inputs and obtain $\mat{V},\mat{A}$.
\State Initialize $\mat{x}$ as all-zero $m$- dimensional vector.\Comment{State vector initialized.}\\
\For{$l$ from $B-1$ down to $0$} 
    \State $y_{j_l} \in \underset{y \in D_l}{argmin}||x-y||$. 
    \State $a_l \gets A[l,j_{l}]$. \Comment{$j_l$ is the index of $y_{j_l}$ in $GR_{\Delta h}$.}
    \State Sample arm $a_l$ of the unknown bandit $\nu$ exactly $T/B$ times and observe the empirical distribution $Q_l$.
    \State $\mat{x} \gets \mat{x}+\Delta t \left(D(Q_{l,a_l}|| \nu^{(i)}_{a_l} )\right)_{i \in [m]}$ . \Comment{Update state vector.}
\EndFor
\Statex
\State \Return $\underset{i \in [m]}{argmin} \hspace{0.2cm} \mat{x}_i$. 
\end{algorithmic}
\end{algorithm}
The proposed algorithm is an example of a "meta" strategy defined in section \ref{sec:approachability}. We provide the following guarantee for GOAP (Algorithm \ref{alg: GOAP}):

\begin{theorem}\label{goap_ldp_guarantee}
If in GOAP, we choose $\Delta h < \frac{\kappa (\Delta t)^2}{\sqrt{m}}$, with $\kappa$ set to a constant, we have that:
\begin{equation}\label{goap_guarantee_eq}
e_{m}(\xi,GOAP) \geq R^{go}_{\infty}-C \sqrt{\Delta t}
\end{equation}
that is, the minimax exponent $e_{m}(\xi,GOAP)$ can be made arbitrarily close to $R^{go}_{\infty}$ if we choose $\Delta t$ small enough. The constant $C$ depends on $m,a$ and lipschitz constants of $g^{'}$ and $H$.
\end{theorem}
The proof of the theorem is in Appendix \ref{proof_goap_ldp_guarantee}. The proof first shows that any optimal response of nature to GOAP play in the differential game is piecewise constant in the intervals $[t,t+\Delta t)$. Thus, we study only piecewise constant paths in the proof. We show using the structure of backward timestepping operator eq. \eqref{optimal_action_policy} a certain recursive bound on $\hat{V}_{l}$. This enables us to prove a pathwise bound required by Lemma \ref{approach_relate_err_exp}. Finally, result follows by an application of Lemma \ref{approach_relate_err_exp} since GOAP is a "meta" strategy in the sense defined in section \ref{sec:approachability}.
\subsubsection{Computational effort in GOAP (Algorithm \ref{alg: GOAP}).}
Since the loop in lines 5-10 of Algorithm \ref{alg: GOAP} takes only $O(m)$ per iteration the total computational cost of the loop is just $O(mB)$. Thus, clearly the computational effort in Algorithm \ref{alg: GOAP} is dominated by the cost of running Algorithm \ref{alg: backward_timestepping} and computing the array $\mat{A}$. From the discussion in Appendix \ref{compute_time_time_stpp_op} it is clear evaluating the time stepping operator at a point $x \in D_l$ takes $O(K B^{m^2}$. Now since there are $O(B^{m}l^{m})$ grid points in each $D_l$ we note that we will have to evaluate the time stepping operator at $O(B^{2m})$ points. Consequently, we need $O(KB^{m^2+2m})$ computational effort to compute $\mat{A}$. This is polynomial in $B$ as opposed to DOT algorithm proposed in \cite{komiyama2022minimax}. We note that the polynomial degree has become quadratic in $m$ as compared to the effort to just compute the value function in Algorithm \ref{alg:upwind_nd} (degree was just linear in $m$). Again, Algorithm \ref{alg: GOAP} is feasible practically only for $m$ small. These algorithms, although not exponential in $B$ and hence an improvement over \cite{komiyama2022minimax}, still suffer from exponential dependence on $m$. In the next few sections, we propose an algorithm based on ideas of Blackwell approachability that does not have such a bad dependence on $m$ and can also be optimal in certain cases. We will also show that this algorithm is uniformly better in all ASHT problems than any static algorithm,
\section{A meta strategy and its relation with Blackwell Approachability.}\label{sec:approachability}
We consider the following \emph{'meta'} strategy to construct a class of fixed budget algorithms:
\begin{enumerate}
    \item We divide the samples into $B$ batches.
    \item We use an allocation function $w_t \in \Delta_K$ such that for all samples in the $t^{th}$ batch, that is  $t*T/B+1$ sample to $(t+1)T/B$ sample, we pull arm $a$ exactly $w_t(a)*T/B$ number of times.
    \item After $T$ samples are exhausted, we declare the best arm of the Maximum Likelihood Bandit (MLB) from $\xi$ as our decision:
    \begin{equation*}
    \hat{i}_T=i^{*}(v_{MLB}(T))
    \end{equation*}
\end{enumerate}
We need to decide how the allocation $w_t$ for each batch in determined to get a concrete algorithm from the meta strategy. In what follows, we will use Blackwell's approachability algorithm for an appropriate target set to determine our allocation functions. In this section we will abuse notation and denote $Q^{t}:=Q^{tB}$ from here onwards. In this section we assume that the each $\xi_a$ is singleton.

\subsection{A generalized Sanov result for the meta strategy.}

We assume that each arm distribution for every bandit in $\xi$ is on a measurable space $(\mathcal{X},2^{\mathcal{X}})$ with $|\mathcal{X}|< \infty$ and support of all probability measures is $\mathcal{X}$. Let the set of all probability measures on $(\mathcal{X},2^{\mathcal{X}})$ be denoted as $\mathcal{P}$.  For any $m \in \mathbb{N}$, we denote any collection of $m.K$ probability measures on $\mathcal{X}$ as $\mu^{m}$, with $\mu_{a,j}$ denoting the $j^{th}$ measure of arm $a$. We also assume that the allocation function $w_t: Q^{t-1} \mapsto \Delta_K$ $( \forall t \in [T])$, is a function of the past empirical distribution. 
\begin{theorem}[Generalized Sanov]\label{sanov_thm}
For any set $\Gamma \subset \mathcal{P}^{BK}$, under the above meta strategy (along with arbitrary allocation functions) we have:
\begin{equation}\label{sanov_thm_eq}
 \underset{T \to \infty}{\liminf}-\frac{\log(P_{\nu}[Q^{B} \in \Gamma])}{T} \geq   \underset{\mu^{B} \in \bar{\Gamma}}{\inf}\frac{\sum^{B}_{j=1}\sum^{K}_{a=1}w_j(a)(\mu^{j-1}) D(\mu_{a,j}||\nu_a)}{B},
\end{equation}
where we have assumed the underlying bandit instance is $\nu$.
\end{theorem}
The point of this result is that it gives us a way to lower bound the error exponent of the meta strategy with any allocation function. It says the error exponent is atleast the the KL divergence between the minimizing distribution in the set $\Gamma$ and the actual arm distribution weighted by the allocation decision and then averaged across the sample batches $B$. The proof of Theorem \ref{sanov_thm} is given in Appendix \ref{sanov_proof}.

\subsection{Achieving an error exponent and Blackwell Approachability.}
When we apply a particular allocation function $w_t$ along with the meta strategy we get an algorithm for the fixed budget problem. Motivated from Theorem \ref{sanov_thm} (and eqn. \ref{sanov_thm_eq}) we define:
\begin{definition}
We say the above meta strategy (with any associated allocation function), denoted by $\mathbb{A}_{w,B}$ \textbf{achieves} atleast a minimax exponent of $R$ if:
\begin{equation*}
e_{m}(\xi,\mathbb{A}_{w,B}) \geq R.
\end{equation*}
\end{definition}
We then have the following result:
\begin{lemma}\label{approach_relate_err_exp}
A concrete realization of the meta algorithm $A_{w,B}$ achieves atleast an exponent $R$ if for every sample path $Q^{B}$ we have:
\begin{equation}
\underset{j}{\max} \hspace{0.2cm} \underset{i \neq j}{\min}\sum^{B}_{t=1}\sum^{K}_{a=1}\frac{w_t(a)(Q^{t-1}) D(\mu_{a,t}||\nu^{i}_a)}{B} \geq R
\end{equation}
\end{lemma}
\begin{proof}
Using eqn. (\ref{sanov_thm_eq}) and the definition of achieveability above we have that $A_{w,B}$ achieves an exponent $R$ if:
\begin{equation}\label{lemma_ineq_1}
\underset{Q^{B} \in \bar{\Gamma}_i}{\inf}\sum^{B}_{t=1}\sum^{K}_{a=1}\frac{w_t(a)(Q^{t-1}) D(\mu_{a,t}||\nu^{i}_a)}{B} \geq R
\end{equation}
for each $i \in \xi$. Here $\Gamma_i$ is the error set associated with bandit instance $\nu^i$. Now let us describe the set $\Gamma_i$- it is the collection of those sample paths where the MLB is not $\nu_i$. In other words for every $Q^{B} \in \Gamma_i$, we must have:
\begin{equation}\label{lemma_ineq_2}
\sum^{B}_{t=1}\sum^{K}_{a=1}\frac{w_t(a)(Q^{t-1}) D(\mu_{a,t}||\nu^{i}_a)}{B} > \sum^{B}_{t=1}\sum^{K}_{a=1}\frac{w_t(a)(Q^{t-1}) D(\mu_{a,t}||\nu^{j}_a)}{B},
\end{equation}
for some $j \neq i$. Here, we have used the fact that $\sum^{B}_{t=1}\sum^{K}_{a=1}\frac{w_t(a)(Q^{t-1}) D(\mu_{a,t}||\nu^{i}_a)}{B}$ represents the negative log-likelihood of the sample path $Q^{B}$. Let $C_j=\underset{i \neq j}{\cap}\bar{\Gamma}_i$. We note that every sample path must belong to the set $\underset{j \in [K]}{\cup} C_j$. WLOG let $Q^{B} \in C_j$ for some $j \in [K]$. Then we  must have:
\begin{equation*}
\sum^{B}_{t=1}\sum^{K}_{a=1}\frac{w_t(a)(Q^{t-1}) D(\mu_{a,t}||\nu^{i}_a)}{B} \geq \sum^{B}_{t=1}\sum^{K}_{a=1}\frac{w_t(a)(Q^{t-1}) D(\mu_{a,t}||\nu^{j}_a)}{B}
\end{equation*}
for every $i \neq j$ from eqn. (\ref{lemma_ineq_2}). Now from the fact that $Q^{B} \in \bar{\Gamma}_i$ for each $i \neq j$ and eqn. (\ref{lemma_ineq_1}) we have that:
\begin{equation*}
\sum^{B}_{t=1}\sum^{K}_{a=1}\frac{w_t(a)(Q^{t-1}) D(\mu_{a,t}||\nu^{i}_a)}{B} \geq R,
\end{equation*}
for every $i \neq j$. This implies the result for the path $Q^{B}$. As $Q^{B}$ was a generic sample path, the result follows.
\end{proof}
The lemma captures the fact that to achieve an exponent of R it is sufficient to ensure the sample path average of weighted empirical divergences are greater than $R$. 

\begin{definition}[Blackwell approachability problem]\label{defn: approachability}
Consider two players 1 and 2, who choose actions from two compact sets $U$ and $V$ respectively. Let $f: U \times V \mapsto R^{d}$ be the loss vector. We assume in each round, player 1 plays first and player 2 can respond after seeing the response of player 1. There are T rounds where the players repeatedly choose actions from $U,V$. Let $H$ be a closed set. The goal of of player 1 is to ensure that:
\begin{equation*}
\limsup_{T \to \infty} d \left(\frac{\sum^{T}_{t=1}f(v_t,u_t)}{T},H \right)=0,
\end{equation*}
that is to get arbitrarily close to the set H in euclidean distance $d(.,.)$. In this case, we say that the closed set $H$ is \textbf{approachable} by player 1.The goal of player 2 is to prevent player 1 from achieving this. We call $H,T,U,V,f$ as the parameters of the Blackwell approachability problem.
\end{definition}
Now consider in the above definition of Blackwell approachability with the following target set $H$:
\begin{equation*}
H=\bigg \{ x \in R^{K} \hspace{0.2cm} \big | \hspace{0.2cm} \underset{j}{\max} \hspace{0.2cm} \underset{i \neq j}{\min} x_i \geq R \bigg \}
\end{equation*}
$T=B$, $U=\Delta_K$, $V=\Delta_{\mathcal{X}}^{K}$ and $f$ whose each component is $f_i(w,Q)=\sum^{K}_{a=1}w(a) D(Q_a||\nu^{i}_a)$. Combining this with the result of Lemma \ref{approach_relate_err_exp} we get the following conclusion:
\\
\\
\emph{ \large An exponent R is achievable in the asymptotic limit $B \to \infty$ if the Blackwell approachability problem with the above parameters is approachable!}
\\
\\
Through this conclusion we have transformed the problem of achieving a minimax exponent $R$ to one about approachability of an appropriate Blackwell Approachability problem. The interesting point is that there are a number of algorithms for approachability that has been discovered over the years (see \citet{abernethy2011blackwell,mannor2014approachability,perchet2015exponential,shimkin2016online} and references therein) beginning with the seminal contribution of Blackwell himself \cite{blackwell1956analog}. We describe his algorithm next.

\begin{remark}
The problem statement of approachability used in this work is stronger than the usual notions of Blackwell approachability considered in literature. The case studied often in literature is asymptotic approachability with only high probability or the stronger notion of almost sure asymptotic approachability. Usually, both players are allowed to play mixed strategies over their action spaces and the randomness is due to these mixed strategies. Also, usually both players are assumed to play simultaneously in each round while here we have a Stackelberg structure of Player 1 being the leader and player 2 being the follower. In spite of these differences, the basic outline of these approachability algorithms will be useful in our work.
\end{remark}

\subsection{Blackwell's Approachability algorithm.}
Blackwell in his seminal work \cite{blackwell1956analog} gave a sufficient condition for the target sets to satisfy - the notion of a \textbf{B-set} and gave a corresponding algorithm to asymptotically approach such sets.
\begin{definition}[B-set]\label{defn: b_set}
A closed set $H \subset R^{d}$ is called a \textbf{B-set} if for any point $x \notin H$ and for some projection point $y \in H$ of $x$, there exists an action $u^{*}(x) \in U$, such that:
\begin{equation}\label{eq:b_set_cond}
\langle x-y,f(u^{*}(x),v)-y \rangle \leq 0
\end{equation}
for all $v \in V$.
\end{definition}
Blackwell after identifying this geometric condition gave a very simple algorithm for such target sets: 
\begin{itemize}
    \item If the set H is a $B$-set and if $\bar{f}_t:= \frac{\sum^{t}_{s=1}f(u_t,v_t)}{t} \notin H$ then at $t+1$ play $u^{*}(\bar{f}_t)$. Play an arbitrary $u \in U$ otherwise.
    \item He further shows that if the target set $H$ is also convex then that $B$-set condition is also necessary. 
\end{itemize}
\begin{remark}
In the next computation we will show that the $B$-set definition is sufficient even for closed sets. As a result any closed set containing a $B$-set is also approachable even if itself isn't one. \cite{spinat2002necessary} shows that for a closed set to be approachable it is also necessary for it to contain a $B$-set. We note that the notion of approachability used in Spinat's work is weaker than ours. But as even this weaker notion necessarily requires a B-set to be a subset of the target set, it is also necessary in our case.
\end{remark}
In what follows let us denote the projection point $y$ of $x$ onto the closed set $H$ as $\prod_{H}(x)$. Because $H$ is a closed set, the projection set is not empty. Now, we derive a recursive inequality assuming the set $H$ is a $B$-set:
\begin{equation*}
\begin{aligned}
\bigg| \bigg|\bar{f}_{t+1}-\prod_H(\bar{f}_{t+1})\bigg| \bigg|^{2}_2 & \stackrel{(a)}{\leq} \bigg| \bigg|\bar{f}_{t+1}-\prod_H(\bar{f}_{t})\bigg| \bigg|^{2}_2 \\
& = \bigg| \bigg|\frac{t}{t+1}\bar{f}_{t+1}+\frac{1}{t+1}f(u_{t+1},v_{t+1})-\frac{t}{t+1}\prod_H(\bar{f}_{t})-\frac{1}{t+1}\prod_H(\bar{f}_{t}) \bigg| \bigg|^2_2\\
&=\left( \frac{t}{t+1} \right)^{2}\bigg| \bigg|\bar{f}_{t}-\prod_H(\bar{f}_{t})\bigg| \bigg|^{2}_2+\left( \frac{1}{t+1} \right)^{2} \bigg| \bigg|f(u_{t+1},v_{t+1})-\prod_H(\bar{f}_{t})\bigg| \bigg|^{2}_2 \\
& \quad +2 \bigg \langle \bar{f}_{t}-\prod_H(\bar{f}_{t}), f(u_{t+1},v_{t+1})-\prod_H(\bar{f}_{t}) \bigg \rangle\\
& \stackrel{(b)}{\leq} \left( \frac{t}{t+1} \right)^{2}\bigg| \bigg|\bar{f}_{t}-\prod_H(\bar{f}_{t})\bigg| \bigg|^{2}_2+\left( \frac{1}{t+1} \right)^{2} \bigg| \bigg|f(u_{t+1},v_{t+1})-\prod_H(\bar{f}_{t})\bigg| \bigg|^{2}_2 \\
& \stackrel{(c)}{\leq}\left( \frac{t}{t+1} \right)^{2}\bigg| \bigg|\bar{f}_{t}-\prod_H(\bar{f}_{t})\bigg| \bigg|^{2}_2+\left( \frac{M}{t+1} \right)^{2},\\
\end{aligned}
\end{equation*}
where $(a)$ is because of the projection property, $(b)$ follows from the B-set condition and $(c)$ follows from compactness of $U,V$, continuity of $f$ and the consequent boundedness of the projection operation. Thus we have shown that:
\begin{equation*}
(t+1)^2\bigg| \bigg|\bar{f}_{t+1}-\prod_H(\bar{f}_{t+1})\bigg| \bigg|^{2}_2 \leq t^2 \bigg| \bigg|\bar{f}_{t}-\prod_H(\bar{f}_{t})\bigg| \bigg|^{2}_2+M^2
\end{equation*}
Recursing this inequality over $t$ we get the following bound:
\begin{equation}\label{eq:approach_compute}
\bigg| \bigg|\bar{f}_{T+1}-\prod_H(\bar{f}_{T+1})\bigg| \bigg|^{2}_2 \leq \frac{||\bar{f}_{1}-\prod_H(\bar{f}_{1})||^{2}_2}{(T+1)^2}+\frac{M^2}{T+1}
\end{equation}
Consequently taking the limit $T \to \infty$ we have that:
\begin{equation*}
\lim_{T \to \infty} ||\bar{f}_{T+1}-\prod_H(\bar{f}_{T+1})||^{2}_2 =0.
\end{equation*}
This computation shows the sufficiency of the B-set condition in approaching such closed sets.
\section{Constructing an appropriate B-set of the target set.}\label{sec:construction b_set}
From the results of previous section we saw that achieving asymptotically ($B \to \infty$) a minimax exponent using the meta strategy boils down to solving a particular Blackwell approachability problem. However, whether the problem is approachable is not clear. Our goal in this section is to identify a particular subset of the target set which we can definitively show is a B-set.
\subsection{Construction based on separating hyperplanes.}\label{hyperplane_subsec}
The target set has the following form:
\begin{equation*}
H=\bigg \{ x \in R^{K} \hspace{0.2cm} \big | \hspace{0.2cm} \underset{j}{\max} \hspace{0.2cm} \underset{i \neq j}{\min} x_i \geq R \bigg \}
\end{equation*}
Our task therefore is to identify a subset of this set which is provably also a B-set. In our approachability problem $U=\Delta_K, V=\Delta_{\mathcal{X}}^{K}$ and $f(w,v)=( \sum^{K}_{a=1}w(a) D(v_a||\nu^{i}_a) )_{i \in [m]}$. We start by trying to construct a hyperplane such that for some $w \in \Delta_K$, the set $f(w,V)$ is to one side of the hyperplane. Noticing the the geometrical structure of the set H (it is always in the first quadrant and unbounded and excludes $0$), we choose a normal vector $\beta \in \Delta_m$. We now need to choose for the given $\beta$, its corresponding intercept value $I(\beta)$ so that there exists some $w(\beta) \in \Delta_m$ such that $\langle \beta, f(w(\beta),v) \rangle \geq I(\beta)$ for all $v \in V$. To this end, first consider the following optimization problem for a fixed arm $a$:
\begin{equation}\label{beta_optimization}
\underset{Q_a}{\inf}\sum^{m}_{i=1}\beta_i D(Q_a||\nu^{i}_a)=-\log \left( \sum_{x \in \mathcal{X}} \prod^{m}_{i=1} (\nu^{i}_a(x))^{\beta_i} \right).
\end{equation}
This optimization is a simple extension of minimizing weighted KL divergence used in information theory (see a proof in Appendix \ref{beta_optim_proof}). Now we observe that choosing any $w \in \Delta_K$ and defining 
\begin{equation*}
  I(w,\beta)= -\sum^{K}_{a=1}w_a \log \left( \sum_{x \in \mathcal{X}} \prod^{m}_{i=1} (\nu^{i}_a(x))^{\beta_i} \right)   
\end{equation*}
ensures that for $ \forall v \in V$ we must have that $\langle \beta, f(w,v)\rangle \geq I(w,\beta)$. A choice we will make is to set 
\begin{equation*}
w(\beta)= \underset{w \in \Delta_K}{argmin} \hspace{0.2cm} I(w, \beta)=  \underset{a \in [K]}{argmin} \hspace{0.2cm} I(\delta_a,\beta).  
\end{equation*}
This ensures that we finally have $ \langle \beta, f(w(\beta),v) \rangle =\sum_{i} \sum_a \beta_i w_a(\beta) D(v_a|| v^{i}_a) \geq I(\beta)=\underset{a \in [K]}{\min}-\log \left( \sum_{x \in \mathcal{X}} \prod^{m}_{i=1} (\nu^{i}_a(x))^{\beta_i} \right).$ Note that in this construction of the halfplane the only assumption is that $\beta \in \Delta_m$. But the halfplanes $H_{\beta}= \{ x | \langle \beta, x \rangle \geq I(\beta) \}$ will never be contained within the target set H (for example the point $x_1=I(\beta)/\beta_1$ and $x_2= \dots=x_m=0$ is not contained in H). To try to create a set contained within the target set H, we choose to consider the following set 
\begin{equation}
S= \underset{\beta \in \Delta_K}{\cap}H_{\beta}.
\end{equation}
\subsection{Sufficient condition to ensure the B-set is a subset of H.}\label{subset_H}
From section \ref{info_th_lb} we know that largest achievable $R$ in 
\begin{equation*}
H=\bigg \{ x \in R^{K} \hspace{0.2cm} \big | \hspace{0.2cm} \underset{j}{\max} \hspace{0.2cm} \underset{i \neq j}{\min} x_i \geq R \bigg \}
\end{equation*}  
is going to be $R^{go}_{\infty}$. So we will try to construct the B-set for $R \leq R^{go}_\infty$. 
We will modify $I(\beta)$ to $I^{'}(\beta)$ in the definition of S above to ensure H contains the B-set. Let us define $H^{'}_{\beta}= \{ x | \langle \beta, x \rangle \geq I^{'}(\beta) \}$ and hence define :
\begin{equation*}
    S^{'}=\underset{\beta \in \Delta_K}{\cap}H^{'}_{\beta}.
\end{equation*}
We will define $I^{'}(\beta)$ later in such a way that $I(\beta) \leq I^{'}(\beta) \leq \underset{a \in [K]}{\max}-\log \left( \sum_{x \in \mathcal{X}} \prod^{m}_{i=1} (\nu^{i}_a(x))^{\beta_i} \right) := L(\beta)$. This ensures that for $ \forall v \in V$ we must have that $\langle \beta, f(w(\beta),v)\rangle \geq I^{'}(\beta)$ with $w(\beta)$ being chosen as a solution to
\begin{equation*}
I^{'}(\beta)=-\sum^{K}_{a=1}w_a \log \left( \sum_{x \in \mathcal{X}} \prod^{m}_{i=1} (\nu^{i}_a(x))^{\beta_i} \right).
\end{equation*}
Let us now define $I^{'}(\beta)$. From Claim \ref{second_order_claim} we know that :
\begin{equation*}
\underset{j}{\max} \hspace{0.2cm} \underset{i \neq j}{\min} \hspace{0.2cm} x_i = \underset{i \neq j}{\min} \max \left( x_i,x_j \right).
\end{equation*}
Let $\beta_{i,j}=(\underbrace{0,\dots,0}_\text{$i-1$ zeros},\beta,\underbrace{0,\dots,0}_\text{$j-i-1$ zeros},1-\beta,\underbrace{0,\dots,0}_\text{$m-j$ zeros})$ with $1 \leq i<j \leq m$ and $\beta \in [0,1]$. Thus, for any $x \in S^{'}$ we must have that:
\begin{equation*}
max(x_i,x_j)=\underset{\beta \in [0,1]}{\max} \beta x_i+(1-\beta)x_j = \underset{\beta \in [0,1]}{\max} \langle \beta_{ij},x \rangle \geq \underset{\beta \in [0,1]}{\max} I^{'}(\beta_{ij}).
\end{equation*}
Now we set $I^{'}(\beta)=I(\beta)$ everywhere except at  one $\tilde{\beta}_{ij}$ associated with each pair $(i,j)$ (with $i<j$), such that $\underset{\beta \in [0,1]}{\sup}I(\beta_{ij})<R$ \footnote{Note that the set of perturbed pairs depend on the chosen R.}. In this case  for such $(i,j)$ pairs we choose any one $\tilde{\beta}_{ij}$ and set $I^{'}(\tilde{\beta}_{ij})=R$. Now by construction we have that:
\begin{equation*}
max (x_i,x_j) \geq \underset{\beta \in [0,1]}{\max} I^{'}(\beta_{ij}) \geq I^{'}(\tilde{\beta}_{ij}) \geq R,
\end{equation*}
for every pair $i \neq j$. This then implies that $S^{'} \subset H$, which is as required.
\begin{remark}
In this way one can view $I^{'}(\beta)$ as perturbation of $I(\beta)$. In this case $I(\beta)$ is perturbed at specific points associated with certain pairs $(i,j)$ that satisfy the condition $\underset{\beta \in [0,1]}{\sup}I(\beta_{ij})<R$. We note that in the next subsection we will use the concave upper envelope of $I^{'}(\beta)$, denoted as $I^{'}_{concave}(\beta)$ over the simplex instead of $I^{'}(\beta)$ itself. Nevertheless, the argument made for $I{'}(\beta)$ in this subsection can be reapplied for $I^{'}_{concave}(\beta)$ as well to ensure the constructed B-set is a subset of $H$.
\end{remark}
\subsection{Putting it all together and constructing the B-set.}\label{subsec:b_set_construction_approach_algo}
We will first define the the upper concave envelope of an real valued function:
\begin{definition}[Upper concave envelope]\label{concv_env}
 The \textbf{upper concave envelope (uce)} of any real valued function $g : Z \mapsto \mathbb{R}$ (with an convex finite dimensional domain $Z$) is the smallest concave function that point-wise dominates the function $g$. The uce of a function $g$ will be denoted as $g_{concave}$.   
\end{definition}
It is also a well known result (see Theorem 1.3.7 in \cite{hiriart2004fundamentals}) that the $f_{concave}$ has the following representation:
\begin{equation*}
g_{concave}(x)=\sup \bigg \{ \sum^{m}_{i=1}\lambda_i g(x_i) \hspace{0.2cm} \bigg | \quad x=\sum^{m}_{i=1}\lambda_i x_i, m \in \mathbb{N},\forall i \in [m] \text{ }\lambda_i \geq 0  \text{ and } \sum^{m}_{i=1} \lambda_i=1  \bigg \}.
\end{equation*}
We will consider the concave envelope of the function $I^{'}(\beta)$. From the above variational representation, the concavity of $I(\beta)$ and the definition of $I^{'}(\beta)$ as a perturbation of $I(\beta)$ we have that:
\begin{equation} \label{var_rep_uce}
\begin{aligned}
I^{'}_{concave}(\beta)=\sup \bigg \{ R \sum_{(i,j) \in M(R)}\lambda_{ij}+\lambda_b I(\beta^{'}) \hspace{0.2cm} \bigg | & \quad \beta=\sum_{(i,j) \in M(R)}\lambda_{ij} \tilde{\beta}_{ij}+\lambda_b \beta^{'},\forall (i,j) \in M(R), \text{ }\lambda_{ij} \geq 0, \lambda_b \geq 0\\
& \text{ and } \lambda_b + \sum_{(i,j) \in M(R)}\lambda_{ij}=1  \bigg \},
\end{aligned}
\end{equation}
where $M(R)$ is the set of $(i,j)$ pairs that were perturbed to create $I^{'}(\beta)$, $\tilde{\beta}_{ij}$ are the associated perturbation points for the perturbed $(i,j)$ pairs (see the previous subsection \ref{subset_H} for precise details) and $\beta^{'}$ is any element from $\Delta_K$. Let us now define halfplanes $\hat{H}(\beta)= \{ x | \langle \beta, x \rangle \geq I^{'}_{concave}(\beta) \}$. And we define the corresponding B-set as:
\begin{equation*}
    \hat{S}=\underset{\beta \in \Delta_m}{\cap}\hat{H}_{\beta}.
\end{equation*}
Further let us define the function $l(x)=\underset{\beta \in \Delta_K}{\inf} \langle \beta, x \rangle - I^{'}_{concave}(\beta).$ Then equivalently we have that $\hat{S}= \{x | l(x) \geq 0$ \} and the boundary $\partial \hat{S}= \{ x | l(x)=0 \}$, that is the level set of $l$.
Next, we state two lemmas from convex analysis that will be useful to us. The proofs are in the appendices \ref{subdiff_conj_lemma_proof} and \ref{normal_sublvl_proof} respectively.
\begin{lemma}\label{subdiff_conj_lemma}
Consider concave conjugate $f^{*}$ of a concave function $f$ defined over the simplex:
\begin{equation*}
f^{*}(x)=\inf_{b \in \Delta_n} \{ \langle b , x\rangle -f(b) \}.
\end{equation*}
Then we can characterize the sub-differential of $f^{*}$ as:
\begin{equation}\label{subdiff_conj}
\partial f^{*}(x)=\underset{b \in \Delta_n}{argmin} \{ \langle b , x\rangle -f(b) \}.
\end{equation}
\end{lemma}
\begin{lemma}\label{normal_sublvl}
 Let $f$ be a concave function that is finite-valued (not taking $\pm \infty$ as values). Assume that $\bar{x}$ is such that $f(\bar{x}) < \sup f(x)$ and define the sub-level set:
 \begin{equation*}
    C=\{ x | f(x) \geq f(\bar{x}) \}.
 \end{equation*}
 Then we have that:
 \begin{equation*}
     N_{C}(\bar{x})=Cone(-\partial f (x)),
 \end{equation*}
 where $N_c(\bar{x})$ denotes the normal cone to C at $\bar{x}$.
\end{lemma}
\begin{theorem}\label{blackwell_approach_thm}
If $R \leq  \underset{\lambda \in \Delta_{M(R)}}{\inf}L \left( \sum_{(i,j) \in M(R)} \lambda_{ij}\tilde{\beta}_{ij} \right)$, then $\hat{S}$ as defined above is a B-set. Thus the Approachability algorithm \ref{alg: approachability} can achieve an exponent of atleast $R$ (assuming $B \to \infty$) in this case.
\end{theorem}
The proof of the Theorem is given in Appendix \ref{proof_blackwell_approach_thm}. 
 We examine the assumption $R \leq \underset{\lambda \in \Delta_{M(R)}}{\inf}L \left( \sum_{(i,j) \in M(R)} \lambda_{ij}\tilde{\beta}_{ij} \right)$. Now as this determines the maximum exponent achievable by the algorithm we choose the variables $\tilde{\beta}_{ij}$ to maximize this exponent. The second point is that this is a fixed point inequality. We try to seek the largest possible $R$ that satisfy the inequality. Thus, we have:
\begin{corollary}\label{optimal_exp_approach}
The Approachability algorithm \ref{alg: approachability} can achieve (assuming $B \to \infty$) the following exponent:
\begin{equation}\label{R_approach_exponent}
R_{approach}=\sup \bigg \{ R \hspace{0,2cm} \bigg|  \hspace{0.2cm} R \leq \underset{\tilde{\beta}_{ij}}{\sup}\underset{\lambda \in \Delta_{M(R)}}{\inf}L \left( \sum_{(i,j) \in M(R)} \lambda_{ij}\tilde{\beta}_{ij} \right) \bigg \}.
\end{equation}
\end{corollary}
The following lemma gives an alternate representation of $l$:
\begin{lemma}\label{alt_rep_l_blackwell}
Given that $\tilde{\beta}_{i,j} \in M(R)$ are the set of perturbations used in defining $I^{'}_{concave}(.)$  we have that:
\begin{equation}\label{alt_rep_eq}
 l(x)= \min \bigg \{ \underset{(i,j) \in M(R)}{\min} \{ \langle \tilde{\beta}_{i,j},x \rangle -R\}, \min_{\beta \in \Delta_m} \{ \langle \beta, x \rangle -I(\beta)\}  \bigg \}.   
\end{equation}
\end{lemma}
The proof of the lemma is given in Appendix \ref{proof_alt_rep_l_blackwell}. This representation is computationally more feasible since we have to minimize over a set of finite points and a convex program. \\

Next, let us define the function $G(R)$ such that:
\begin{equation*}
G(R)= \begin{cases}
 +\infty & R< \underset{i \neq j}{\min}\underset{\beta \in [0,1]}{\sup}I(\beta_{ij}).\\
 \underset{\tilde{\beta}_{ij}}{\sup}\underset{\lambda \in \Delta_{M(R)}}{\inf}L \left( \sum_{(i,j) \in M(R)} \lambda_{ij}\tilde{\beta}_{ij} \right) & R \geq \underset{i \neq j}{\min}\underset{\beta \in [0,1]}{\sup}I(\beta_{ij}).
\end{cases}
\end{equation*}
\begin{lemma}\label{prop_G_R}
The function $G(.)$ has the following properties:
\begin{enumerate}
    \item $G(.)$ is non-increasing in R.
    \item Denote by $a_{s}$ as the $s^{th}$ $(i,j)$ pair when ordered according to $\underset{\beta \in [0,1]}{\sup}I(\beta_{ij})$. Then the function $G(.)$ is constant between $R \in (a_{s},a_{s+1}]$.    
\end{enumerate}
\end{lemma}
Proof of the lemma is given in appendix \ref{prop_G_R_proof}. The above lemma implies that $G(.)$ is a piecewise constant function with jumps at $a_s$ and these jumps always decrease the value as $R$ increases. This implies the optimal exponent $R_{approach}$ defined through a fixed point inequality  in Corollary \ref{optimal_exp_approach} is uniquely attained. Let us rewrite this optimal exponent $R_{approach}$ as:
\begin{equation*}
R_{approach}=\sup \{ R: R \leq G(R) \}.
\end{equation*}
We then have that:
\begin{lemma}\label{better_than_static}
The Approachability algorithm \ref{alg: approachability} outperforms (assuming $B \to \infty$) the best possible static exponent:
\begin{equation*}
R_{static} \leq R_{approach}. 
\end{equation*}
\end{lemma}
The proof is in \ref{proof_better_than_static}.We present the pseudocode for the Approachability Algorithm:
\begin{algorithm}
\caption{Approachability algorithm.}
\label{alg: approachability}
\begin{algorithmic}[1]
\Statex \textbf{Inputs:}
\Statex Unknown bandit $\nu \in \xi$, No. of Samples $T$, Bandit class $\xi$, , No of batches - $B$.
\Statex \textbf{Initialization:}
\State $R_{approach} \gets \sup \{ R: R \leq G(R) \}$.
\State Obtain the optimal $\tilde{\beta^{*}_{i,j}}$'s for $R_{approach}$ from eq. \eqref{R_approach_exponent}.
\State $\hat{S} \gets \{x \hspace{0.2cm} | \hspace{0.2cm} l(x) \geq 0. \}$.
\State Initialize $\mat{x}$ as all-zero $m$- dimensional vector.\Comment{State vector initialized.}\\
\For{$n$ from $0$  to $B-1$} 
    \State \quad \textbf{if} \(l(\mat{x}) \geq 0\) \textbf{then}
        \State \quad \quad \( w_{n} \gets \text{ Any distribution } w \in \Delta_K \) 
    \State \quad \textbf{else} 
        \State \quad \quad \( y_n \in \underset{y \in \hat{S}}{argmin}||\mat{x}-y|| \) \Comment{Project onto B-set.}
        \State \quad \quad \( \beta_n \in \underset{ \beta \in \Delta_m}{argmin} \{ \langle \beta, y_n \rangle-I^{'}_{concave}(\beta) \}  \)
        \State \quad \quad \( w_n \gets \text{ Any } w \in \Delta_K \text{ such that } I(w,\beta_n)=\langle \beta_n , y_n \rangle \).
    \State \quad \textbf{end if}
    \State Sample each arm $a$ of bandit $\nu$ exactly $w_n(a)T/B$ times and observe the empirical distribution $Q_n$.
    \State $\mat{x} \gets \frac{n}{n+1}\mat{x}+\frac{1}{n+1}\left(\sum_a w_n(a) D(Q_{n,a}|| \nu^{(i)}_{a} )\right)_{i \in [m]}$ . \Comment{Update state vector.}
\EndFor
\Statex
\State \Return $\underset{i \in [m]}{argmin} \hspace{0.2cm} \mat{x}_i$. 
\end{algorithmic}
\end{algorithm}
\FloatBarrier
From the general bound for Blackwell Approachability in eq. \eqref{eq:approach_compute} and the fact that $\hat{S}$ defined for $R=R_{approach}$ is a $B$-set we have the following result:
\begin{equation*}
d_{\hat{S}}(\mat{x}) \leq \frac{C}{B},
\end{equation*}
where $\mat{x}$ is the final state vector obtained after running Approachabiltiy algorithm \ref{alg: approachability} and $d_{\hat{S}}(.)$ denotes the distance from set $\hat{S}$. From the considerations in Section \ref{subsec:differential_game_interp}, we know that:
\begin{equation*}
g(x)=\max_{j} \underset{i \neq j}{\min} x_i=x_{(2)}
\end{equation*}
is a Lipschitz function. Since $\hat{S} \subset \{ g(x) \geq R_{approach} \}$ we have that for some $y \in \{ g(x) \geq R_{approach} \}$ that:
\begin{equation*}
|g(\mat{x})-g(y)| \leq L_{g} ||x-y|| \leq \frac{L_g C}{B}
\end{equation*}
From this we conclude that:
\begin{equation*}
R_{approach}-\frac{L_g C}{B} \leq g(y)-\frac{L_g C}{B} \leq g(\mat{x}).
\end{equation*}
As the approachability algorithm is a "meta" strategy and we have shown a pathwise lower bound we can apply Lemma \ref{approach_relate_err_exp} to get the following result:
\begin{theorem}\label{finite_b_minimax_approach_bd}
The Approachability algorithm \ref{alg: approachability} has the following guarantee:
\begin{equation}\label{approach_guarantee_eq}
e_{m}(\xi,Approachability) \geq R_{approach}-\frac{L_g C}{B}
\end{equation}
that is, the minimax exponent $e_{m}(\xi,Approachability)$ can be made arbitrarily close to $R_{approach}$ if we choose $B$ large enough. The constant $C$ depends on the distance of origin to the set $\hat{S}$.
\end{theorem}
We also observe in some problems numerically that the $R_{approach}$ is equal to the optimal adaptive exponent $R^{go}_{\infty}$. In the following example:
\begin{table}[h]
\centering
\begin{tabular}{|c|c|}
\hline
Quantity &Value \\
\hline
K & 3 \\
\hline
m & 3 \\
\hline
$\nu^{(1)}$ & $ (0.6, 0.48, 0.23)$\\
$\nu^{(2)}$ & $ (0.48, 0.6, 0.23)$\\
$\nu^{(3)}$ & $ (0.43, 0.5, 0.6)$\\
\hline
$R^{ub}$ & $0.0073001 \pm 2 \times 10^{-6}$  \\
\hline
$R^{go}_{1}$ & $0.0073001 \pm 1 \times 10^{-6}$\\
\hline
$R_{static}$ & $0.007003 \pm 3 \times 10^{-5}$\\
\hline
$R_{apporach}$ & $0.0073001 \pm 2 \times 10^{-6}$\\
\hline
\end{tabular}
\caption{A simple bandit class $\xi=  \{ \nu^{(1)}, \nu^{(2)}, \nu^{(3)} \}$, with three $(m=3)$ three-armed bernoulli bandits ($K=3$). In this case we have that $R^{ub}=R^{go}_1=R^{go}_{\infty}=R_{approach}>R_{static}$.}
\label{tab:num_example_2}
\end{table}
\FloatBarrier
we have that $R_{approach}=R^{go}_{\infty}=R^{go}_1>R_{static}$.
\begin{remark}
We observe in the numerical experiments that $R_{approach}$ can be close to $R^{go}_{\infty}$. It will be of interest to quantify the gap and identify conditions under which both the exponents are equal.
\end{remark}
\subsubsection{Computational effort in Approachability algorithm \ref{alg: approachability}.}
Using the properties listed in Lemma \ref{prop_G_R}, one can show that to compute $R_{approach}$  we need to compute the maximin program $G(.)$ $O(m^{2})$ times. The maximin $G(.)$ itself is a bi-level optimization and in the worst case can require $O(B^{m})$computational effort if solved to $1/B$ accuracy. Further in the loop (lines 6-13 in Algorithm \ref{alg: approachability}) we have to first evaluate $l(\mat{x})$ which can be efficiently done based on the representation given in Lemma \ref{alt_rep_l_blackwell}. The projection step (line 10) is also a convex program and can be done efficiently using projected sub-gradient methods for moderate values of $m$. Thus, the computational effort in Algorithm \ref{alg: approachability} is dominated by solving the non-convex bi-level optimization problem. While this can still be exponential in $m$, in practice this is much easier to compute compared to GOAP (Algorithm \ref{alg: GOAP}) and Explicit Upwind Scheme (Algorithm \ref{alg:upwind_nd}). Global optimization methods like differential evolution and particle swarm  optimization methods in particular can be very effective in practice and require a fraction of the time the grid based methods of section \ref{sec:numerical_pde_scheme} which in turn are more effective than the methods of \cite{komiyama2022minimax}.
\section{Summary.}\label{Summary_of_work}
In this work we consider the problem of finding the minimax exponent for the Active Simple Hypothesis Testing (ASHT) problem. We were able to re-characterize the upper bounds $R^{go}_B$ on the minimax exponents derived by \cite{komiyama2022minimax} as value of certain dynamic Stackelberg games (Proposition \ref{eq_rep_R_go_B}). The limit of these bounds $R^{go}_{\infty}$ was shown to be the value of a certain zero sum differential game (Theorem \ref{hji_pde_R_go_infty}). This recasting of the upper bounds into game-theoretic form gives new insight into the problem - Both players, nature and agent, are trying to control the final likelihood of the second most likely bandit in each sample path. The interpretation as a differential game allows one to utilize PDE methods for Hamilton Jacobi equations to effectively compute (Algorithm \ref{alg:upwind_nd} and Theorem \ref{convergence_thm_fd_scheme}) $R^{go}_{\infty}$ (atleast for small $m$). This allows us to numerically disprove a conjecture of \cite{komiyama2022minimax} in ASHT setting (subsection \ref{subsec: num_compute_conj_disprov}). The synthesis of optimal controls from the computation of value function has long been an open problem (see \cite{falcone2006numerical}, \cite{falcone2013semi}) in differential games. In our specific case, using the structure of our problem and the methods of \cite{taras1994approximation}, \cite{tarasyev1999control} we were able to propose GOAP (Algorithm \ref{alg: GOAP}) and rigorously prove its approximate optimality (Theorem \ref{goap_ldp_guarantee}). The grid based methods of section \ref{sec:numerical_pde_scheme} based on the PDE formulation (eq. \eqref{HJI_pde}) have theoretical guarantees and are effective for small $m$ in contrast to the methods proposed in \cite{komiyama2022minimax}. However, these methods do suffer from curse of dimensionality in $m$ and are not effective for moderate to large values of $m$. We instead propose an alternative algorithm based on Blackwell Approachability (section \ref{sec:approachability}). The usage of Blackwell Approachability to pure exploration bandit problems is novel as far as we are aware. Using this connection to Blackwell Approachability, we construct an appropriate B-set (subsection \ref{subsec:b_set_construction_approach_algo}) and provide guarantees (Theorem \ref{finite_b_minimax_approach_bd}) for the Approachability algorithm (\ref{alg: approachability}). Further, we characterize the exponent $R_{approach}$ achievable by Algorithm \ref{alg: approachability} (Corollary \ref{optimal_exp_approach}) and show that it outperforms any static algorithm for all problem instances of ASHT (Lemma \ref{better_than_static}). Further, numerically we show instances where $R_{approach}=R^{go}_\infty$ and generally observe that $R_{approach}$ can be quite close to $R^{go}_{\infty}$ numerically.
\section{Future Work.}\label{future_work}
The current work can be extended in a number of directions. We mention a few concrete problems that we consider to be of interest:
\begin{enumerate}
    \item As mentioned in the introduction \ref{intro}, the current work should be viewed as contributing to broader program of understanding minimax optimal algorithms for the more general fixed budget problem. In effect we need to remove the restriction that the hypotheses are simple. We need to be able to handle composite continuum hypotheses classes. From this work, it is clear that ideas from differential games will be important. But the associated dynamical equations of the differential game will be a set of infinite-dimensional ODEs (see for example \cite{kocan1997differential}) and the corresponding PDE would likely have state spaces as subsets of infinite dimensional  spaces (\cite{crandall1986hamilton}). It would be of interest to extend the current framework to the infinite dimensional case.
    \item For some closely related classes of PDEs, recently methods have been proposed that overcome the curse of dimensionality in the value function approximation (non-linear parabolic PDEs through multilevel Picard iteration \cite{hutzenthaler2019multilevel} and Hamilton Jacobi equations with convex Hamiltonian \cite{darbon2016algorithms}) have been proposed. Can these methods be extended to include the case of the Hamilton Jacobi Isaac PDE \eqref{HJI_pde} we consider?
    \item It would be of interest to get greater detail on the solution $V(t,x)$ of the PDE \eqref{HJI_pde}. However, the fact that both the Hamiltonian and the terminal functions being non-convex prevent application of direct representations like Hopf-Lax and Hopf formulae. Can we get more details about the solution $V(t,x)$ using PDE methods, for example those presented in \cite{evans2014envelopes} and \cite{melikyan2012generalized}?
    \item It would be of interest to characterize when $R_{approach}<R^{go}_{\infty}$ and to quantify the gap. One way to make this approach work is to use general minimax theorems. $R_{approach}$ and both the upper bounds $R^{go}_1$ and $  R_{Hopf}$ can be shown to be dual to $R_{approach}$. Although problem does not have the convexity/ quasi-convexity required to apply the usual Sion's minimax theorem, one can try to apply more general minimax theorems (for example see results in \cite{simons1995minimax}, \cite{mcclendon1984minimax}, \cite{repovvs2006minimax}).
\end{enumerate}
\newpage
\appendix
\section{Proofs of result in Section \ref{info_th_lb}.}
\subsection{Proof of Proposition \ref{eq_rep_R_go_B}.}\label{proof_prop_1}
\begin{proof}
From Theorem \ref{batched_B_lb} we have that:
\begin{equation*}
R^{go}_{B} = \underset{r^{B}(.),J(.)}{\sup} \hspace{0.2cm} \underset{Q^{B}}{\inf} \underset{\underset{J(Q^{B}) \neq i^{*}(\nu)}{\nu \in \xi}}{\inf} \frac{1}{B} \sum^{B}_{l=1}\sum_{a \in [K]}r_{l}(Q^{l})(a)D(Q_{l,a}||\nu_a). 
\end{equation*}
As $r^{B}(.)$ and $J(.)$ are functions taking $Q^{B}$ as argument we can just optimize the function for each fixed $Q^{B}$. So assuming $Q^{B}$ is fixed we want to optimize  
\begin{equation*}
\underset{\underset{J(Q^{B}) \neq i^{*}(\nu)}{\nu \in \xi}}{\inf} \frac{1}{B} \sum^{B}_{l=1}\sum_{a \in [K]}r_{l}(Q^{l})(a)D(Q_{l,a}||\nu_a),
\end{equation*}
while still respecting the sequential nature of the functions $r_1,r_2,\ldots, r_B,J$. This is then just a simple backward induction process. So, we start with assuming all $r_1,\ldots,r_B$ have been chosen and optimize only $J(.)$ given $Q^{B}$. Clearly the optimal choice here is :
\begin{equation*}
J^{*}(Q^{B})=\underset{j}{argmax} \hspace{0.1cm} \underset{j \neq i }{\min} \hspace{0.1cm} \underset{\nu \in \xi_i}{\inf} \frac{1}{B} \sum^{B}_{l=1}\sum_{a \in [K]}r_{l}(Q^{l})(a)D(Q_{l,a}||\nu_a).
\end{equation*}
Now after this optimization over $J(.)$ and fixing the choices $r_1,\ldots,r_{B-1}$, we optimize over $r_B$. This entails just taking a supremum over $\Delta_K$:
\begin{equation*}
r^{*}_{B}=\underset{w_B \in \Delta_K}{argmax} \hspace{0.1cm} \underset{j}{\max} \hspace{0.1cm} \underset{j \neq i }{\min} \hspace{0.1cm} \underset{\nu \in \xi_i}{\inf} \frac{1}{B} \sum^{B-1}_{l=1}\sum_{a \in [K]}r_{l}(Q^{l})(a)D(Q_{l,a}||\nu_a)+\frac{1}{B}\sum_{a \in [K]}w_B(Q^{B})(a)D(Q_{B,a}||\nu_a). 
\end{equation*}
Note that the argmax here is nonempty because $\Delta_K$ is compact and the optimizing objective is continuous in $w_{B}$, since $\xi_i$ are also compact. Now we observe that $r_1,\ldots,r-{B-1}$ need only $Q^{B-1}$ as input. The $R^{go}_B$ definition has an infimization over $Q^{B}$. This means we must now infimize over $Q_B$ first and then again optimize over $r^{B-1}$ assuming $Q^{B-1}$ and $r_1,\ldots, r_{B-2}$ are held fixed. In this way we also respect the sequential nature inherent in $Q^{B}$. Hence we have:
\begin{equation*}
r^{*}_{B-1}=\underset{w_{B-1}}{argmax} \hspace{0.1cm} \underset{Q_B}{\inf} \hspace{0.1cm}\underset{w_B}{\max}\hspace{0.1cm}\underset{j}{\max} \hspace{0.1cm} \underset{j \neq i }{\min} \hspace{0.1cm} \underset{\nu \in \xi_i}{\inf} \frac{1}{B} \sum^{B-2}_{l=1}\sum_{a \in [K]}r_{l}(Q^{l})(a)D(Q_{l,a}||\nu_a)+\frac{1}{B}\sum^{B}_{l=B-1}\sum_{a \in [K]}w_l(Q^{l})(a)D(Q_{l,a}||\nu_a). 
\end{equation*}
The nonemptiness of the argmax follows from the compactness of $\Delta_K$, $\xi_i$ and the fact that $Q_B$ belongs to compact set by the finite support assumption. Now clearly the procedure to define the optimal $r^{*}_{B-2}$ to $r^{*}_1$ is similar and maybe similarly carried to get the equivalent representation of $R^{go}_{B}$. This completes the proof.
\end{proof}
\subsection{Proof of Proposition \ref{consistency_result}.}\label{proof_prop_2}
\begin{proof}
We first show that $R_{static} \leq R^{go}_B$ for any $B \in \mathbb{N}$. In the alternate representation of $R^{go}_B$ given by Proposition \ref{eq_rep_R_go_B} we choose $w_l=w$ for all $1 \leq l \leq B$. Further we have the following claim:
\begin{claim}\label{second_order_claim}
Given $N$ arbitrary sets $A_i$, $ i \in [N]$ and any collection of functions $f_i(.)$ which takes elements from $A_i$ as input, we have:
\begin{equation*}
\underset{j}{\max}\hspace{0.1cm}\underset{j \neq i}{\min} \hspace{0.1cm} \underset{a \in A_i}{\inf} f_i(a)=\underset{i \neq j}{\min} \max \bigg \{ \underset{a \in A_i}{\inf} f_i(a),\underset{a \in A_j}{\inf} f_j(a)\bigg \}. 
\end{equation*}
\end{claim}
Combining we get the inequality for any $w \in \Delta_K$ we have:
\begin{equation*}
\begin{aligned}
R^{go}_B & \stackrel{(a)} {\geq}\underbrace{\underset{Q_1}{\inf}\hspace{0.1cm}\underset{Q_2}{\inf}\hspace{0.1cm}\ldots\underset{Q_B}{\inf}}_{B \text{-times}} \underset{j}{\max}\hspace{0.1cm}\underset{j \neq i}{\min} \underset{\nu \in \xi_i}{\inf} \left( \frac{1}{B} \sum^{B}_{l=1}\sum_{a \in [K]}w(a)D(Q_{l,a}||\nu_a) \right)\\
&\stackrel{(b)}{=} \underset{Q^{B}}{\inf}\underset{i \neq j}{\min} \max \bigg \{ \underset{\nu \in \xi_i}{\inf} \frac{1}{B} \sum^{B}_{l=1}\sum_{a \in [K]}w(a)D(Q_{l,a}||\nu_a),\underset{\nu \in \xi_j}{\inf} \frac{1}{B} \sum^{B}_{l=1}\sum_{a \in [K]}w(a)D(Q_{l,a}||\nu_a) \bigg \} \\
&=\underset{i \neq j}{\min} \underset{Q^{B}}{\inf} \max \bigg \{ \underset{\nu \in \xi_i}{\inf} \frac{1}{B} \sum^{B}_{l=1}\sum_{a \in [K]}w(a)D(Q_{l,a}||\nu_a),\underset{\nu \in \xi_j}{\inf} \frac{1}{B} \sum^{B}_{l=1}\sum_{a \in [K]}w(a)D(Q_{l,a}||\nu_a) \bigg \} \\
&=\underset{i \neq j}{\min} \underset{Q^{B}}{\inf} \underset{\underset{\nu^{'} \in \xi_j}{\nu \in \xi_i}}{\inf}\max \bigg \{\frac{1}{B} \sum^{B}_{l=1}\sum_{a \in [K]}w(a)D(Q_{l,a}||\nu_a),\frac{1}{B} \sum^{B}_{l=1}\sum_{a \in [K]}w(a)D(Q_{l,a}||\nu^{'}_a) \bigg \} \\
& = \underset{i \neq j}{\min}\underset{Q^{B}}{\inf} \underset{\underset{\nu^{'} \in \xi_j}{\nu \in \xi_i}}{\inf} \underset{s \in [0,1]}{\max} \bigg \{ s \left(\frac{1}{B} \sum^{B}_{l=1}\sum_{a \in [K]}w(a)D(Q_{l,a}||\nu_a) \right)+(1-s) \left(\frac{1}{B} \sum^{B}_{l=1}\sum_{a \in [K]}w(a)D(Q_{l,a}||\nu^{'}_a) \right)\bigg\}\\
& \stackrel{(c)}{\geq} \underset{i \neq j}{\min}\underset{\underset{\nu^{'} \in \xi_j}{\nu \in \xi_i}}{\inf} \underset{s \in [0,1]}{\max} \hspace{0.1cm} \underset{Q^{B}}{\inf} \bigg \{ s \left(\frac{1}{B} \sum^{B}_{l=1}\sum_{a \in [K]}w(a)D(Q_{l,a}||\nu_a) \right)+(1-s) \left(\frac{1}{B} \sum^{B}_{l=1}\sum_{a \in [K]}w(a)D(Q_{l,a}||\nu^{'}_a) \right) \bigg\}\\
& = \underset{i \neq j}{\min}\underset{\underset{\nu^{'} \in \xi_j}{\nu \in \xi_i}}{\inf} \underset{s \in [0,1]}{\max} \hspace{0.1cm} \underset{Q^{B}}{\inf} \bigg \{\frac{1}{B} \sum^{B}_{l=1}\sum_{a \in [K]}w(a) \left(sD(Q_{l,a}||\nu_a) +(1-s)D(Q_{l,a}||\nu^{'}_a)\right) \bigg\}\\
& \stackrel{(d)}{=}\underset{i \neq j}{\min}\underset{\underset{\nu^{'} \in \xi_j}{\nu \in \xi_i}}{\inf} \underset{s \in [0,1]}{\max} \hspace{0.1cm} \underset{Q}{\inf} \sum_{a \in [K]}w(a) \left(sD(Q_{a}||\nu_a) +(1-s)D(Q_{a}||\nu^{'}_a)\right)\\
&=\underset{i \neq j}{\min}\underset{\underset{\nu^{'} \in \xi_j}{\nu \in \xi_i}}{\inf} \underset{s \in [0,1]}{\max} \hspace{0.1cm} \sum_{a \in [K]}w(a) \underset{Q_a}{\inf}\left(sD(Q_{a}||\nu_a) +(1-s)D(Q_{a}||\nu^{'}_a)\right)\\
& \stackrel{(e)}{=}\underset{i \neq j}{\min}\underset{\underset{\nu^{'} \in \xi_j}{\nu \in \xi_i}}{\inf} \underset{s \in [0,1]}{\max} \hspace{0.1cm} \sum_{a \in [K]}w(a) -\log\left( \sum_{x \in \mathcal{X}} \nu_a(x)^{s}(\nu^{'}_a(x))^{1-s}\right),
\end{aligned}
\end{equation*}
where $(a)$ is because of setting $w_l=w$ for $1 \leq l \leq B$ in \eqref{eqv_rep_R_go_B_eq}, $(b)$ is because of Claim \ref{second_order_claim}, $(c)$ is because maxmin is smaller than minmax, $(d)$ is because the optimization over $Q^{B}$ reduces to an optimization over a single empirical bandit distribution and $(e)$ is a well known optimization result in information theory (see Theorem 32 in \cite{van2014renyi}). As this result holds for all $w \in \Delta_K$, we must have then that:
\begin{equation*}
R^{go}_B \geq \underset{w \in \Delta_K}{\sup}\underset{i \neq j}{\min}\underset{\underset{\nu^{'} \in \xi_j}{\nu \in \xi_i}}{\inf} \underset{s \in [0,1]}{\max} \hspace{0.1cm} -\sum_{a \in [K]}w(a)\log\left( \sum_{x \in \mathcal{X}} \nu_a(x)^{s}(\nu^{'}_a(x))^{1-s}\right) =R_{static}.
\end{equation*}
As this is true for all $B \in \mathbb{N}$, taking limit $B \to \infty$ gives $R^{go}_\infty \geq R_{static}$ as well. Since $\underset{B \in \mathbb{N}}{\inf}R^{go}_B= R^{go}_\infty $ we also have $R^{go}_B \geq R^{go}_\infty$. Corollary 3 in \cite{komiyama2022minimax} showed that $R^{go}_B \leq R^{go}_1$.
\\
We next show that $R^{go}_1 \leq R^{ub}$. Using the representation of $R^{go}_1$ from \eqref{eqv_rep_R_go_B_eq} we get that:
\begin{equation*}
\begin{aligned}
R^{go}_1 & = \underset{Q}{\inf} \underset{w \in \Delta_K}{\sup} \underset{j}{\max}\hspace{0.1cm}\underset{j \neq i}{\min} \underset{\nu \in \xi_i}{\inf} \left( \sum_{a \in [K]}w(a)D(Q_{a}||\nu_a) \right)\\
& \stackrel{(a)}{\leq} \underset{Q}{\inf} \underset{j}{\max} \hspace{0.1cm}\underset{j \neq i}{\min} \underset{\nu \in \xi_i}{\inf}  \underset{w \in \Delta_K}{\sup}  \left( \sum_{a \in [K]}w(a)D(Q_{a}||\nu_a) \right) \\
&= \underset{Q}{\inf} \underset{j}{\max} \hspace{0.1cm}\underset{j \neq i}{\min} \underset{\nu \in \xi_i}{\inf}  \underset{a \in [K]}{\max} D(Q_{a}||\nu_a)\\
& \stackrel{(b)}{=} \underset{Q}{\inf} \max \bigg\{ \underset{\nu \in \xi_i}{\inf}  \underset{a \in [K]}{\max} D(Q_{a}||\nu_a), \underset{\nu \in \xi_j}{\inf}  \underset{a \in [K]}{\max} D(Q_{a}||\nu_a)\bigg \}\\
&=\underset{Q}{\inf} \underset{\underset{\nu^{'} \in \xi_j}{\nu \in \xi_i}}{\inf} \max  \underset{a \in [K]}{\max} \bigg\{ D(Q_{a}||\nu_a), D(Q_{a}||\nu^{'}_a)\bigg \}\\
& =  \underset{\underset{\nu^{'} \in \xi_j}{\nu \in \xi_i}}{\inf} \underset{Q}{\inf}   \underset{a \in [K]}{\max} \max \bigg\{ D(Q_{a}||\nu_a), D(Q_{a}||\nu^{'}_a)\bigg \}\\
& \stackrel{(c)}{=}\underset{\underset{\nu^{'} \in \xi_j}{\nu \in \xi_i}}{\inf}  \underset{a \in [K]}{\max} \underset{Q_a}{\inf} \max \bigg\{ D(Q_{a}||\nu_a), D(Q_{a}||\nu^{'}_a)\bigg \}\\
&\stackrel{(d)}{=}\underset{\underset{\nu^{'} \in \xi_j}{\nu \in \xi_i}}{\inf}  \underset{a \in [K]}{\max} C(\nu^{'}_a,\nu_a)\\
&=R^{ub},
\end{aligned}
\end{equation*}
where $(a)$ is because maxmin is less than minmax, $(b)$ is because of Claim \ref{second_order_claim}, $(c)$ is true because Q is tuple of distributions and hence can be interchanged with  $\underset{a \in [K]}{\max}$ and $(d)$ is an alternative definition of chernoff information (see Theorem 30 in \cite{van2014renyi}). We will show the proof of Claim \ref{second_order_claim} below and complete the proof.
\\
\\
\textbf{\underline{Proof of claim \ref{second_order_claim}}.}: Let us define $h_i:= \underset{a \in A_i}{\inf} f_i(a)$. Then we just need to show that :

\begin{equation*}
\underset{j}{\max}\hspace{0.1cm}\underset{j \neq i}{\min} \hspace{0.1cm} h_i =\underset{i \neq j}{\min} \max \bigg \{ h_i,h_j\bigg \}. 
\end{equation*}
Let us reorder the $h_i$'s such that $h_{(1)} \leq h_{(2)} \leq \dots, h_{(N)}$. Now let us interpret the LHS of the above equation. For any fixed $j \neq (1)$, we observe that $\underset{j \neq i}{\min} h_i= h_{(1)}$. However when $j=(1)$ we then have that $\underset{j \neq i}{\min} h_i= h_{(2)}$, Since $h(2) \geq h(1)$, we conclude that the LHS equals $h_{(2)}$. On the RHS it is clear the ordered pair $((1),(2))$ attains the minima of $max(h_{(1)},h_{(2)})=h_{(2)}$. Thus we have shown that both RHS and LHS are equal to $h_{(2)}$.
\end{proof}

\subsection{ Proof of Lemma \ref{Hamiltonian_formula_lemma}.}\label{Hamiltonian_formula_lemma_proof}
\begin{proof}
  The upper Hamiltonian is by definition:
\begin{equation*}
H^{+}(p)=\underset{Q}{\min}  \underset{w \in \Delta_K}{\max}  \bigg \{\sum^{m}_{i=1}\sum_{a \in [K]}w(a)D(Q_{a}||\nu^{i}_a)p_i \bigg  \}. 
\end{equation*}
Re-writing it and expanding $Q$ for clarity we have:
\begin{equation*}
 H^{+}(p)=\underset{Q=(Q_a)_{a \in [K]}}{\min}  \underset{w \in \Delta_K}{\max}  \bigg \{\sum_{a \in [K]}w(a)\sum^{m}_{i=1} D(Q_{a}||\nu^{i}_a)p_i \bigg  \}. 
\end{equation*}   
Notice nature has to respond with a $Q_a$ for each arm $a$ in response to the agents play $w$. Further the objective as rewritten is linear in  $w$. Suppose nature chooses $Q^{*}_a \in \underset{Q_a}{argmin} \sum^{m}_{i=1} D(Q_{a}||\nu^{i}_a)p_i$ for each arm $a$. Then we have by definition:
\begin{equation*}
H^{+}(p) \leq \underset{w \in \Delta_K}{\max} \bigg \{\sum_{a \in [K]}w(a)\sum^{m}_{i=1} D(Q^{*}_{a}||\nu^{i}_a)p_i \bigg  \}.
\end{equation*}
But for this choice $Q^{*}_a$ we also have for every $Q_a$, $a \in [K]$ that:
\begin{equation*}
\sum^{m}_{i=1} D(Q_{a}||\nu^{i}_a)p_i \geq \sum^{m}_{i=1} D(Q^{*}_{a}||\nu^{i}_a)p_i = \underset{Q_a}{\inf} \sum^{m}_{i=1} D(Q_{a}||\nu^{i}_a)p_i
\end{equation*}
and hence for every $w \in \Delta_K$:
\begin{equation*}
\sum_a w(a) \sum^{m}_{i=1} D(Q_{a}||\nu^{i}_a)p_i \geq   \sum_a w(a) \underset{Q_a}{\inf} \sum^{m}_{i=1} D(Q_{a}||\nu^{i}_a)p_i  
\end{equation*}
and hence:
\begin{equation*}
\underset{w \in \Delta_K}{\sup}\sum_a w(a) \sum^{m}_{i=1} D(Q_{a}||\nu^{i}_a)p_i \geq  \underset{w \in \Delta_K}{\sup}  \sum_a w(a) \underset{Q_a}{\inf} \sum^{m}_{i=1} D(Q_{a}||\nu^{i}_a)p_i.  
\end{equation*}
As this is true for each $Q_a$, we can take an infimum over the LHS to get:
\begin{equation*}
\begin{aligned}
\underset{Q}{\inf}\underset{w \in \Delta_K}{\sup}\sum_a w(a) \sum^{m}_{i=1} D(Q_{a}||\nu^{i}_a)p_i & \geq  \underset{w \in \Delta_K}{\sup}  \sum_a w(a) \underset{Q_a}{\inf} \sum^{m}_{i=1} D(Q_{a}||\nu^{i}_a)p_i\\
& = \underset{w \in \Delta_K}{\max} \bigg \{\sum_{a \in [K]}w(a)\sum^{m}_{i=1} D(Q^{*}_{a}||\nu^{i}_a)p_i \bigg  \}
\end{aligned}
\end{equation*}
Thus we have that:
\begin{equation*}
H^{+}(p)= \underset{w \in \Delta_K}{\max} \bigg \{\sum_{a \in [K]}w(a)\sum^{m}_{i=1} D(Q^{*}_{a}||\nu^{i}_a)p_i \bigg  \}.
\end{equation*}
But we note that:
\begin{equation*}
\begin{aligned}
\underset{w \in \Delta_K}{\max} \bigg \{\sum_{a \in [K]}w(a)\sum^{m}_{i=1} D(Q^{*}_{a}||\nu^{i}_a)p_i \bigg  \} & = \underset{w \in \Delta_K}{\max} \bigg \{\sum_{a \in [K]}w(a) \underset{Q_a}{\inf}\sum^{m}_{i=1} D(Q_{a}||\nu^{i}_a)p_i \bigg  \}\\
&=\underset{w \in \Delta_K}{\max} \underset{Q}{\inf} \bigg \{\sum_{a \in [K]}w(a) \sum^{m}_{i=1} D(Q_{a}||\nu^{i}_a)p_i \bigg  \}\\
&=H^{-}(p).
\end{aligned}
\end{equation*}
Thus we have shown that $H^{+}(p)=H^{-}(p)$ for every $p$. From the above we also see that:
\begin{equation*}
\begin{aligned}
H^{+}(p)=H^{-}(p) & =\underset{w \in \Delta_K}{\max} \bigg \{\sum_{a \in [K]}w(a)\sum^{m}_{i=1} D(Q^{*}_{a}||\nu^{i}_a)p_i \bigg  \}\\
& = \max_a  \bigg \{ \sum^{m}_{i=1} D(Q^{*}_{a}||\nu^{i}_a)p_i  \bigg \}.
\end{aligned}
\end{equation*}
Hence to get an exact expression of the Hamiltonians we next study the following optimization problem in detail:
\begin{equation*}
\underset{Q_a}{\inf} \hspace{0.2cm} \sum^{m}_{i=1} D(Q_{a}||\nu^{i}_a)p_i.
\end{equation*}
To that end we re-write the objective as follows:
\begin{equation}\label{obj_alt_form}
\begin{aligned}
\sum^{m}_{i=1} D(Q_{a}||\nu^{i}_a)p_i &= \sum^{m}_{i=1} p_i\sum_{x \in \mathcal{X}} Q_a(x) \log \left( \frac{Q_a(x)}{\nu^{i}_a(x)} \right)\\
&= \sum_{x \in \mathcal{X}} \sum^{m}_{i=1} p_i Q_a(x) \log \left( \frac{Q_a(x)}{\nu^{i}_a(x)} \right)\\
& = \sum_{x \in \mathcal{X}} Q_a(x) \sum^{m}_{i=1} p_i \log \left( \frac{Q_a(x)}{\nu^{i}_a(x)} \right).
\end{aligned}
\end{equation}
Define $P=\sum_i p_i$. If $P \neq 0$, we can normalize eqn \eqref{obj_alt_form} by $P$ to get:
\begin{equation*}
\begin{aligned}
\sum^{m}_{i=1} D(Q_{a}||\nu^{i}_a)p_i&= \sum_{x \in \mathcal{X}} Q_a(x) \sum^{m}_{i=1} p_i \log \left( \frac{Q_a(x)}{\nu^{i}_a(x)} \right)\\
&=P\sum_{x \in \mathcal{X}} Q_a(x) \sum^{m}_{i=1} \frac{p_i}{P} \log \left( \frac{Q_a(x)}{\nu^{i}_a(x)} \right)\\
&=P\sum_{x \in \mathcal{X}} Q_a(x) \log \left( \frac{Q_a(x)}{\prod^{m}_{i=1}(\nu^{i}_a(x))^{p_i/P}} \right).
\end{aligned}     
\end{equation*}
We then normalize inside the logarithm using the partition term $Z=\sum_{x \in \mathcal{X}}\prod^{m}_{i=1}(\nu^{i}_a(x))^{p_i/P}$ to get:
\begin{equation*}
\begin{aligned}
\sum^{m}_{i=1} D(Q_{a}||\nu^{i}_a)p_i&= P\sum_{x \in \mathcal{X}} Q_a(x) \log \left( \frac{Q_a(x)Z}{\prod^{m}_{i=1}(\nu^{i}_a(x))^{p_i/P}Z} \right)\\
&=P\sum_{x \in \mathcal{X}} Q_a(x) \log \left( \frac{Q_a(x)Z}{\prod^{m}_{i=1}(\nu^{i}_a(x))^{p_i/P}}\right)-P \log(Z). 
\end{aligned}
\end{equation*}
Now observing that $\tilde{Q}=\left(\frac{\prod^{m}_{i=1}(\nu^{i}_a(x))^{p_i/P}}{Z}\right)_{x \in \mathcal{X}}$ is a probability distribution over $\mathcal{X}$ we have that:
\begin{equation}\label{non_zero_form}
 \sum^{m}_{i=1} D(Q_{a}||\nu^{i}_a)p_i= PD(Q_a||\tilde{Q})-P \log \left( \sum_{x \in \mathcal{X}}\prod^{m}_{i=1}(\nu^{i}_a(x))^{p_i/P}\right).   
\end{equation}
Let us now study the three different cases: (i) $P>0$, (ii) $P<0$ and (iii) $P=0$.\\

\textbf{Case 1 - $P>0$: } 
As $P>0$, from eqn \eqref{non_zero_form} the minimization over $Q_a$ is achieved when $Q^{*}_a=\tilde{Q}$. Thus whenever $P>0$ we have that:
\begin{equation*}
\underset{Q_a}{\inf} \hspace{0.2cm} \sum_i p_i D(Q_a||\nu^{i}_a)=-P \log \left( \sum_{x \in \mathcal{X}}\prod^{m}_{i=1}(\nu^{i}_a(x))^{p_i/P}\right).
\end{equation*}
\textbf{Case 2 - $P<0$:}  As $P<0$, from eqn \eqref{non_zero_form} we have that minimization of the objective obtained by maximizing $D(Q_a||\tilde{Q})$. As KL divergence is a strictly convex function the maxima is attained at $\delta$ distributions over $\mathcal{X}$. Hence we have:
\begin{equation*}
\begin{aligned}
\underset{Q_a}{\inf} \hspace{0.2cm} \sum_i p_i D(Q_a||\nu^{i}_a)&= P \max_{Q_a} \{D(Q_a||\tilde{Q})\}-P \log \left( \sum_{x \in \mathcal{X}}\prod^{m}_{i=1}(\nu^{i}_a(x))^{p_i/P}\right)\\
&=P \max_{x \in \mathcal{X}}\log \left( \frac{Z}{\prod^{m}_{i=1}(\nu^{i}_a(x))^{p_i/P}} \right)-P \log(Z)\\
&=P \log(Z)+ P \underset{x \in \mathcal{X}}{\max}-\log \left(\prod^{m}_{i=1}(\nu^{i}_a(x))^{p_i/P} \right)-P \log(Z)\\
&=\underset{x \in \mathcal{X}}{\min}-P\log \left(\prod^{m}_{i=1}(\nu^{i}_a(x))^{p_i/P} \right)\\
&=\underset{x \in \mathcal{X}}{\min}-\sum^{m}_{i=1}p_i \log \left(\nu^{i}_a(x)\right).
\end{aligned}
\end{equation*}
\textbf{Case 3 - $P=0$:} As $P=0$, the eqn. \eqref{non_zero_form} is not valid. Instead, we will work with eqn. \eqref{obj_alt_form}. Thus, we have:
\begin{equation*}
\begin{aligned}
\sum^{m}_{i=1} D(Q_{a}||\nu^{i}_a)p_i &=\sum_{x \in \mathcal{X}} Q_a(x) \sum^{m}_{i=1} p_i \log \left( \frac{Q_a(x)}{\nu^{i}_a(x)} \right)\\
&=\sum_{x \in \mathcal{X}} Q_a(x)  \log \left( \frac{Q_a(x)^{P}}{\prod^{m}_{i=1}(\nu^{i}_a(x))^{p_i}} \right)\\
&= \sum_{x \in \mathcal{X}} Q_a(x)  \log \left( \frac{1}{\prod^{m}_{i=1}(\nu^{i}_a(x))^{p_i}} \right),
\end{aligned}
\end{equation*}
where last equality is because $P=0$. The final expression is also linear in $Q_a$ and hence we have that:
\begin{equation*}
\begin{aligned}
\underset{Q_a}{\inf} \hspace{0.2cm} \sum_i p_i D(Q_a||\nu^{i}_a)&= \underset{Q_a}{\inf}\sum_{x \in \mathcal{X}} Q_a(x)  \log \left( \frac{1}{\prod^{m}_{i=1}(\nu^{i}_a(x))^{p_i}} \right)\\
&= \underset{x \in \mathcal{X}}{\min} \log \left( \frac{1}{\prod^{m}_{i=1}(\nu^{i}_a(x))^{p_i}} \right)\\
&=\underset{x \in \mathcal{X}}{\min} -\sum^{m}_{i=1} p_i \log(\nu^{i}_a(x)). 
\end{aligned}
\end{equation*}
Next, we show the Lipschitzness of the Hamiltonian function $H(p)$. We observe that for any fixed $a$, the function
\begin{equation*}
H_a(p)=\underset{Q_a}{\inf}\sum_i p_i D(Q_a||\nu^{i}_a)   
\end{equation*} 
is a concave function in $p$ over $\mathbb{R}^{m}$ and hence continuous over $\mathbb{R}^{m}$. Further, we have that:
\begin{equation*}
H(p)=\max_a H_a(p),
\end{equation*}
is maximum of finite number of continuous concave functions and hence is also continuous over $\mathbb{R}^{m}$. We have shown that :
\begin{equation*}
H_a(p)=\begin{cases}
- (\sum_i p_i) \log \left ( \sum_{x \in \mathcal{X}} \prod^{m}_{i=1}(\nu^{i}_a(x))^{\frac{p_i}{\sum_i p_i}}\right), & \text{ if } \sum_i p_i>0\\
\underset{x \in \mathcal{X}}{\min}- \sum^{m}_{i=1}p_i \log( \nu^{i}_a(x)), & \text { if } \sum_{i} p_i \leq 0.
\end{cases}
\end{equation*}
In the case $\sum_i p_i \leq 0$, we observe that $H_a(p)$ is the minimum of finite number (since $|\mathcal{X}|< \infty$) of linear functions in $p$ and hence is going to be Lipschitz in $p$ over the domain $\mathcal{D}_{\leq 0}= \{ p : \sum_i p_i \leq 0 \}$. It is straightforward to calculate the lipschitz constant by invoking Lemma \ref{finite_lipschitz_lemma}. Since $|\log( \nu^{i}_a(x))| \leq \log(\frac{1}{\epsilon})$ we have that the Lipschitz constant in the domain $\mathcal{D}_{\leq 0}$ is $\sqrt{m}\log(\frac{1}{\epsilon})$. \\

In the case $\sum_i p_i >0$, we note that $H_a(p)$ is differentiable. It is useful to define $s_p(x)=\prod^{m}_{i=1}(\nu^{i}_a(x))^{\frac{p_i}{\sum_i p_i}}$ and $s_p=\sum_{x \in \mathcal{X}}s_p(x)$. So we differentiate wrt to $p_j$:
\begin{equation*}
\begin{aligned}
\frac{\partial H_a(p)}{\partial p_j}&=- \log(s_p)-\frac{(\sum_i p_i)}{s_p} \left( \sum^{m}_{i=1}\sum_{x \in \mathcal{X}} \left (\prod^{m}_{i=1}(\nu^{i}_a(x))^{\frac{p_i}{\sum_i p_i}} \right) \log (\nu^{i}_a(x)) \frac{\partial \left(\frac{p_i}{\sum_i p_i} \right) }{ \partial p_j}\right)\\
&=-\log(s_p)-\frac{(\sum_i p_i)}{s_p} \left( \sum_{x \in \mathcal{X}} s_p(x) \sum^{m}_{i=1} \log (\nu^{i}_a(x)) \frac{\partial \left(\frac{p_i}{\sum_i p_i} \right) }{ \partial p_j}\right)\\
&=-\log(s_p)-\frac{(\sum_i p_i)}{s_p} \left( \sum_{x \in \mathcal{X}} s_p(x) \sum^{m}_{i=1} \log (\nu^{i}_a(x)) \left( \frac{\delta_{ij}(\sum_i p_i)-p_i}{(\sum_i p_i)^2}\right)\right)\\
&=-\log(s_p)-\frac{(\sum_i p_i)}{s_p} \left( \sum_{x \in \mathcal{X}} s_p(x) \left( \frac{\log (\nu^{j}_a(x))}{(\sum_i p_i)} -\sum^{m}_{i=1} \frac{p_i \log (\nu^{i}_a(x))}{(\sum_i p_i)^2} \right)\right)\\
&=-\log(s_p)-\sum_{x \in \mathcal{X}}\frac{s_p(x)}{s_p}\log( \nu^{j}_a(x))+\sum_{x \in \mathcal{X}}\frac{s_p(x)}{s_p}\sum^{m}_{i=1}\frac{p_i}{(\sum_i p_i)}\log ( \nu^{i}_a(x))\\
&=-\log(s_p)-\sum_{x \in \mathcal{X}}\frac{s_p(x)}{s_p}\log( \nu^{j}_a(x))+\sum_{x \in \mathcal{X}}\frac{s_p(x)}{s_p} \log(s_p(x)).
\end{aligned}
\end{equation*}
We observe that $s_p(x) \geq 0, \forall x \in \mathcal{X}$ and that $\sum_{x \in \mathcal{X}} \frac{s_p(x)}{s_p}=1$. Denote this probability distribution as $S(p)$. Hence we have that:
\begin{equation*}
\begin{aligned}
\frac{\partial H_a(p)}{\partial p_j}&=\sum_{x \in \mathcal{X}}\frac{s_p(x)}{s_p} \log \left(\frac{s_p(x)/s_p}{\nu^{j}_a(x)} \right)\\
&=D(S(p)||\nu^{j}_a),
\end{aligned}
\end{equation*}
Now under our assumptions we have that $D(S(p)|| \nu^{j}_a) \leq \log \left( \frac{1}{\epsilon} \right)$. This immediately implies $||\nabla H_a || \leq \sqrt{m} \log \left( \frac{1}{\epsilon} \right) $ and is the lipschitz constant in the domain $\mathcal{D}_{>0}=\{ p : \sum_i p_i >0\}$.\\

Now we have found the Lipschitz constant of $H_a(p)$ in the two disjoint domains $\mathcal{D}_{\leq 0}$ and $\mathcal{D}_{>0}$. Further, we have argued that $H_a(p)$ is continuous throughout $\mathbb{R}^{m}$. Thus, we apply Lemma \ref{disjoint_lipschitz_lemma} to conclude that $H_a(p)$ is lipschitz throughout $\mathbb{R}^{m}$ and has a Lipschitz constant of $\sqrt{m} \log \left( \frac{1}{\epsilon} \right)$.\\

Since we have $H(p)=\max_a H_a(p)$ we can invoke Lemma \ref{finite_lipschitz_lemma} to conclude that $H(p)$ is also Lipschitz with Lipschitz constant $\sqrt{m} \log \left( \frac{1}{\epsilon} \right)$. \\

Note that as $H_a(P)$ when $p \in \mathcal{D}_{\leq 0}$ is expressed as a $\min_{x \in \mathcal{X}}$ with each co-efficient of $p_i$ being positive ($-\log(\nu^{i}_a(x))> 0$ under our assumptions) we conclude that $H_a(p)$ is strictly increasing in each co-ordinate $p_j$ for $p \in \mathcal{D}_{\leq 0}$.This concludes the proof of the lemma. In the case $p \in \mathcal{D}_{>0}$, as directly calculated above, $\frac{\partial H_a(p)}{\partial p_j} \geq 0$ and this implies that in $\mathcal{D}_{>0}$, $H_a(p)$ is non-decreasing in $p_j$. Thus $H_a(p)$ is non-decreasing over the entire domain for each $a$. From this the non-decreasing nature of $H(p)$ follows. This concludes the proof of the lemma.
\end{proof}
\begin{lemma}\label{finite_lipschitz_lemma}
Consider a finite index set $I$, indexing a collection of real valued lipschitz functions $f_i$ (we assume all $f_i$ are defined over a common domain $\mathcal{D} \subseteq \mathbb{R}^{n}$), $i \in I$ with respective lipschitz constants $L_i$. Then $g := \max_{i \in I} f_i$ and $h= \min_{i \in I} f_i$ are lipschitz with Lipschitz constant $\max_{i \in I} L_i$.
\end{lemma}
\begin{proof}
From the Lipschitzness of $f_i$ we have that $\forall i \in I$ and any pair $x,y \in \mathcal{D}$:
\begin{equation*}
f_i(x) \leq f_i(y)+L_i ||x-y|| \leq f_i(y)+ (\max_i L_i)||x-y||. 
\end{equation*}
This implies by taking $\max$ amd $\min$ over $I$ on the above inequality that:
\begin{equation*}
\begin{aligned}
& g(x) \leq g(y) + (\max_{i \in I} L_i) ||x-y|| \\
& h(x) \leq h(y) + (\max_{i \in I} L_i) ||x-y||.
\end{aligned}
\end{equation*}
As the role of $x$ and $y$ can be interchanged in the above argument, we conclude that for all $x,y \in \mathcal{D}$:
\begin{equation*}
\begin{aligned}
& |g(x)-g(y)| \leq (\max_{i \in I} L_i) ||x-y||\\
& |h(x)-h(y)| \leq (\max_{i \in I} L_i) ||x-y||.
\end{aligned}
\end{equation*}
This proves the claim.
\end{proof}
\begin{lemma}\label{disjoint_lipschitz_lemma}
Suppose we have two lipschitz functions $f_A$ and $f_{A^c}$ defined on a set $A \subsetneq \mathbb{R}^{n}$ and its complement $A^{c} \subsetneq \mathbb{R}^{n}$ with respective Lipschitz constants $L_A$ and $L_{A^c}$. Define the glued up function:
\begin{equation*}
h(x)= \begin{cases}
 f_A(x), & x \in A,\\
 f_{A^{c}}(x), & x \in A^c.
\end{cases}
\end{equation*}
Suppose further that $h$ is continuous over $\mathbb{R}^{n}$, then $h$ is Lipschitz continuous with Lipschitz constant $\max \{ L_A, L_{A^{c}} \}$.
\end{lemma}
\begin{proof}
 Clearly when both $x,y \in A$ or $x,y \in A^{c}$ the following statement holds by Lipschitzness of $f_A$ and $f_{A^{c}}$:
\begin{equation*}
|h(x)-h(y)| \leq \max \{ L_A, L_{A^{c}} \} ||x - y||.
\end{equation*}
So we need to analyse only the case $x \in A$ and $y \in A^{c}$. Consider the closed straight line segment $S$ between $x$  and $y$ ($x,y$ are included in S). There exists atleast one point $s \in \partial A \cap S$. Now because $s$ lies in the line segment $S$ between $x$ and $y$ we have that:
\begin{equation*}
||x-y||=||x-s||+||y-s||.
\end{equation*}
Assume that $s \in A$ (the case $s \in A^{c}$ can be analysed similarly). Consider a sequence $y_r \in A^{c}, \forall r \in \mathbb{N}$  such that $\lim_{r \to \infty} y_r=s$ (such a sequence exist because $s \in \partial A$), then we have for each $r \in \mathbb{N}$:
\begin{equation*}
\begin{aligned}
|h(x)-h(y)| & \leq |h(x)-h(s)|+|h(s)-h(y_r)|+|h(y_r)-h(y)|\\
& \leq |h(x)-h(s)|+|h(s)-h(y_r)|+L_{A^{c}}||y_r-y||.
\end{aligned}
\end{equation*}
Taking the limit $r \to \infty$ and using the continuity of $h$ and the norm function $d_y(z)=||z-y||$ we have that:
\begin{equation*}
|h(x)-h(y)| \leq |h(x)-h(s)|+L_{A^c}||s-y||.
\end{equation*}
As $s,x \in A$ using the lipschitzness of $f_A$ we have that:
\begin{equation*}
\begin{aligned}
 |h(x)-h(y)| & \leq L_{A}||x-s||+L_{A^c}||s-y|| \\
 & \leq \max \{L_A,L_{A^c} \} (||x-s||+||s-y||)\\
 & = \max \{ L_A, L_{A^c} \} ||x-y||.   
\end{aligned}
\end{equation*}
This shows the Lipschitzness of $h$.
\end{proof}
\subsection{Proof of Proposition \ref{alt_R_go_infty_ub}.}\label{alt_R_go_infty_ub_proof}
\begin{proof}
Fix a $\lambda \in \Delta_S$. Consider the the function :
\begin{equation*}
    g_{\lambda}(x)=\sum_{(i,j) \in S} \lambda_{i,j} \max \{x_i ,x_j\}.
\end{equation*}
We observe that $g(x) \leq g_{\lambda}(x)$. Further, we note that this is a support function of set $A_{\lambda}$ :
\begin{equation*}
A_{\lambda}=\sum_{(i,j) \in S} \lambda_{i,j} \Delta_{\{i,j\}}
\end{equation*}
where $\Delta_{\{i,j\}}$ is viewed as subsets of $\Delta_m$. That is $g_{\lambda}$ is the support function of the minkowski sum of $\Delta_{\{i,j\}}$ simplexes. As $g_{\lambda}$ is convex and lipschitz, we can apply the comparison theorem (Proposition 1.1 \cite{crandal1984two}) and the Hopf formula (Theorem 3.1 \cite{bardi1984hopf}) to get that :
\begin{equation*}
R^{go}_{\infty} \leq \underset{\beta}{\sup} \hspace{0.2cm} \underset{a} {\max} -\log \left(\sum_{x \in \mathcal{X}} \prod^{m}_{k=1} (\nu^{k}_a(x))^{P_k(\lambda,\beta)} \right).
\end{equation*}
As the choice of $\lambda$ was arbitrary the result follows.
\end{proof}
\section{Proofs of results in Section \ref{sec:numerical_pde_scheme}.}
\subsection{Proof of lemma \ref{viscosity_properties}.}\label{proof_viscosity_properties}
\begin{proof}
(1) is Theorem 3.4.1 in \cite{krasovskij1988game} and the equivalence between viscosity solution and the notion minimax solution to Hamilton Jacobi PDE defined in \cite{krasovskij1988game}.\\
(2) uniqueness and existence is guaranteed by Theorem VI.1 in \cite{crandall1987remarks}.\\
(3) Consider the following change of variables :
\begin{equation*}
\begin{aligned}
   & r=t+s(1-t)\\
   & \frac{x(r)}{(1-t)}=\tilde{x}(s),
\end{aligned}
\end{equation*}
with $s \in [0,1]$ and hence $r \in [t,1]$ and the initial condition $x(t)=x$ and $\tilde{x}(0)=\frac{x}{1-t}$. Now consider the following derivative:
\begin{equation*}
\begin{aligned}
\frac{d \tilde{x}(s)}{d s} &= \frac{1}{(1-t)} \frac{d x(r) }{d s}\\
&=\frac{1}{(1-t)}\frac{dx (r)}{dr} \frac{dr}{ds}\\
&=\frac{d x(r)}{d r}.
\end{aligned}
\end{equation*}
Thus the dynamics for the game in the new co-ordinates $(s,\tilde{x}(s))$ is the same as the old co-ordinates $(r,x(r))$. In the new co-ordinates it begin with $s=0$ and an initial condition of $\frac{x}{1-t}$. For any trajectory the final payoff in the new co-ordinates will $g(\tilde{x}(1))=g \left (\frac{x(1)}{1-t} \right)=\frac{1}{1-t}g(x(1))$. From this relation the result immediately follows.
\end{proof}
\subsection{Proof of lemma \ref{domain_of_dependence}.}\label{proof_domain_of_dependence}
\begin{proof}
We will use the fact that :
\begin{equation*}
0 \leq \dot x_{t,i} \leq \log \left ( \frac{1}{\epsilon} \right).
\end{equation*}
This implies that from any starting point $x_0$, we have that each co-ordinate of $x_{t,i}$, for $t \in [0,1]$, is contained in the hypercube $[x_{0,1},x_{0,1}+\log \left ( \frac{1}{\epsilon} \right) t ] \times \dots \times [x_{0,m},x_{0,m}+\log \left ( \frac{1}{\epsilon} \right) t ]$. Now, we are given that when $x_0=0$, we have that over the hypercube $[0,\log \left ( \frac{1}{\epsilon} \right)]^{m}$, $g(x)=g^{'}(x)$. Thus, the set of all possible reachable states from $x_0=0$ in the time interval $[0,1]$ is contained in the hypercube $[0,\log \left ( \frac{1}{\epsilon} \right)]^{m}$. Similarly one can argue that for an initial condition $(x,t)$ satisfying $0 \leq x_i \leq \log \left( \frac{1}{\epsilon} \right)t$ for any $t \in [0,1)$ the reachable set of states at terminal time $t=1$ is included within the hypercube $[0,\log \left ( \frac{1}{\epsilon} \right)]^{m}$.\\
From lemma \ref{DP_value_formulation} eq. \eqref{terminal_value_representation} and the fact that both PDE formulations (mentioned in the statement of lemma \ref{domain_of_dependence})  have the same dynamics (they differ only in the terminal value function $g$ and $g^{'}$) we have that:
\begin{equation}\label{value_func_equality_in hypercube}
\begin{aligned}
& V(x,t)= \underset{w \in \mathcal{A}(t)} {\sup} \underset{Q_{[t,1]} \in \mathcal{Q}(t)}{\inf} g(x(1))\\
& V^{'}(x,t)=\underset{w \in \mathcal{A}(t)} {\sup} \underset{Q_{[t,1]} \in \mathcal{Q}(t)}{\inf} g^{'}(x(1)).
\end{aligned}
\end{equation}
Now, since as the $g=g^{'}$ whenever $x(1) \in [0,\log \left ( \frac{1}{\epsilon} \right)]^{m}$ and we have argued that the set of reachable terminal states $x(1)$ are in $[0,\log \left ( \frac{1}{\epsilon} \right)]^{m}$ whenever
$0 \leq x_i \leq \log \left( \frac{1}{\epsilon} \right)t$ for any $t \in [0,1)$ and all $i \in [m]$, we conclude that $V(x,t)=V^{'}(x,t)$ from eq. \eqref{value_func_equality_in hypercube}.
\end{proof}
\subsection{Proof of lemma \ref{alt_g_properties}.}\label{proof_alt_g_properties}
\begin{proof}
The fact that $g^{'}$ is bounded, continuous and has a compact support $C_{a/2}$ follows directly from the construction of $g^{'}$. Further if $x \in C$, then $\prod^{(2)}_{C}(x)=x$ and $d_{\infty}(x.C)=0$. This implies that $g^{'}(x)=g(x)$ whenever $x \in C$. We now show the Lipschitzness.\\

We observe that $\prod^{(2)}_C(.)$ is a lipschitz function with lipschitz constant 1 which follows from the non-expansiveness of the euclidean projection onto a closed convex set. Further $g(.)$ was also a lipschitz function with lipschitz constant 1. Their composition is therefore also lipschitz. The distance metric $d_{\infty}(.,C)$ is lipschitz in $l_{\infty}$ metric and hence also lipschitz in $l_2$ distance. Consequently so is $1-\frac{2 d_{\infty}(.,C)}{a}$. The product of two lipschitz functions over a compact set is also lipschitz over the compact set. This implies that over the set $C_{a/2}$ we have that $g \left(\prod^{(2)}_C(.) \right) \left(1-\frac{2d_{\infty}(.,C)}{a}\right)$ is also Lipschitz. Clearly the zero function is lipschitz over the set $\mathbb{R}^{m} \setminus C_{a/2}$. And clearly $g^{'}$ is continuous over the entire euclidean space $\mathbb{R}^{m}$. We can then apply Lemma \ref{disjoint_lipschitz_lemma} to conclude that $g^{'}$ is lipschitz.\\

Consider any $x \notin [-3a/2,3a/2]^{m}$. Using this as the starting point we observe that all reachable positions from this point is the set $[x,x+a]^{m}$ (see similar argument in the proof of lemma \ref{domain_of_dependence}). But we have that :
\begin{equation*}
[x,x+a]^{m} \cap C_{a/2}= \phi.
\end{equation*}
If not, there exists $z \in C_{a/2}$ which implies:
\begin{equation*}
-a/2 \leq z_i \leq 3a/2
\end{equation*}
for each $i \in [m]$ but also $z \in [x,x+a]^{m}$ that is:
\begin{equation*}
x_i \leq z_i \leq x_i+a
\end{equation*}
for each $i \in [m]$. But as $x \notin [-3a/2,3a/2]^{m}$, there must exist some $j \in [m]$ such that $x_j<-3a/2$ or $x_j>3a/2$. If $x_j<-3a/2$, then from above this implies $z_j< a/2$ contradicting that $a/2 \leq z_i$ for all $i \in [m]$. If $x_j>3a/2$, then $3a/2<z_j$ and this contradicts the fact that $z_i \leq 3a/2$ for all $i \in [m]$. Thus we have that $[x,x+a]^{m} \cap C_{a/2}= \phi$. Using the representation eq \eqref{terminal_value_representation} for $V^{'}(x,t)$, the fact that the reachable set of $x \notin [-3a/2,3a/2]^{m}$ doesn't intersect with the support $C_{a/2}$ of $g^{'}$ we conclude that $V^{'}(x,t)=0$ for such $x$ for all $t \in [0,1]$. This concludes the proof of the lemma.
\end{proof}
\subsection{Proof of lemma \ref{explicit_scheme_mono}.}\label{proof_explicit_scheme_mono}
\begin{proof}
For every $j \in [m]$, define for each $j \in [m]$:
\begin{equation*}
i_{j}=(i_1,\dots, i_{j+1}, \dots,i_m).
\end{equation*}
Further, denote the RHS of eq. \eqref{explicit_upwind} as:
\begin{equation}\label{RHS_upwind}
G(V^{n}_i, V^{n}_{i_1}, \dots, V^{n}_{i_m})=V^{n}_i-\Delta t \tilde{H} \left( \left(\frac{\Delta_j V^{n}_{i}}{\Delta h} \right)_{j \in [m]} \right).
\end{equation}
Note that terms like $V^{n}_{i_j}$ are implicitly present in the difference term $\Delta_j V^{n}_{i}$. To show the upwind scheme is monotone, we need to show that $G(.)$ is non-decreasing in each of the aforementioned terms.\\

First we deal with terms like $V^{n}_{i_j}$. From monotonicity of hamiltonian proved in lemma \ref{Hamiltonian_formula_lemma} and the fact that we are using the hamiltonian of the reversed time PDE \ref{time_revr_pde} implies that $\tilde{H}$ is co-ordinate wise non-increasing. From this it immediately follows that the $G(.)$ is non-decreasing in each $V^{n}_{i_j}$ for each $j \in [m]$.\\

Next, we prove the non-decreasing behavior of $G(.)$ in $V^{n}_i$. Suppose, we perturb $V^{n}_i$ to $V^{n}_i+\delta$, with $\delta>0$. Then we have that:
\begin{equation*}
\begin{aligned}
G(V^{n}_i+\delta, V^{n}_{i_1}, \dots, V^{n}_{i_m})-G(V^{n}_i, V^{n}_{i_1}, \dots, V^{n}_{i_m})&=\delta-\Delta t  \left( \tilde{H} \left( \left(\frac{\Delta_j V^{n}_{i}}{\Delta h} -\frac{\delta}{\Delta h} \right)_{j \in [m]} \right)-\tilde{H} \left( \left(\frac{\Delta_j V^{n}_{i}}{\Delta h} \right)_{j \in [m]} \right)\right)\\
& \geq \delta -\Delta t \bigg | \bigg| \left (\frac{\delta}{\Delta h} \right)_{j \in [m]} \sqrt{m} \log \left( \frac{1}{\epsilon}\right) \bigg | \bigg |\\
&= \delta \left( 1-\frac{m \Delta t}{\Delta h} \log \left( \frac{1}{\epsilon}\right) \right)\\
& \geq 0,
\end{aligned}
\end{equation*}
where the first inequality follows from the Lipschitzness of the hamiltonian proven in lemma \ref{Hamiltonian_formula_lemma}. The second inequality comes from the stability conditions eq. \eqref{stability_criteria} imposed in the hypothesis of the lemma. This shows that $G(.)$ is non-decreasing in $V^{n}_i$ as well. This shows the requisite monotonicity of the upwind scheme \eqref{explicit_upwind}.
\end{proof}
\subsection{Computational effort to compute $R^{go}_{\infty}$ using various representations.}\label{compute_effort_R_go_infty}
Suppose we are interested in solving the repeated Stackelberg game in eq. \eqref{eqv_rep_R_go_B_eq} for $B$ large enough. This can be re-written as:
\begin{equation*}
R^{go}_B=\underbrace{\underset{Q_1}{\inf} \hspace{0.1cm}\underset{w_1}{\sup}\hspace{0.1cm}\underset{Q_2}{\inf}\hspace{0.1cm}\underset{w_2}{\sup}\ldots\underset{Q_B}{\inf}\hspace{0.1cm}\underset{w_B}{\sup}}_{B \text{-times}} \underset{j}{\max}\hspace{0.1cm}\underset{j \neq i}{\min} \underset{\nu \in \xi_i}{\inf} \left( \frac{1}{B} \sum^{B}_{l=1}\sum_{a \in [K]}w_{l}(Q^{l})(a)D(Q_{l,a}||\nu_a) \right)    
\end{equation*}
which can be re-written as :
\begin{equation*}
 R^{go}_B=\underbrace{\underset{Q_1}{\inf} \hspace{0.1cm}\underset{w_1}{\sup}\hspace{0.1cm}\underset{Q_2}{\inf}\hspace{0.1cm}\underset{w_2}{\sup}\ldots\underset{Q_B}{\inf}\hspace{0.1cm}\underset{w_B}{\sup}}_{B \text{-times}} \underset{j}{\max}\hspace{0.1cm} g(w^{B},Q^{B}),
\end{equation*}
where $g$ is the terminal value function defined in eq. \eqref{term_cost}, and $w^{B},Q^{B}$ refers to the controls applied by agent and nature respectively, in the B-times repeated game. Suppose we discretize the control spaces $\Delta_K$ and $(\Delta_{\mathcal{X}})^{K}$, then for the very last stage $w_B$ is always chosen by the agent to maximize the final cost $g(w^{B},Q^{B})$. If they brute force this over the discretized grid it requires $O(me^{K})$ effort (since we have to evaluate over a grid exponential in dimension of control set of the agent). Now when choosing $Q_B$ nature has to be mindful of the agents last response $w_B$ and choose $Q_B$ such that it minimizes the terminal cost. This would require $O(me^{K})$ effort for each fixed $Q_B$ and hence overall this step will require $O(me^{K |\mathcal{X}|+K})$. Now, this procedure has to be repeated $B$- times and would entail an overall effort of $O(me^{B K |\mathcal{X}|})$. Note that there is no good way of solving such repeated Stackelberg games and such exponential computational effort is typical. Thus even when $m,K,\mathcal{|X|}$ is small, if we evaluate for moderate $B \sim 10$ this is computationally impossible.\\
Now, let us analyze the computational effort to compute $R^{go}_{\infty}$ through the finite difference scheme in section \ref{FD_scheme}. We have a grid which is exponential in $m$ (as that is the state space of the differential game). Now, we have to evaluate the Hamiltonian (eq. \eqref{Hamiltonian_formula}) which is linear in $K$ and $\mathcal{|X|}$ at each grid point. Then, we have to move forward in time using eq. \eqref{explicit_upwind}). Thus the overall computational effort of this step will be $O(K |\mathcal{X}|\frac{1}{(\Delta t)^{m}})$. As this has to be done for $1/\Delta t$ times we get the overall computational effort to be $O \left( \frac{K }{(\Delta t)^{m+1}}\right)$. Now with $B= \frac{1}{\Delta t}$ we get that the overall effort is $O(K |\mathcal{X}|B^{m+1})$. Whenever $m$ is small this is much better than solving the repeated game structure.

\subsection{Proof of lemma \ref{lemma_grid_scheme_consistency}.}\label{proof_lemma_grid_scheme_consistency}
We will show the stronger result:
\begin{equation*}
B_{0}(\sqrt{m}a (t_l+\Delta t)+\kappa t_l^{2}) \subseteq D^{conv}_{l+1}, \hspace{0.2cm} \forall l \in \{0,1,2, \dots,B-1\}.
\end{equation*}
The result follows since for any $y \in B_x(\sqrt{m}a \Delta t)$ with $x \in D_{l}$ we have that:
\begin{equation*}
||y|| \leq ||y-x||+||x|| \leq \sqrt{m}a \Delta t+ \sqrt{m}a t_l+\kappa t^2_l.
\end{equation*}
We will show the result by studying the properties of the support functionals $\sigma_{B_{0}(\sqrt{m}a (t_l+\Delta t)+\kappa t_l^{2})}$ and $\sigma_{D^{conv}_{l+1}}$. To this end consider any $y \in \mathbb{R}^{m}$. Then we have that:
\begin{equation*}
x^{*}(y):= \underset{x \in B_{0}(\sqrt{m}a (t_l+\Delta t)+\kappa t_l^{2}) }{argmax} \langle x, y \rangle=(\sqrt{m}a (t_l+\Delta t)+\kappa t_l^{2}) \frac{y}{||y||}.
\end{equation*}
Now clearly $x^{*}(y) \in H_{\sqrt{m}a+\kappa}$. This implies that there exist $x^{(n)} \in GR_{\Delta h}$ such that $x^{*}(y) \in [x^{(n)}_1,x^{(n)}_1+\Delta h] \times \dots \times [x^{(n)}_m,x^{(n)}_m+\Delta h]=:H_y$. Now we have that:
\begin{equation*}
||z-x^{*}(y)|| \leq \sqrt{m} \Delta h, \hspace{0.2cm} \forall z \in H_y.
\end{equation*}
Consequently, we have that, $\forall z \in H_y$:
\begin{equation*}
||z|| \leq ||z-x^{*}(y)|| +||x^{*}(y)||\leq  \sqrt{m} \Delta h+\sqrt{m}a (t_l+\Delta t)+\kappa t_l^{2}.
\end{equation*}
Using the condition $\Delta h< \frac{\kappa (\Delta t)^2}{\sqrt{m}}$, we then have that:
\begin{equation*}
\begin{aligned}
||z|| & \leq \sqrt{m} \Delta h+\sqrt{m}a (t_l+\Delta t)+\kappa t_l^{2}\\
& < \kappa((\Delta t)^2+t^{2}_l)+\sqrt{m}a (t_l+\Delta t)\\
& \leq \kappa((\Delta t)^2+t^{2}_l+2 \Delta t t_l)+\sqrt{m}a (t_l+\Delta t)\\
&= \kappa (t_l+\Delta t)^2+\sqrt{m}a (t_l+\Delta t)\\
&=\kappa t^{2}_{l+1}+\sqrt{m}a t_{l+1}
\end{aligned}
\end{equation*}
The last inequality in the chain above is a consequence of $t_l \geq 0$. Thus $H_y \subset G^{t_{l+1}}_{\sqrt{m}a,\kappa}$. Denote $\mathcal{V}(H_y)$ as the set of vertices of $H_y$. Since by construction $\mathcal{V}(H_y) \subset GR_{\Delta h}$ we have that $\mathcal{V}(H_y) \subset D_{l+1}$.\\
Let us study the support function of $B_{0}(\sqrt{m}a (t_l+\Delta t)+\kappa t_l^{2})$:
\begin{equation*}
\begin{aligned}
\sigma_{B_{0}(\sqrt{m}a (t_l+\Delta t)+\kappa t_l^{2})}(y)&= \underset{x \in B_{0}(\sqrt{m}a (t_l+\Delta t)+\kappa t_l^{2})}{\max} \langle x , y \rangle\\
&=\langle x^{*}(y), y \rangle \\
& \leq \underset{x \in H_y}{\max} \langle x, y \rangle,
\end{aligned}
\end{equation*}
where the last inequality follows since $x^{*}(y) \in H_y$.
Since the optima of a linear function over a compact polytope is always attained at the polytopes vertices we have that:
\begin{equation*}
\underset{x \in H_y}{\max} \langle y,x \rangle=\underset{x \in \mathcal{V}(H_y)}{\max} \langle y,x \rangle.
\end{equation*}
Using the previously derived conclusion $\mathcal{V}(H_y) \subset D_{l+1}$ and the fact that $D_{l+1} \subset D^{conv}_{l+1}$ we have :
\begin{equation*}
\begin{aligned}
\sigma_{B_{0}(\sqrt{m}a (t_l+\Delta t)+\kappa t_l^{2})}(y) & \leq  \underset{x \in H_y}{\max} \langle x, y \rangle\\
& =\underset{x \in \mathcal{V}(H_y)}{\max} \langle y,x \rangle\\
& \leq \underset{x \in D_{l+1}}{\max} \langle y,x \rangle\\
& \leq \underset{x \in D^{conv}_{l+1}}{\max} \langle y,x \rangle\\
&=\sigma_{D^{conv}_{l+1}}(y).
\end{aligned}
\end{equation*}
From the dominance of the support function we can conclude that $B_{0}(\sqrt{m}a (t_l+\Delta t)+\kappa t_l^{2}) \subseteq D^{conv}_{l+1}$ using Theorem 3.3.1 in \cite{hiriart2004fundamentals}.
\subsection{Computation of time stepping operator.}\label{compute_time_time_stpp_op}
Let us assume for a point $x \in D_l$ we are able to identify the other grid points of $D_{l+1}$ within $B_x(\sqrt{m}a \Delta t)$. Let us first re-write the time stepping operator:
\begin{equation}\label{alt_rep_tim_stp_op}
\begin{aligned}
F_{lce}(l,u_{l+1},GR_{\Delta h})(x) & =\underset{y \in B_x(\sqrt{m}a\Delta t)}{\sup} \hspace{0.2cm} \underset{s \in \partial g_{l+1}(y)}{\max} \{ \Delta t H(s)+ g_{l+1}(y)+\langle s , x -y \rangle \}\\
&=\underset{y \in B_x(\sqrt{m}a\Delta t)}{\sup} \hspace{0.2cm} \underset{s \in \partial g_{l+1}(y)}{\max} \{ \Delta t H(s)+ g_{l+1}(y)-\langle s , y \rangle+\langle s, x \rangle \}\\ 
&=\underset{y \in B_x(\sqrt{m}a\Delta t)}{\sup} \hspace{0.2cm} \underset{s \in \partial g_{l+1}(y)}{\max} \{ \Delta t H(s)- g^{*}_{l+1}(s)+\langle s, x \rangle \}\\
&=\underset{y \in B_x(\sqrt{m}a\Delta t)}{\sup} \hspace{0.2cm} \underset{s \in \partial g_{l+1}(y)}{\max} \{ \Delta t H(s)- g^{*}_{l+1}(s)+\langle s, x \rangle \}\\
&= \underset{a \in [K]}{\max}\underset{y \in B_x(\sqrt{m}a\Delta t)}{\sup} \hspace{0.2cm} \underset{s \in \partial g_{l+1}(y)}{\max} \{ \Delta t H_a(s)- g^{*}_{l+1}(s)+\langle s, x \rangle \}
\end{aligned}
\end{equation}
where $g^{*}_{l+1}$ refers to the convex conjugate of lce $g_{l+1}$. The final equation is a maximization over all possible sub-gradient of $g_{l+1}$ over the set $B_x(\sqrt{m}a\Delta t)$ for a fixed arm $a$ and then a further maximization over the arms. As the $\hat{V}_{l+1}$ is a piecewise linear function, its lce $g_{l+1}$ is also piecewise linear. The lce maybe constructed by constructing the convex hull of the epigraph of $\hat{V}_{l+1}$ over the set $B_x(\sqrt{m}a\Delta t)$. The number of grid points in $B_x(\sqrt{m}a\Delta t)$ is going to be $O( \left(\frac{\Delta t}{\Delta h} \right)^{m})$. We need to calculate the convex hull of $\{(y,\hat{V}_{l+1}(y)) | y \in D_{l+1} \cap B_x(\sqrt{m}a\Delta t) \}$. Algorithms like QuickHull (see \cite{barber1996quickhull}) have a worst case complexity of $O(n^{m/2})$ while performing much better on average $O(n/log(n) poly(m))$, where $n$ is the number of points under consideration. So, assuming $\Delta h= \theta(\Delta t^2)$ (this was a requirement in Lemma \ref{lemma_grid_scheme_consistency}), we can compute the lower convex envelope in $O(\frac{1}{\Delta t^{m^2/2}})$ computations. Once, the convex hull is computed, the sub-differential at each point $y \in B_x(\sqrt{m}a\Delta t)$ can be computed in $O(m)$ time as the normal vectors to the convex hull. \\

Note that the objective in the final equation in \eqref{alt_rep_tim_stp_op}, 
\begin{equation*}
\Delta t H_a(s)- g^{*}_{l+1}(s)+\langle s, x \rangle
\end{equation*}
depends only on the sub-differential $s$ directly and only implicitly on $y$.  Further the objective function is concave in $s$. These observations imply that we need to restrict $y$ to just the vertices present in $g_{l+1}$ in the maximization over $y \in B_x(\sqrt{m}a\Delta t)$. Due to the piecewise linearity of $g_{l+1}$ to evaluate $g^{*}_{l+1}$ it is enough to evaluate linear functionals at only the vertices present in $g_{l+1}$. Thus by using for example, a projected sub-gradient method (see \cite{boyd2003subgradient} for details) one can solve the innermost maximization with complexity $O( \left( \frac{1}{\Delta t}\right)^{m^2})$. As this has to be done for $K$ arms we get the total computational complexity to be $O\left (  \frac{K}{(\Delta t)^{m^2}}\right)$. Thus the computational effort of building the convex hull $O(\frac{1}{\Delta t^{m^2/2}})$ and then solving the optimization problem $O\left (  \frac{K}{(\Delta t)^{m^2}}\right)$ implies the total computational effort at evaluating the time stepping operator at a point $x \in D_l$ is $O\left (  \frac{K}{(\Delta t)^{m^2}}\right)$.
\subsection{Proof of Theorem \ref{goap_ldp_guarantee}.}\label{proof_goap_ldp_guarantee}
\begin{proof}
We will utilize the method of proof introduced in \cite{tarasyev1999control}. Given an initial position $x_{t_l}$ at $t_l \in GR_{\Delta t}$ and let us assume GOAP plays $a_l$ in the time interval $[t_l,t_l+\Delta t)$, then the state vector $x_{t_l+\Delta t}$ will have the property:
\begin{equation*}
x_{t_l+\Delta t} \in \bigg \{x_{t_l}+\Delta t\left(D(Q_{a_l}||\nu^{i}_{a_l}) \right)_{i\in [m]}: \text{ for any } Q \in (\Delta_{\mathcal{X}})^{m}\bigg\} \tag{P},
\end{equation*}
that is, the state vector would have evolved in a manner as if nature plays a constant control $Q$ in response to the constant control $a_l$ over the interval $[t_l,t_l+\Delta t)$. This is because even if nature was playing a non-constant control $Q_s$ over the interval $[t_l,t_l+\Delta t)$, by convexity of KL divergence we have:
\begin{equation}\label{state_ineq_kl_div_conv}
x_{t_l,i}+\int^{t_l+\Delta t}_{t_l} D(Q_{s,a_l}|| \nu^{(i)}_{a_l}) ds \geq x_{t_l,i}+\Delta t D \left(\frac{\int^{t_l+\Delta t}_{t_l}Q_{s,a_l}ds}{\Delta t} \bigg| \bigg| \nu^{(i)}_{a_l} \right)
\end{equation}
for each $i \in [m]$. Now whether nature played the non-constant control $Q_s$ in the interval or played the constant control $\frac{\int^{t_l+\Delta t}_{t_l}Q_{s,a_l}ds}{\Delta t}$, the GOAP algorithm is indifferent. It would still update its state vector (see line 9 in Algorithm \ref{alg: GOAP}) using only $\frac{\int^{t_l+\Delta t}_{t_l}Q_{s,a_l}ds}{\Delta t}$. However, for nature by inequality \eqref{state_ineq_kl_div_conv} we know that non-constant evolution vector will be greater than the constant evolution vector in all co-ordinates. Since the terminal cost $g^{'}$ is non-decreasing in $x_1$, this implies that the optimal response of nature to GOAP play is a piecewise constant response between intervals $[t,t+\Delta t)$. This means we need to consider only paths with property (P), since in other paths GOAP will do better. In particular, this argument is true for batched algorithms considered under the "meta" strategy in section \ref{sec:approachability}. This property (P) of the specific differential game (eqs. \eqref{dyn_eq}, \eqref{term_cost}) we consider is what allows us to extend the results of \cite{tarasyev1999control} to the nonlinear dynamics (eq. \ref{dyn_eq}) \footnote{nonlinear in player controls}. The results in \cite{tarasyev1999control} were restricted to just linear dynamics because in general zero sum differential games property $(P)$ is not true.\\

Let us introduce the notation $x_{t+\Delta t}(x_t,w_t)$ for the set of all possible state vectors that can evolve from the initial position $x_t$ at time $t \in [0,1-\Delta t)$ under the constant agent control $w_t$ over the time interval $[t,t+\Delta t]$. Further, from property (P) we know that every point in $x_{t+\Delta t}(x_t,w_t)$ corresponds to some constant response $Q$ of nature. Let us denote the point in $x_{t+\Delta t}(x_t,w_t)$ corresponding to $Q$ as $x_{t+\Delta t}(x_t,w_t)(Q)$. We can show the following claim:
\begin{claim}\label{basic_recursive_claim}
Suppose $y_{j_l} \in D_l$ and $a^{*}_l(y_{j_l})$ is the optimal control defined for $y_{j_l}$ by eq. \eqref{optimal_action_policy} then
\begin{equation}
 \underset{z \in x_{t_l+\Delta t}(y_{j_l},a^{*}_l(y_{j_l}))}{\inf}  \hat{V}_{l+1}(z) \geq \hat{V}_l(y_{j_l}). 
\end{equation}
\end{claim}
Now for any point $x_{t_l} \in G^{t_l}_{\sqrt{m}a,\kappa}$, we let (as in eq. \eqref{proj_x_l_grid}) that:
\begin{equation*}
y_{j_l} \in \underset{y \in D_l}{argmin}||x_{t_l}-y||.    
\end{equation*}
This step ensures that $a^{*}_l(x_{t_l})=a^{*}_l(y_{j_l})$ \footnote{$a^{*}_l(.)$ is the action chosen by GOAP at  $t_l \in GR_{\Delta t}$ for any $x \in G^{t_l}_{\sqrt{m}a,\kappa}$.} . Further, by Property (P) for any of nature's control $Q$ we have that:
\begin{equation}
||x_{t_l+\Delta t}(x_{t_l},a^{*}_l(x_{t_l}))(Q)-x_{t_l+ \Delta t}(y_{j_l},a^{*}_l(y_{j_l}))(Q)||=||x_{t_l}-y_{j_l}||.
\end{equation}
Now, we know that $\hat{V}_l$ is a lipschitz function (see the discussion after lemma \ref{lemma_grid_scheme_consistency}) with lipschitz constant of $c^{'}(m) L_{g^{'}}$. Thus, we have that for any $Q$:
\begin{equation}
\begin{aligned}
\hat{V}_l(x_{t_l}) & \stackrel{(a)}{\leq }\hat{V}_{l}(y_{j_l})+c^{,}(m)L_{g^{'}}||x_{t_l}-y_{j_l}||\\
& \stackrel{(b)}{\leq }\hat{V}_{l+1}(x_{t_l+ \Delta t}(y_{j_l},a^{*}_l(y_{j_l}))(Q))+c^{,}(m)L_{g^{'}}||x_{t_l}-y_{j_l}||\\
& \stackrel{(a)}{\leq }\hat{V}_{l+1}(x_{t_l+ \Delta t}(x_{t_l},a^{*}_l(x_{t_l}))(Q))+2c^{,}(m)L_{g^{'}}||x_{t_l}-y_{j_l}||\\
& \stackrel{(c)}{\leq }\hat{V}_{l+1}(x_{t_l+ \Delta t}(x_{t_l},a^{*}_l(x_{t_l}))(Q))+C(m)L_{g^{'}} \sqrt{m} (\Delta t)^2,
\end{aligned}
\end{equation}
where the steps marked as $(a)$ follow from the aforementioned Lipschitzness of $\hat{V}_l$, $(b)$ follows from Claim \ref{basic_recursive_claim} and $(c)$ follows from the hypothesis $\Delta h =O((\Delta t)^2)$ assumed in the Theorem statement. Now in the above inequality if we substitute $Q=Q_l$, the actual observed empirical distribution in response to the constant control $a^{*}_l(x_{t_l})$ in $[t_l,t_{l+1}=t_l+\Delta t]$, we get that:
\begin{equation}\label{recursive_approx_value_ineq}
 \hat{V}_l(x_{t_l}) \leq \hat{V}_{l+1}(x_{t_{l+1}}) +C(m)L_{g^{'}} \sqrt{m} (\Delta t)^2 . 
\end{equation}
This is because we have that $x_{t_{l+1}}=x_{t_l+ \Delta t}(x_{t_l},a^{*}_l(x_{t_l}))(Q_l)$ by property (P). Using induction on $l$ and the terminal condition $\hat{V}_{B}(x)=g^{'}(x)$ we can conclude:
\begin{equation*}
\begin{aligned}
\hat{V}_0(x_{t_0}) & \leq  g^{'}(x_1)+C(m)BL_{g^{'}} \sqrt{m} (\Delta t)^2 \\
& = g^{'}(x_1)+C(m)L_{g^{'}} \sqrt{m} \Delta t.
\end{aligned}
\end{equation*}
Since $x_{t_0}=0$ and $x_1$ is the final state that evolved at $t=1$ under the actions played by GOAP and nature we have that:
\begin{equation*}
\hat{V}_0(0) \leq g^{'}(x_1)+C(m)L_{g^{'}} \sqrt{m} \Delta t.
\end{equation*}
Now using Theorem \ref{convg_thm_backwrd_stp_op}, lemma \ref{domain_of_dependence} and Theorem \ref{hji_pde_R_go_infty} , we conclude that:
\begin{equation*}
R^{go}_{\infty}-C \sqrt{\Delta t} \leq \hat{V}_0(0) \leq g^{'}(x_1)+C(m)L_{g^{'}} \sqrt{m} \Delta t.
\end{equation*}
Thus we have that:
\begin{equation}
\begin{aligned}
g^{'}(x_1) & \geq R^{go}_{\infty}-C \sqrt{\Delta t}-C(m)L_{g^{'}} \sqrt{m} \Delta t\\
&=R^{go}_{\infty}-C^{'} \sqrt{\Delta t}.
\end{aligned}
\end{equation}
We have effectively shown that for any sample path produced by GOAP and nature the terminal function $g^{'}$ (which is the likelihood of the second most likely bandit) is atleast $R^{go}_{\infty}-C^{'} \sqrt{\Delta t}$. Observing that GOAP is a particular case of the "meta" strategy defined in Section \ref{sec:approachability} we can utilize the lemma \ref{approach_relate_err_exp} to conclude that :
\begin{equation*}
e_{m}(\xi,GOAP) \geq R^{go}_{\infty}-C \sqrt{\Delta t}.
\end{equation*}
This  concludes the proof of Theorem \ref{goap_ldp_guarantee}. We finally give the proof of Claim \ref{basic_recursive_claim}.\\
\\
\textbf{Proof of Claim \ref{basic_recursive_claim}:}
As $g_{l+1}(.)$ is the  local lower convex envelope (lce) of $\hat{V}_{l+1}$ we have that:
\begin{equation*}
\hat{V}_{l+1}(z) \geq g_{l+1}(z)
\end{equation*}
for all $z \in B_{y_{j_l}}(\sqrt{m} a \Delta t)$. We note that by construction:
\begin{equation*}
\underset{w \in \Delta_K}{\cup} x_{t_l+\Delta t}(y_{j_l},w) \subset B_{y_{j_l}}(\sqrt{m} a \Delta t).
\end{equation*}
Using the sub-differential inequality of $g_{l+1}$ we get that:
\begin{equation*}
\hat{V}_{l+1}(z) \geq g_{l+1}(z) \geq g_{l+1}(y)+\langle s, z-y \rangle
\end{equation*}
where $s \in \partial g_{l+1} (y)$ for any $y,z \in B_{y_{j_l}}(\sqrt{m} a \Delta t)$. Suppose, we choose $y=y^{*}(y_{j_l})$ and $s^{*}(y_{j_l}) \in \partial g_{l+1}(y^{*})$, the respective maximizers in the optimization:
\begin{equation*}
\underset{y \in B_{y_{j_l}}(\sqrt{m}a\Delta t)}{\sup} \hspace{0.2cm} \underset{s \in \partial g_{l+1}(y)}{\max} \{ \Delta t H(s)+ g_{l+1}(y)+\langle s , y_{j_l} -y \rangle \},
\end{equation*}
we get that:
\begin{equation*}
\hat{V}_{l+1}(z) \geq g_{l+1}(z) \geq g_{l+1}(y^{*}(y_{j_l}))+\langle s^{*}(y_{j_l}), z-y^{*}(y_{j_l}) \rangle.
\end{equation*}
If $z \in x_{t_l+ \Delta t}(y_{j_l},a^{*}_l(y_{j_l}))$, then we have that 
\begin{equation*}
z=y_{j_l}+\Delta t \left(D(Q_{a^{*}_l(y_{j_l})}|| \nu^{(i)}_{a^{*}_l(y_{j_l})} \right)_{i \in [m]}
\end{equation*}
for some $Q$ and $z=x_{t_l+ \Delta t}(y_{j_l},a^{*}_l(y_{j_l}))(Q)$. Substituting this in the above inequality and infimizing over $Q$ we get:
\begin{equation*}
 \underset{z \in x_{t_l+ \Delta t}(y_{j_l},a^{*}_l(y_{j_l}))}{\inf}\hat{V}_{l+1}(z) \geq g_{l+1}(y^{*}(y_{j_l}))+\langle s^{*}(y_{j_l}), y_{j_l} -y^{*}(y_{j_l}\rangle + \underset{Q}{\inf} \Delta t \bigg \langle s^{*}(y_{j_l}), \left(D(Q_{a^{*}_l(y_{j_l})}|| \nu^{(i)}_{a^{*}_l(y_{j_l})} \right)_{i \in [m]} \bigg \rangle
\end{equation*}
From the definition of $H_a(.)$ (recall from Lemma \ref{Hamiltonian_formula_lemma}) we then get that:
\begin{equation*}
 \underset{z \in x_{t_l+ \Delta t}(y_{j_l},a^{*}_l(y_{j_l}))}{\inf} \hat{V}_{l+1}(z) \geq g_{l+1}(y^{*}(y_{j_l}))+\langle s^{*}(y_{j_l}), y_{j_l} -y^{*}(y_{j_l}\rangle + \Delta t H_{a^{*}_l(y_{j_l})}(s^{*}(y_{j_l}))
\end{equation*}
But from eqs. \eqref{lce_time_stp_oper}, \eqref{optimal_action_policy} we get that the RHS in the above inequality is just $\hat{V}_{l}(y_{j_l})$. This proves the claim.
\end{proof}
\section{Proofs of result in Section \ref{sec:approachability}.}
\subsection{Proof of Theorem \ref{sanov_thm}.}\label{sanov_proof}
 Let $\mathcal{P}$ be set of all probability measures on $\mathcal{X}$. Denote all measures on $[m]$ of the form $(n_1/n,n_2/n, \dots,n_m/n)$ where $n_1,\dots,n_m$ are integers as $\mathcal{P}_{n}(m)$. For any $P \in \mathcal{P}_{n}(m)$, denote by $\mathcal{T}_{n}(P)$ the set of $n$ length sequences in which each $i \in [m]$ occurs $n_i$ may times.

We have the following two simple lemma (see chapter 2 \cite{dembo2009large}):
\begin{lemma}[Lemma 2.1.2 (a) in \cite{dembo2009large}]\label{lemma_b_1}
\[
|\mathcal{P}_{n}(m)| \leq (n+1)^{m}.\]
\end{lemma}
\begin{lemma}[Lemma 2.1.9 in \cite{dembo2009large}]\label{lemma_b_2}
For any $P \in \mathcal{P}_{n}(m)$, with $n$ i.i.d draws from a distribution $\mu$ on $[m]$ we have that:
\[
(n+1)^{-m}e^{-n D(P||\mu)} \leq \mu^{n}(\mathcal{T}_{n}(P)) \leq e^{-n D(P||\mu)}
\]
\end{lemma}
 Let us denote the set $\mathcal{P}_{\alpha T/B}(|\mathcal{X}|)$ as $\mathcal{P}_{\alpha}$ where $\alpha \in [0,1]$. If $w \in \Delta_K$ then we denote the collection $(\mathcal{P}_{w(a)})_{a \in [K]}$ as $\mathcal{P}^{K}_w$. Further, we denote the collection of $(\mathcal{P}^{K}_{w_j(Q^{j-1})})_{j \in [B]}$ for every possible $Q^{B}$ as $\mathcal{P}^{K}(w^{B})$. Consider any set $\Gamma \subset \mathcal{P}^{BK}$.
\begin{equation*}
\begin{aligned}
P_{\nu}[Q^{B} \in \Gamma]&=\sum_{\mu^{B} \in \Gamma \cap \mathcal{P}^{K}(w^{B})}P_{\nu}[Q^{B}=\mu^{B}]\\
&=\sum_{\mu^{B} \in \Gamma \cap \mathcal{P}^{K}(w^{B})}\prod^{B}_{j=1}P_{\nu}[Q_j=\mu_j|Q^{j-1}=\mu^{j-1}].
\end{aligned}
\end{equation*}
Define the following sets for $j=1,2,\dots,B$
$$
\Gamma^{-1}_j(\mu^{j-1})=\bigg\{ \mu \in \mathcal{P}^{K} \bigg| \exists \mu^{'}_{j+1},\mu^{'}_{j+2},\dots, \mu^{'}_B \in \mathcal{P}^{K} , \text{ s. that } (\mu^{j-1},\mu,\mu^{'}_{j+1},\mu^{'}_{j+2},\dots, \mu^{'}_B) \in \Gamma \bigg\}
$$. Using the notation defined we have that:
\begin{equation*}
\begin{aligned}
P_{\nu}[Q^{B} \in \Gamma]&=\sum_{\mu^{B} \in \Gamma \cap \mathcal{P}^{K}(w^{B})}\prod^{B}_{j=1}P_{\nu}[Q_j=\mu_j|Q^{j-1}=\mu^{j-1}]\\
&=\sum_{\mu_1 \in \Gamma^{-1}_1(\phi) \cap \mathcal{P}^{K}_{w_1}}\sum_{\mu_2 \in \Gamma^{-1}_2(\mu^{1}) \cap \mathcal{P}^{K}_{w_2(\mu^{1})}}\dots \sum_{\mu_B \in \Gamma^{-1}_B(\mu^{B-1}) \cap \mathcal{P}^{K}_{w_B(\mu^{B-1})}} \prod^{B}_{j=1}P_{\nu}[Q_j=\mu_j|Q^{j-1}=\mu^{j-1}].
\end{aligned}
\end{equation*}
Bow we note that:
\begin{equation*}
\begin{aligned}
P_{\nu}[Q_j=\mu_j|Q^{j-1}=\mu^{j-1}]&=\prod^{K}_{a=1}P_{\nu}[Q_{a,j}=\mu_{a,j}|Q^{j-1}=\mu^{j-1}]\\
&=\prod^{K}_{a=1}\nu_a^{w_j(a)(\mu^{j-1})T/B}(\mathcal{T}_{w_j(a)T/B}(\mu_{a,j}))\\
& \leq \prod^{K}_{a=1}e^{-w_j(a)(\mu^{j-1})T/B D(\mu_{a,j}||\nu_a)}\\
&=e^{-\frac{T}{B}\sum^{K}_{a=1}w_j(a)(\mu^{j-1}) D(\mu_{a,j}||\nu_a)}.
\end{aligned}
\end{equation*}
where for the last step we used Lemma \ref{lemma_b_2}. Thus, we have the following.
\begin{equation*}
\begin{aligned}
P_{\nu}[Q^{B} \in \Gamma]& \leq \sum_{\mu_1 \in \Gamma^{-1}_1(\phi) \cap \mathcal{P}^{K}_{w_1}}\sum_{\mu_2 \in \Gamma^{-1}_2(\mu^{1}) \cap \mathcal{P}^{K}_{w_2(\mu^{1})}}\dots \sum_{\mu_B \in \Gamma^{-1}_B(\mu^{B-1}) \cap \mathcal{P}^{K}_{w_B(\mu^{B-1})}} e^{-\frac{T}{B}\sum^{B}_{j=1}\sum^{K}_{a=1}w_j(a)(\mu^{j-1}) D(\mu_{a,j}||\nu_a)}\\
& \leq  \bigg|\Gamma \cap \mathcal{P}^{K}(w^{B}) \bigg|e^{-\underset{\mu^{B} \in \Gamma \cap \mathcal{P}^{K}(w^{B})}{\inf}\frac{T}{B}\sum^{B}_{j=1}\sum^{K}_{a=1}w_j(a)(\mu^{j-1}) D(\mu_{a,j}||\nu_a)}
\end{aligned}
\end{equation*}
Now we observe from Lemma \ref{lemma_b_1} and $w_j \in \Delta_K$ that:
$$
|\mathcal{P}_{w_j(a)(Q^{j-1})}| \leq (1+w_j(a)(Q^{j-1})T/B)^{|\mathcal{X}|} \leq (1+T/B)^{|\mathcal{X}|}
$$
and hence it follows that:
\begin{equation*}
\bigg|\mathcal{P}^{K}(w^{B}) \bigg| \leq (1+T/B)^{BK |\mathcal{X}|}.
\end{equation*}
From the fact that $\bigg|\Gamma \cap \mathcal{P}^{K}(w^{B}) \bigg| \leq \bigg|\mathcal{P}^{K}(w^{B}) \bigg|$ it follows that:
\begin{equation*}
P_{\nu}[Q^{B} \in \Gamma] \leq (1+T/B)^{B|\mathcal{X}|} e^{-\underset{\mu^{B} \in \Gamma \cap \mathcal{P}^{K}(w^{B})}{\inf}\frac{T}{B}\sum^{B}_{j=1}\sum^{K}_{a=1}w_j(a)(\mu^{j-1}) D(\mu_{a,j}||\nu_a)}.
\end{equation*}
From this we conclude that:
\begin{equation}
\begin{aligned}
\underset{T \to \infty}{\liminf}-\frac{\log(P_{\nu}[Q^{B} \in \Gamma])}{T} & \geq  \underset{T \to \infty}{\liminf}\underset{\mu^{B} \in \Gamma \cap \mathcal{P}^{K}(w^{B})}{\inf}\frac{\sum^{B}_{j=1}\sum^{K}_{a=1}w_j(a)(\mu^{j-1}) D(\mu_{a,j}||\nu_a)}{B}\\
& \geq \underset{\mu^{B} \in \Gamma}{\inf}\frac{\sum^{B}_{j=1}\sum^{K}_{a=1}w_j(a)(\mu^{j-1}) D(\mu_{a,j}||\nu_a)}{B}
\end{aligned}
\end{equation}
\section{Proofs related to Section \ref{sec:construction b_set}.} \label{section_5_proofs}
\subsection{Proof of equation \eqref{beta_optimization}.}\label{beta_optim_proof}
We first consider the optimization:
\begin{equation*}
\underset{Q_a}{\inf} \hspace{0.2cm} \sum^{K}_{i=1}\beta_i D(Q_a||\nu^{i}_a).   
\end{equation*}
Now the optimization objective function can be re-written as:
\begin{equation*}
\begin{aligned}
\sum^{K}_{i=1}\beta_i D(Q_a||\nu^{i}_a) &= \sum^{K}_{i=1}\beta_i \sum_{x \in \mathcal{X}} Q_a(x) \log \left( \frac{Q_a(x)}{\nu_a^{i}(x)}\right)\\
&= \sum_{x \in \mathcal{X}} Q_a(x) \sum^{K}_{i=1}\beta_i \log \left( \frac{Q_a(x)}{\nu_a^{i}(x)}\right)\\
& = \sum_{x \in \mathcal{X}} Q_a(x) \log \left( \prod^{K}_{i=1} \left(\frac{Q_a(x)}{\nu_a^{i}(x)} \right)^{\beta_i} \right)\\
& = \sum_{x \in \mathcal{X}} Q_a(x) \log \left( \frac{Q_a(x)}{\prod^{K}_{i=1} (\nu_a^{i}(x))^{\beta_i}}  \right)\\
& = \sum_{x \in \mathcal{X}} Q_a(x) \log \left( \frac{Q_a(x)}{\prod^{K}_{i=1} (\nu_a^{i}(x))^{\beta_i}}  \right)\\
& = \sum_{x \in \mathcal{X}} Q_a(x) \log \left( \frac{Q_a(x) (\sum_{x \in \mathcal{X}}\prod^{K}_{i=1} (\nu_a^{i}(x))^{\beta_i})}{\prod^{K}_{i=1} (\nu_a^{i}(x))^{\beta_i}}  \right) - \log  \left( \sum_{x \in \mathcal{X}}\prod^{K}_{i=1} (\nu_a^{i}(x))^{\beta_i} \right) \\
&= D(Q_a || p(\xi,\beta)) - \log  \left( \sum_{x \in \mathcal{X}}\prod^{K}_{i=1} (\nu_a^{i}(x))^{\beta_i} \right)
\end{aligned}
\end{equation*}
where $p(\xi,\beta)$ is a probability measure whose pmf at $x \in \mathcal{X}$ satisfies:
\begin{equation*}
p(\xi,\beta)(x) =\frac{\prod^{K}_{i=1} (\nu_a^{i}(x))^{\beta_i}}{\sum_{x \in \mathcal{X}}\prod^{K}_{i=1} (\nu_a^{i}(x))^{\beta_i}}.
\end{equation*}
But from the last expression in the earlier algebraic manipulation as the KL divergence has a minimum of zero we note that the optimizer is $Q^{*}_a=p(\xi,\beta)$ with the optimal value being $-\log  \left( \sum_{x \in \mathcal{X}}\prod^{K}_{i=1} (\nu_a^{i}(x))^{\beta_i} \right)$.

\subsection{Proof of Lemma \ref{subdiff_conj_lemma}.}\label{subdiff_conj_lemma_proof}
Suppose $\beta \in \underset{b \in \Delta_n}{argmin} \{ \langle b , x\rangle -f(b) \}$ (we know from basic convex optimization that this argmin set is convex and compact) then we have:
\begin{equation*}
\begin{aligned}
f^{*}(y)&=\underset{b \in \Delta_n}{\inf} \{ \langle b , y\rangle -f(b) \}\\
&= \underset{b \in \Delta_n}{\inf} \{ \langle b , x\rangle+\langle b, y-x \rangle -f(b) \}\\
& \leq  \langle \beta , x\rangle -f(\beta) +\langle \beta, y-x \rangle\\
&= f^{*}(x)+ \langle \beta, y-x \rangle
\end{aligned}
\end{equation*}
Thus $\beta \in \partial f^{*}(x)$ and hence $\underset{b \in \Delta_n}{argmin} \{ \langle b , x\rangle -f(b) \} \subseteq \partial f^{*}(x)$.

Now in the other direction we first make the observation that $f$ is continuous over the simplex (using Theorem 10.2 of \cite{rockafellar1997convex}). Now consider the extended value function:
\begin{equation*}
\tilde{f}(b)= \begin{cases}
f(b), & b \in \Delta_n\\
-\infty, & b \notin \Delta_n.
\end{cases}
\end{equation*}
which extends $f$ over the entire euclidean space. Due to the continuity of $f$ in $\Delta_n$ we see that $\tilde{f}$ is both proper and upper semi-continuous. Further we make the observation that the concave conjugate of $f$ and $\tilde{f}$ is the same, that is, $f^{*}$. Now by the Fenchel-Moreau Theorem (Theorem 4.2.1 \cite{borwein2006convex}) we have that:
\begin{equation*}
\tilde{f}^{**}=\tilde{f}
\end{equation*}
and hence we have the following equality holding:
\begin{equation*}
\tilde{f}(b)=\underset{x \in R^{n}}{\min} \{ \langle b ,x \rangle -f^{*}(x) \}.    
\end{equation*}
and whenever $b \in \Delta_n$, $f(b)=\underset{x \in R^{n}}{\min} \{ \langle b ,x \rangle -f^{*}(x) \}$. Now from the very definition of a subdifferential we have:
\begin{equation*}
f^{*}(y) \leq f^{*}(x)+ \langle \partial f^{*}(x),y-x \rangle
\end{equation*}
which implies that $\forall y \in \mathbb{R}^{n}$:
\begin{equation*}
\langle \partial f^{*}(x),x\rangle-f^{*}(x) \leq  \langle \partial f^{*}(x),y\rangle-f^{*}(y)  
\end{equation*}
Taking infimum over $y$ on the RHS we obtain that:
\begin{equation*}
\langle \partial f^{*}(x),x\rangle-f^{*}(x) \leq \tilde{f}(\partial f^{*}(x)).
\end{equation*}
Hence from the variational form of $\tilde{f}$ we obtain that:
\begin{equation*}
\langle \partial f^{*}(x),x\rangle-f^{*}(x) = \tilde{f}(\partial f^{*}(x)).
\end{equation*}
Now from the fact that the subdifferential is bounded in interior of $dom(f^{*})=\mathbb{R}^{n}$ we have that:
\begin{equation}
-\infty < \tilde{f}(\partial f^{*}(x)).
\end{equation}
But this by definition of $\tilde{f}$ implies that $\partial f^{*}(x)$ is in $\Delta_m$ and we have:

\begin{equation*}
\langle \partial f^{*}(x),x\rangle-f^{*}(x) = f(\partial f^{*}(x)).
\end{equation*}.
But this equality can only hold if $\partial f^{*}(x) \in \underset{b \in \Delta_n}{argmin} \{ \langle b , x\rangle -f(b) \}$. Hence $\partial f^{*}(x) \subseteq \underset{b \in \Delta_n}{argmin} \{ \langle b , x\rangle -f(b) \}$.

\subsection{Proof of lemma \ref{normal_sublvl}}\label{normal_sublvl_proof}
We directly invoke the following result from \cite{hiriart2004fundamentals}:
\begin{theorem}[Theorem 1.3.5, Chapter VI \cite{hiriart2004fundamentals}]
Let $h$ be a finite valued convex function and let $z$ be such that $0 \notin \partial h(z)$. Then, for the sublevel set :
\begin{equation*}
    D=\{y | h(y) \leq h(z) \}
\end{equation*}
we have that:
\begin{equation*}
N_D(z)=Cone(\partial h(z))
\end{equation*}
\end{theorem}
Using the above theorem by setting $h=-f$ and $z=\bar{x}$ and observing that:
\begin{equation*}
f(\bar{x}) < \sup f(x) \implies 0 \notin \partial f(\bar{x})  
\end{equation*}
and observing that $C=D$ and $\partial h(\bar{x})= \partial -f(\bar{x})$ we immediately get the result in the lemma.

\subsection{Proof of Theorem \ref{blackwell_approach_thm}}\label{proof_blackwell_approach_thm}
\begin{proof}
Let us show that under the given assumptions we have $I^{'}_{concave}(\beta) \leq \underset{a \in [K]}{\max}-\log \left( \sum_{x \in \mathcal{X}} \prod^{K}_{i=1} (\nu^{i}_a(x))^{\beta_i} \right) := L(\beta)$. We have that:
\begin{equation*}
\begin{aligned}
\underset{\beta \in \Delta_K}{\sup} \{I^{'}_{concave}(\beta)-L(\beta)\}& \stackrel{(a)}{=}\underset{\alpha \in [0,1]}{\sup}\underset{\lambda \in \Delta_{M(R)}}{\sup} \underset{\beta^{'} \in \Delta_K}{\sup} \bigg \{(1-\alpha)R+\alpha I( \beta^{'})-L \left( (1-\alpha)\sum_{(i,j) \in M(R)}\lambda_{ij}\tilde{\beta}_{ij}+ \alpha \beta^{'}\right)\bigg \}\\
&\stackrel{(e)}{=}\underset{\alpha \in [0,1]}{\sup}\underset{\lambda \in \Delta_{M(R)}}{\sup} \underset{\beta^{'} \in \Delta_K}{\sup} \bigg \{(1-\alpha)R+\alpha I( \beta^{'})- \underset{a \in [K]}{\max}I \left( \delta_a, (1-\alpha)\sum_{(i,j) \in M(R)}\lambda_{ij}\tilde{\beta}_{ij}+ \alpha \beta^{'}\right)\bigg \}\\
& \stackrel{(b)}{ \leq }\underset{\alpha \in [0,1]}{\sup}\underset{\lambda \in \Delta_{M(R)}}{\sup} \underset{\beta^{'} \in \Delta_K}{\sup} \bigg \{(1-\alpha)R+\alpha I( \beta^{'})- \underset{a \in [K]}{\max} \bigg[(1-\alpha)I \left( \delta_a, \sum_{(i,j) \in M(R)}\lambda_{ij}\tilde{\beta}_{ij} \right)+ \alpha I(\delta_a ,\beta^{'})\bigg] \bigg \}\\
& \stackrel{(c)}{ \leq } \underset{\alpha \in [0,1]}{\sup}\underset{\lambda \in \Delta_{M(R)}}{\sup} \underset{\beta^{'} \in \Delta_K}{\sup} \bigg \{(1-\alpha)R+\alpha I( \beta^{'})- \underset{a \in [K]}{\max} \bigg[(1-\alpha)I \left( \delta_a, \sum_{(i,j) \in M(R)}\lambda_{ij}\tilde{\beta}_{ij} \right)+ \alpha I(\beta^{'}) \bigg]\bigg \}\\
&= \underset{\alpha \in [0,1]}{\sup}\underset{\lambda \in \Delta_{M(R)}}{\sup} \underset{\beta^{'} \in \Delta_K}{\sup} \bigg \{(1-\alpha) \bigg[R- \underset{a \in [K]}{\max} I \left( \delta_a, \sum_{(i,j) \in M(R)}\lambda_{ij}\tilde{\beta}_{ij} \right) \bigg]\bigg \}\\
& \stackrel{(d)}{=}\underset{\alpha \in [0,1]}{\sup}\underset{\lambda \in \Delta_{M(R)}}{\sup}  \bigg \{(1-\alpha) \bigg[R- \underset{a \in [K]}{\max} I \left( \delta_a, \sum_{(i,j) \in M(R)}\lambda_{ij}\tilde{\beta}_{ij} \right) \bigg]\bigg \}\\
&=\underset{\alpha \in [0,1]}{\sup} (1-\alpha) \bigg[R-\underset{\lambda \in \Delta_{M(R)}}{\inf} \underset{a \in [K]}{\max} I \left( \delta_a, \sum_{(i,j) \in M(R)}\lambda_{ij}\tilde{\beta}_{ij} \right) \bigg]\\
&\stackrel{(e)}{=}\underset{\alpha \in [0,1]}{\sup} (1-\alpha) \bigg[R-\underset{\lambda \in \Delta_{M(R)}}{\inf}L \left( \sum_{(i,j) \in M(R)} \lambda_{ij}\tilde{\beta}_{ij} \right) \bigg]\\
& \stackrel{(f)}{\leq}0,
\end{aligned}
\end{equation*}
where $(a)$ is due to the variational representation of $I^{'}_{concave}(\beta)$ (see eq \eqref{var_rep_uce}), $(b)$ is due to the concavity of $I(w,\beta)$ for a fixed $w \in \Delta_K$, $(c)$ is because $I(\beta) \leq I(w,\beta)$ (see subsection \ref{hyperplane_subsec}) , $(d)$ is because objective in the previous step is independent of $\beta^{'}$, $(e)$ is due to definition of $L(\beta)$ and $(f)$ is because of the assumption $R \leq  \underset{\lambda \in \Delta_{M(R)}}{\inf}L \left( \sum_{(i,j) \in M(R)} \lambda_{ij}\tilde{\beta}_{ij} \right)$.\\
\\
As a consequence of $I^{'}_{concave}(\beta) \leq L(\beta)$, we know there exist some $w(\beta) \in \Delta_K$ for every $\beta \in \Delta_K$ such that $I^{'}_{concave}(\beta)=I(w(\beta),\beta)$. Now from Theorem 10.2 of \cite{rockafellar1997convex}, we know that there exist a $w(.)$ that is continuous in $\beta \in \Delta_K$.\\

Now consider any point $y \notin \hat{S}$. Let the unique projection (because $\hat{S}$ is closed convex set) of this point $y$ be $\prod_{\hat{S}}(y)$. Then $l(\prod_{\hat{S}}(y))=0$. Now we observe that $-\infty< l(x) < \infty$ and that $\sup l(x)=+\infty$, hence we apply lemma \ref{normal_sublvl} with $f=l$ to conclude that $y-\prod_{\hat{S}}(y)=-\lambda \partial l(\prod_{\hat{S}}(y))$ for some $\lambda >0$. Now, we apply lemma \ref{subdiff_conj_lemma} with $f=I^{'}_{concave}(.)$ and $f^{*}=l$ to obtain that $\partial l(\prod_{\hat{S}}(y)) = \underset{\beta \in \Delta_n}{argmin} \{ \langle \beta  , \prod_{\hat{S}}(y)\rangle -I^{'}_{concave}(\beta ) \}$. Thus, there exists a $\beta (y) \in \underset{\beta  \in \Delta_n}{argmin} \{ \langle \beta  , \prod_{\hat{S}}(y)\rangle -I^{'}_{concave}(\beta ) \}$ such that $y-\prod_{\hat{S}}(y)=-\lambda \beta (y)$.
We have then that:
\begin{equation*}
\begin{aligned}
 \langle y, \beta (y) \rangle &= \langle \prod_{\hat{S}}(y), \beta (y) \rangle -\lambda || \beta (y)||^{2} \\
 & < I^{'}_{concave}(\beta (y)),
\end{aligned}
\end{equation*}
where we used the fact that $l(\prod_C(y))=0$ and that $\beta (y) \in \underset{\beta  \in \Delta_n}{argmin} \{ \langle \beta , \prod_{\hat{S}}(y)\rangle -I^{'}_{concave}(\beta ) \}$.
Thus the hyperplane $\{x : \langle \beta (y), x \rangle =I^{'}_{concave}(\beta (y))\}$ separates $y$ from $\hat{S}$. But by construction there exists a $w (\beta(y)) \in \Delta_K$ such that (see similar argument in subsection \ref{hyperplane_subsec}):
\begin{equation*}
\langle \beta(y), f(w(\beta(y)),v) \rangle =\sum_{i} \sum_a \beta_i(y) w_a(\beta(y)) D(v_a|| v^{i}_a) \geq  \underset{Q \in V}{\inf}\sum_{i} \sum_a \beta_i(y) w_a(\beta(y)) D(Q_a|| v^{i}) =I^{'}_{concave}(\beta(y)),
\end{equation*}
holds for each $v \in V$. Thus $y$ and $f(w(\beta(y)),V)$ lie on opposing halfspaces. Thus, the B-set condition holds for $\hat{S}$.
\end{proof}
\subsection{Proof of lemma \ref{alt_rep_l_blackwell}.}\label{proof_alt_rep_l_blackwell}
\begin{proof}
From the definition of $l$ we have that:
\begin{equation*}
l(x)=\underset{\beta \in \Delta_m}{\inf} \{ \langle \beta, x \rangle -I^{'}_{concave}(\beta) \}.
\end{equation*}
Using the representation eq. \eqref{var_rep_uce} of $I^{'}_{concave}(\beta)$ we then have that:
\begin{equation*}
 l(x)= \underset{\beta \in \Delta_m}{\inf} \bigg \{ \langle \beta, x \rangle -\underset{\underset{\beta=\alpha(\sum_{(i,j) \in M(R)}\lambda_{i,j}\tilde{\beta}_{i,j})+(1-\alpha)\beta^{'}}{\lambda \in \Delta_{M(R)}; \alpha \in [0,1], \beta^{'} \in \Delta_m}}{\sup}  \{ \alpha R+(1-\alpha)I(\beta^{'})\} \bigg \}
\end{equation*}
From which it follows that:
\begin{equation*}
 l(x)= \underset{\lambda \in \Delta_{M(R)}; \alpha \in [0,1], \beta^{'} \in \Delta_m}{\inf} \bigg \{ \bigg \langle \alpha \left(\sum_{(i,j) \in M(R)}\lambda_{i,j}\tilde{\beta}_{i,j} \right)+(1-\alpha)\beta^{'} , x \bigg \rangle -   \alpha R-(1-\alpha)I(\beta^{'}) \bigg \}
\end{equation*}
We get the result by noticing that the objective is linear in $\lambda$ and $\alpha$.
\end{proof}
\subsection{Proof of lemma \ref{prop_G_R}.}\label{prop_G_R_proof}
The $\tilde{\beta}_{ij}$ are chosen for each pair such that $R> \underset{\beta \in [0,1]}{\sup}I(\beta_{ij})$. Thus only for those choices of $R$ will the function $G(.)$ change its value. This implies that $G(.)$ is a constant between $a_{s}$ and $a_{s+1}$.\\
Now for every $R$ we can attach (as before) the set of all $(i,j)$ pairs that are being perturbed:
\begin{equation*}
M(R)=\{ (i,j) : R> \underset{\beta \in [0,1]}{\sup}I(\beta_{ij}) \}.
\end{equation*}
Now, clearly this set can only grow with $R$, that is if $R_1< R_2$ then  $M(R_1) \subseteq M(R_2)$. Now suppose for $R_1<R_2$ such that $M(R_1) \subsetneq M(R_2)$. We then have that:
\begin{equation*}
\underset{\lambda \in \Delta_{M(R_1)}}{\inf}L \left( \sum_{(i,j) \in M(R_1)} \lambda_{ij}\tilde{\beta}_{ij} \right) \geq \underset{\lambda \in \Delta_{M(R_2)}}{\inf}L \left( \sum_{(i,j) \in M(R_2)} \lambda_{ij}\tilde{\beta}_{ij} \right)
\end{equation*}
If we supremize over $\tilde{\beta_{ij}}$ over all the $(i,j) \in M(R_2)$ pairs we have:
\begin{equation*}
\underset{\tilde{\beta}_{ij}: (i,j) \in M(R_2)}{\sup}\underset{\lambda \in \Delta_{M(R_1)}}{\inf}L \left( \sum_{(i,j) \in M(R_1)} \lambda_{ij}\tilde{\beta}_{ij} \right) \geq \underset{\tilde{\beta}_{ij}: (i,j) \in M(R_2)}{\sup} \underset{\lambda \in \Delta_{M(R_2)}}{\inf}L \left( \sum_{(i,j) \in M(R_2)} \lambda_{ij}\tilde{\beta}_{ij} \right).
\end{equation*}
Notice that $(i,j) \in M(R_2)/M(R_1)$ play no role in the LHS. Hence we have that:
\begin{equation*}
 \underset{\tilde{\beta}_{ij}: (i,j) \in M(R_2)}{\sup}\underset{\lambda \in \Delta_{M(R_1)}}{\inf}L \left( \sum_{(i,j) \in M(R_1)} \lambda_{ij}\tilde{\beta}_{ij} \right)=\underset{\tilde{\beta}_{ij}: (i,j) \in M(R_1)}{\sup}\underset{\lambda \in \Delta_{M(R_1)}}{\inf}L \left( \sum_{(i,j) \in M(R_1)} \lambda_{ij}\tilde{\beta}_{ij} \right)   
\end{equation*}
Therefore:
\begin{equation*}
\underset{\tilde{\beta}_{ij}: (i,j) \in M(R_1)}{\sup}\underset{\lambda \in \Delta_{M(R_1)}}{\inf}L \left( \sum_{(i,j) \in M(R_1)} \lambda_{ij}\tilde{\beta}_{ij} \right) \geq \underset{\tilde{\beta}_{ij}: (i,j) \in M(R_2)}{\sup} \underset{\lambda \in \Delta_{M(R_2)}}{\inf}L \left( \sum_{(i,j) \in M(R_2)} \lambda_{ij}\tilde{\beta}_{ij} \right)
\end{equation*}
which is equivalent to $G(R_1) \geq G(R_2)$ whenever $R_1<R_2$.
\subsection{Proof of lemma \ref{better_than_static}}\label{proof_better_than_static}
Since $R_{approach}=\sup \{ R \leq G(R) \}$ we have that $R_{approach} \leq G(R_{approach})$. Now if $R_{approach}=G(R_{approach})$ then using the fact that $G(R)$ is non-increasing we have that 
\begin{equation*}
R_{approach}=G(R_{approach}) \geq \underset{\tilde{\beta}_{i,j}: (i,j)}{\sup} \underset{\lambda \in \Delta_{K(K-1)/2}}{\inf} L \left(\sum_{(i,j)}\lambda_{i,j}\tilde{\beta}_{i,j} \right).
\end{equation*}
Now if $R_{approach}<G(R_{approach})$ then it must be because:
\begin{equation*}
\underset{\tilde{\beta}_{ij}: (i,j) \in M(a_s)}{\sup}\underset{\lambda \in \Delta_{M(a_s)}}{\inf}L \left( \sum_{(i,j) \in M(a_s)} \lambda_{ij}\tilde{\beta}_{ij} \right) > R_{approach} \geq \underset{\tilde{\beta}_{ij}: (i,j) \in M(a_{s+1})}{\sup}\underset{\lambda \in \Delta_{M(a_{s+1})}}{\inf}L \left( \sum_{(i,j) \in M(a_{s+1})} \lambda_{ij}\tilde{\beta}_{ij} \right) 
\end{equation*}
for some $s \in [K(K-1)/2]$.
Again using the non-increasing nature of G(.) we have that:
\begin{equation*}
R_{approach} \geq \underset{\tilde{\beta}_{ij}: (i,j) \in M(a_{s+1})}{\sup}\underset{\lambda \in \Delta_{M(a_{s+1})}}{\inf}L \left( \sum_{(i,j) \in M(a_{s+1})} \lambda_{ij}\tilde{\beta}_{ij} \right) \geq \underset{\tilde{\beta}_{i,j}: (i,j)}{\sup} \underset{\lambda \in \Delta_{K(K-1)/2}}{\inf} L \left(\sum_{(i,j)}\lambda_{i,j}\tilde{\beta}_{i,j} \right).
\end{equation*}
Thus, in either case we have that:
\begin{equation*}
    R_{approach} \geq \underset{\tilde{\beta}_{i,j}: (i,j)}{\sup} \underset{\lambda \in \Delta_{K(K-1)/2}}{\inf} L \left(\sum_{(i,j)}\lambda_{i,j}\tilde{\beta}_{i,j} \right).
\end{equation*}
Now we have that for any $w \in \Delta_K$ that:
\begin{equation*}
L\left(\sum_{(i,j)}\lambda_{i,j}\tilde{\beta}_{i,j} \right) \geq -\sum_a w_a \log \left( \sum_{x \in \mathcal{X}} \prod^{K}_{i=1} (\nu^{i}_a(x))^{\sum_{(i,j)}\lambda_{i,j}\tilde{\beta}_{i,j}} \right).
\end{equation*}
Now as each function $-\log \left( \sum_{x \in \mathcal{X}} \prod^{K}_{i=1} (\nu^{i}_a(x))^{\beta} \right)$ is concave wrt $\beta$ we have then that :
\begin{equation*}
L\left(\sum_{(i,j)}\lambda_{i,j}\tilde{\beta}_{i,j} \right) \geq \sum_{(i,j)} \lambda_{i,j} \sum_a w_a \left(-\log \left( \sum_{x \in \mathcal{X}} \prod^{K}_{i=1} (\nu^{i}_a(x))^{\tilde{\beta}_{i,j}} \right)\right)
\end{equation*}
From which we obtain that:
\begin{equation*}
\underset{\tilde{\beta}_{i,j}: (i,j)}
{\sup} \underset{\lambda \in \Delta_{K(K-1)/2}}{\inf} L \left(\sum_{(i,j)}\lambda_{i,j}\tilde{\beta}_{i,j} \right) \geq \underset{\tilde{\beta}_{i,j}: (i,j)}
{\sup} \underset{(i,j)}{\min}  \sum_a w_a \left(-\log \left( \sum_{x \in \mathcal{X}} \prod^{K}_{i=1} (\nu^{i}_a(x))^{\tilde{\beta}_{i,j}} \right)\right).
\end{equation*}
Now for each $(i,j)$ we have a corresponding perturbation $\tilde{\beta}_{i,j}$, hence we can take the supremum over $\tilde{\beta}_{i,j}$ inside the minimization over the pairs $(i,j)$ to get that:
\begin{equation*}
\underset{\tilde{\beta}_{i,j}: (i,j)}
{\sup} \underset{\lambda \in \Delta_{K(K-1)/2}}{\inf} L \left(\sum_{(i,j)}\lambda_{i,j}\tilde{\beta}_{i,j} \right) \geq \underset{(i,j)}{\min} \underset{\beta \in [0,1]}{\sup}\sum_a w_a \left(-\log \left( \sum_{x \in \mathcal{X}}  (\nu^{i}_a(x))^{\beta}(\nu^{j}_a(x))^{1-\beta} \right)\right).
\end{equation*}
As this was for arbitrary $w \in \Delta_K$ we have that:
\begin{equation*}
\underset{\tilde{\beta}_{i,j}: (i,j)}
{\sup} \underset{\lambda \in \Delta_{K(K-1)/2}}{\inf} L \left(\sum_{(i,j)}\lambda_{i,j}\tilde{\beta}_{i,j} \right) \geq \underset{w \in \Delta_K}{\sup}\underset{(i,j):i \neq j}{\min} \underset{\beta \in [0,1]}{\sup}-\sum_a w_a \left(\log \left( \sum_{x \in \mathcal{X}}  (\nu^{i}_a(x))^{\beta}(\nu^{j}_a(x))^{1-\beta} \right)\right)=R_{static}
\end{equation*}
From which we conclude that $R_{approach} \geq R_{static}$.
\bibliography{reference}
\bibliographystyle{apalike}
\end{document}